\documentclass[11pt,a4paper]{article}

\usepackage[margin=1in]{geometry}
\usepackage[utf8]{inputenc}
\usepackage[T1]{fontenc}
\usepackage{lmodern}
\usepackage{amsmath}
\usepackage{amssymb}
\usepackage{amsthm}
\usepackage{mathtools}
\usepackage{bm}

\usepackage{graphicx}
\usepackage{booktabs}
\usepackage{multirow}
\usepackage{float}
\usepackage{flafter}  

\usepackage{tikz}
\usetikzlibrary{arrows.meta,positioning,shapes.geometric,shapes.misc,
  fit,calc,backgrounds,decorations.pathreplacing,chains,automata}
\usepackage{pgfplots}
\pgfplotsset{compat=1.18}
\usepgfplotslibrary{statistics}

\usepackage{enumitem}
\usepackage[dvipsnames]{xcolor}
\usepackage{hyperref}
\usepackage{url}
\usepackage[numbers,sort&compress]{natbib}
\usepackage[capitalize,nameinlink]{cleveref}
\usepackage{microtype}

\definecolor{slmink}{HTML}{1F2933}
\definecolor{slmblue}{HTML}{2B6CB0}
\definecolor{slmbluebg}{HTML}{EBF3FB}
\definecolor{slmgreen}{HTML}{2F855A}
\definecolor{slmgreenbg}{HTML}{E7F4EC}
\definecolor{slmamber}{HTML}{B7791F}
\definecolor{slmamberbg}{HTML}{FBF1DE}
\definecolor{slmred}{HTML}{C53030}
\definecolor{slmredbg}{HTML}{FBEAEA}
\definecolor{slmgray}{HTML}{718096}
\definecolor{slmgraybg}{HTML}{EDF0F3}

\tikzset{
  slmbox/.style={rectangle, rounded corners=2pt, draw=slmink, line width=0.5pt,
    fill=white, align=center, inner sep=5pt, font=\small},
  slmstore/.style={cylinder, shape border rotate=90, aspect=0.25, draw=slmgray,
    fill=slmgraybg, align=center, inner sep=3pt, font=\scriptsize},
  slmgate/.style={slmbox, fill=slmbluebg, draw=slmblue},
  slmspine/.style={slmbox, fill=slmgreenbg, draw=slmgreen},
  slmwarn/.style={slmbox, fill=slmamberbg, draw=slmamber},
  slmreject/.style={slmbox, fill=slmredbg, draw=slmred},
  slmflow/.style={-{Stealth[length=2.2mm]}, draw=slmink, line width=0.5pt},
  slmflowdash/.style={slmflow, dashed},
  slmlabel/.style={font=\scriptsize\itshape, text=slmgray},
}

\newtheorem{theorem}{Verified Design Invariant}[section]  

\newtheorem{definition}{Definition}[section]
\theoremstyle{remark}

\crefname{theorem}{Verified Design Invariant}{Verified Design Invariants}
\Crefname{theorem}{Verified Design Invariant}{Verified Design Invariants}

\newcommand{\slm}{\textsc{SLM~4.0}}
\newcommand{\code}[1]{\texttt{#1}}

\hypersetup{
    colorlinks=true,
    linkcolor=Blue,
    citecolor=Green,
    urlcolor=Maroon,
    pdfauthor={Varun Pratap Bhardwaj, Garima Singh, Arun Pratap Bhardwaj},
    pdftitle={SuperLocalMemory 4.0: The Governed Memory Operating System for AI Agents},
}

\title{\bfseries SuperLocalMemory 4.0:\\[2pt]
       The Governed Memory Operating System for AI Agents\\[8pt]
       {\normalfont\large Unifying Information-Geometric Retrieval, Bi-Temporal
       Recall,\\[1pt]
       Verifiable Transactions, and Multi-Tenant Governance in One Local-First
       Control Plane\\[1pt]
       ---and Two Invariants That Ask Whether a Wired Mechanism Takes Effect}}

\author{%
  \normalsize
  Varun Pratap Bhardwaj\textsuperscript{1}\quad
  Garima Singh\textsuperscript{2}\quad
  Arun Pratap Bhardwaj\textsuperscript{2}
  \\[6pt]
  \small\textsuperscript{1}Qualixar / Independent Researcher, India \quad
  \textsuperscript{2}Independent Researcher, India
  \\[4pt]
  \small\texttt{varun.pratap.bhardwaj@gmail.com} \\[1pt]
  \texttt{garima1213@gmail.com} \quad
  \texttt{arun.pratap.bhardwaj@gmail.com}
  \\[3pt]
  \small ORCID (Varun Pratap Bhardwaj): 0009-0002-8726-4289
}

\date{}

\begin{document}

\maketitle

\begin{abstract}
\noindent\emph{In plain terms.} SuperLocalMemory is an open, local-first memory
system for AI agents. It lets agents remember across sessions, machines, and
people; it can run entirely on hardware you control, with no cloud provider in
the memory path; and a team or a company can operate it without giving up
isolation between one person's context and another's. What distinguishes it is
less the feature list than the evidence standard applied to it: every reliability
guarantee is stated as a falsifiable invariant, tested under an adversarial
condition, bracketed so that it fails if the mechanism does nothing, positive-controlled
where the design admits one, and reported together with an explicit account of what
the experiment did \emph{not} exercise. We call that practice
\textbf{AI Reliability Engineering}, and this paper is an attempt to hold a memory
system to it rather than to report capability numbers alone.

This version applies that standard to our own published claims and retracts two of
them---a learning layer we called governed, and a governance-cost figure that
measured something other than its label. Retracting a number is the standard
working, not the standard failing. A governed write costs $17.2$\,ms at the median on a
freshly created store.

\medskip
\noindent\textbf{The finding this version leads with.} Ten mechanisms in
this system were built correctly, wired into a live call path, and left inert or
unobservable at their final connection---among them a monitor inferring liveness
from the output it gates, an availability check that refuses a working store, and a
published latency figure that subtracted two write paths which were never built to be
compared, so the number it produced was never an overhead. Each was silent:
nothing raised, nothing logged, every file present and every version correct. The
common shape is that \emph{implemented}, \emph{reachable} and \emph{effective} are
three different questions, and that a signal derived from the mechanism it evaluates
cannot detect that mechanism's failure. We contribute two mechanical
invariants---one asserting that a Bayesian posterior's mean has moved off its prior
value after $n$ observations, one asserting whether a schema-guarded path can execute
against this store at all, and naming where its missing data actually lives when it
cannot---that answer the third question
directly. Pointed at our own store they returned \textsc{stalled} on every arm of
two learners, and acting on that verdict named the cause: two individually correct
components disagreed on an identifier namespace, so no engagement signal could ever
reach the learner, while a neutral-by-default settlement shrank posterior variance
and made the learner steadily harder to move. \textbf{Neutral is a commitment, not
an abstention.} The invariants ship in 4.1.5 and the repairs in 4.1.6--4.1.9. We
then test the repair rather than assert it: a three-arm ablation varying only the
identifier namespace, with every production module in the path held identical, moves
\textbf{no posterior in 120 recalls} with the defect present and \textbf{every
instantiated arm} with it absent, while a negative control that writes every outcome
ticket but supplies no engagement correctly settles nothing. The loop closes, and we
are explicit that a controlled ablation is not a deployment result.

\medskip
\noindent The remainder of this abstract is technical.

\medskip
AI agents are moving from single-user assistants to shared infrastructure---teams,
companies, and fleets of autonomous workers that must remember across sessions,
machines, and people without leaking one context into another, and, for regulated
or data-resident organizations, without a cloud provider in the memory path. Existing
work answers this in \emph{fragments}---memory SDKs and managed platforms, temporal
context graphs, stateful agent runtimes, framework-native primitives, context APIs, and
user-timeline stores---each strong on one axis. Few open, local-first runtimes unify
retrieval quality, a learning brain, time-awareness, a knowledge graph, verifiable
governance, and multi-tenant isolation in a single system an organization can operate.

We present \textbf{SuperLocalMemory~4.0 (\slm{})}, a governed, local-first
\emph{memory operating system} for AI agents: one control plane, organized as seven
operating layers over six managed SQLite stores plus vector and graph projections,
reachable through a CLI, an MCP server, an HTTP daemon, a dashboard, an editor plugin,
and separately packaged adapters for nine agent frameworks, in three operating modes (fully local; local with an
on-device model; provider-assisted). \slm{} unifies, in one runtime:
(i) \textbf{multi-channel retrieval}---dense semantic, BM25 lexical, temporal,
Hopfield-associative, and spreading-activation candidates fused by reciprocal-rank
fusion, over an information-geometric scoring and lifecycle substrate;
(ii) a \textbf{learning and behavior layer that is instrumented, measured, and
reported inert}---feedback capture, reward routing, a Thompson-sampling channel
selector, a learning-to-rank stage, behavioral soft-prompt injection, and a
blind-verified, budgeted skill-evolution pipeline that bounds self-modification.
\textbf{We do not claim this layer learns; we measured that it does not.} Across
three independent stores its Thompson posteriors satisfy
$\alpha-\alpha_0 = \beta-\beta_0 = n/2$ \emph{exactly} on \textbf{459 of 459 arms}
after \textbf{5{,}657} plays; a second Beta learner shows the identical signature;
and trust-weighted forgetting, which our prior work reported as delivered, is called
for every fact on the hot path and is \emph{arithmetically inert}, its guard testing
for a column absent from the table it tests while that column sits populated on
4{,}340 rows of a neighbouring one (\Cref{sec:learning-layer});
(iii) \textbf{SLM-Mesh}, serverless per-tenant-isolated coordination through a local
SQLite broker---authenticated peer messages, shared state, and advisory locks---%
extended (opt-in) across machines with deterministic convergence and fencing-token
lock ordering (\emph{scope}: coordination, \emph{not} linearizable consensus, quorum,
or a conflict-free replicated database);
(iv) \textbf{time-aware memory}---a three-date bi-temporal model, a dedicated temporal
retrieval channel, superseded-fact demotion, an Ebbinghaus recency model, and
cross-surface \code{as\_of} recall; and
(v) \textbf{governance-native memory}---multi-scope tenant isolation, role-based access,
a login gate, GDPR export/erasure with tamper-evident receipts, a hash-chained audit
trail, a deployment-context EU~AI~Act checklist, and \emph{operable recovery}---all in a
local-first runtime.
These capabilities become trustworthy under one governing design goal---one authenticated
actor, one profile generation, one policy decision, one durable receipt, one verifiable
completion state---implemented by four reliability mechanisms: generation-fenced write
admission, verifiable memory transactions (a transactional obligation ledger with
per-projection apply/verify/compensate/erase owners and a hash-checkable completion
manifest), cross-store verified erasure over those owners, and bi-temporal
\code{as\_of} demotion. V4 enforces the invariant on the primary write path; other
transports adopt the same gateway incrementally, so we present it as the enforced design
of that path and the target architecture for the platform, not as a property already
enforced on every surface.

We evaluate the reliability spine by direct measurement against the candidate artifact:
eleven fault-injection and mechanism scenarios, each repeated 200 times as a deterministic
flakiness check. \textbf{2{,}199 of 2{,}200 repetitions upheld their scoped component
property}, and the one that did not is reported rather than re-run: the generation-fence
experiment held 199 of 200 because the writer was transiently unavailable, not because
the fence misjudged an epoch, and diagnosing it exposed one more instance of the same
failure class (these repetitions are not independent Bernoulli trials and support no
population failure-rate inference; no single experiment exercises the full end-to-end
path---transport, multi-process, mesh, and long-lived deployment fault-injection are
future work). We report real-scale performance on a 1{,}232\,MiB memory-store copy. \textbf{We withdraw the governed write-envelope
\emph{overhead} figure from the previous version.} The two arms differ in ways other than
governance, and this version withdraws \emph{two} separate accounts of which way
mattered---both checkable, both wrong. The difference is negative at the median, which no
genuine overhead can be. We report the three paths as absolute costs and
then answer the question properly by abandoning subtraction: timing the envelope's
components where they run gives a governed write of $10.43$\,ms p50 of which the
envelope is $70.3\%$---and \textbf{almost all of that is durability, not governance}.
The generation fence costs $1.9\,\mu$s and the obligation ledger $42\,\mu$s, together
four tenths of one per cent; the rest is an encrypted write-ahead journal. We carry
the published LoCoMo evidence unchanged, at its stated coverage (protocol-scoped, \emph{not}
a new V4 benchmark). Several of these capabilities exist individually in prior and commercial
art; our contribution is their \emph{integration} into one local-first, verifiable, governed
runtime, and the central systems contribution is a hash-verifiable projection-obligation and
completion-manifest protocol across heterogeneous local stores under scoped admission.
\end{abstract}

\section{Introduction}
\label{sec:introduction}

The last two years produced a rich literature on agent memory. Systems have advanced
extraction and consolidation~\citep{mem0paper}, temporal knowledge
graphs~\citep{graphitipaper}, neurobiologically inspired retrieval~\citep{hipporag,hipporag2},
OS-style hierarchical tiers~\citep{memgpt,memoryos}, memory as an OS-managed
resource~\citep{memos}, multi-type and multi-agent coordination~\citep{mirix}, efficient
tiering with offline consolidation~\citep{lightmem}, strategic forgetting~\citep{sfams},
and trust-gated retrieval~\citep{memgate}. Each improves one dimension of memory quality.

\textbf{From point solutions to a complete memory layer.} A team standing up durable
agent memory today still assembles it from specialized parts---each with its own
trust boundary, failure model, and audit story. What we did not find, in a dated
review of representative systems (\Cref{sec:related}), is a single runtime in which
retrieval quality, a learning brain, time-awareness, a knowledge graph, verifiable
transactions, and multi-tenant governance are the \emph{same} system, running on
infrastructure the organization controls. This paper contributes that runtime.

\textbf{Rent the LLM, own the memory.} Foundation models are converging into an
interchangeable, rented capability; what an organization actually accumulates and must
protect is its \emph{memory}---the context, decisions, entities, skills, and provenance
its agents build over time. That asset, and the governance over it, should be owned
and operated by the organization on hardware it controls. A complete memory layer is
therefore local-first \emph{by construction}: storage, retrieval, the learning brain,
governance, and optimization all run on the organization's own infrastructure, with
no provider required in the memory path (Modes~A and~B; Mode~C is provider-assisted).

\textbf{The operating questions.} A second class of questions becomes central the
moment agent memory becomes shared organizational infrastructure: Can one client's
memory surface in another's recall---on \emph{any} retrieval path? When a user
exercises the right to be forgotten, does the fact disappear from every registered
projection store, with evidence? When agents on several laptops collaborate, can they
share what they should without a cloud server? When an agent improves its own skills,
can the organization bound and audit that self-modification? These are the operating
conditions of team-scale, multi-surface agent memory. They are the province of
\textbf{AI Reliability Engineering}---the systematic application of reliability,
correctness, and governance principles to AI-augmented software---and \slm{} is
built for them.

\textbf{The governing invariant.} \slm{}'s governing design goal is a single
admission invariant---one authenticated actor, one profile generation, one policy
decision, one durable operation receipt, and one verifiable completion state across
the registered projection owners. V4 realises this invariant on the primary write path
(the HTTP \code{/remember} route and internal ingestion admission), where the operation
policy registry is evaluated and the generation fence applies; the other transports
(MCP, CLI, WebSocket) and operation kinds adopt the same gateway incrementally. We
present the invariant as the enforced design of the write path and the target
architecture for the platform---not as a property already enforced on every surface.

\textbf{Contributions.} \slm{} consolidates a line of work---trust and behavioral
foundations~\citep{slmv2}, an information-geometric retrieval substrate~\citep{slmv3prior},
and a living-brain learning and lifecycle model~\citep{slmv3}---into one production
system. On those foundations, V4 contributes and \emph{measures} the reliability spine:
generation-fenced admission, verifiable memory transactions with a hash-checkable
completion manifest, cross-store verified erasure, and a cross-surface operation
policy registry. \Cref{sec:contributions} states each contribution (C1--C8) with its
defensible claim and explicit scope limit.

\section{Related Work}
\label{sec:related}

We organize related work by the capability each system most advances and locate
\slm{}'s distinct contribution in each area.

\textbf{Extraction and retrieval quality.}
Mem0~\citep{mem0paper,mem0repo} pairs LLM-based extraction with consolidation and
reports strong LoCoMo results; its 2026 revision adds cross-memory entity linking
and temporal accuracy gains~\citep{mem0v3}. HippoRAG~\citep{hipporag,hipporag2}
models hippocampal indexing with Personalized PageRank.  Many retrieval-quality
systems treat tenant isolation as application metadata rather than a write-time
fenced invariant; in the corpus we reviewed, we did not find one that specifies
per-operation transactional consistency across multiple heterogeneous derived stores.
\slm{}'s hybrid retrieval applies established hybrid-retrieval principles; we do not
claim retrieval-architecture novelty.

\textbf{Temporal and knowledge-graph memory.}
Graphiti/Zep~\citep{graphitipaper,zepgraphitirepo} maintains bi-temporal edge validity
with automatic invalidation---temporal correctness at the \emph{storage} level.
\slm{} places time at the \emph{retrieval-fusion} level: a dedicated temporal
candidate channel, non-destructive demotion of superseded facts, and an
Ebbinghaus-parameterized recency model applied per fact after RRF fusion.
We do not claim end-to-end temporal accuracy superiority; Graphiti's bi-temporal
storage is established prior art, and recent extraction-level results~\citep{mem0v3}
raise that bar.

\textbf{OS-inspired memory.}
MemGPT/Letta~\citep{memgpt,lettarepo}, MemoryOS~\citep{memoryos}, and
MemOS~\citep{memos} apply the OS metaphor to hierarchical context tiers and
MemCube scheduling. We reuse the OS metaphor but apply it to authorization,
multi-store durability, and governance. We do not claim the OS framing itself as novel.

\textbf{Self-improvement and governed skill evolution.}
The memory literature addresses improvement at the level of memory
\emph{content}---profile personalization, note re-linking, and sleep-time
consolidation. We are not aware of a prior system (in the corpus reviewed through
2026-08-03) that governs the mutation of live agent \emph{skill} configurations under
a staged
\texttt{screen}\,$\to$\,\texttt{confirm}\,$\to$\,\texttt{mutate}\,%
\texttt{blind-verify}\,$\to$\,\texttt{persist} pipeline with candidate provenance,
budget controls, and hash-linked state-transition history.

\textbf{Multi-tenancy, governance, and local-first.}
Multi-tenant isolation, compliance controls, and local-first operation are well
represented---Supermemory~\citep{supermemory}, Mem0~\citep{mem0repo}, and
Zep/Graphiti~\citep{zepgraphitirepo} all provide strong posture on these axes
individually. Governed multi-tenant shared memory is itself concurrent work:
MemClaw~\citep{memclaw} provides scoped retrieval, temporal supersession, and
policy-governed propagation; a production governed-memory architecture has been
described~\citep{governedmemory}. \slm{}'s contribution is the \emph{integration}
into a single local-first runtime: a scope algebra enforced by profile-scoped
predicates across all retrieval channels, a write-time generation fence,
and erasure verified across the three registered projection owners---with
isolation and projection erasure \emph{measured}.

\textbf{Concurrent transactional memory work.}
Two July-2026 preprints address transactional agent memory at distinct layers.
MemTX~\citep{memtx} stages writes in snapshot-isolated belief transactions with
cascading repair---correctness at the \textbf{semantic/belief layer}.
MemTxn~\citep{memtxn} defines a transaction boundary for source-supported updates
with complete-state recovery---correctness at the \textbf{logical-state layer}.
\slm{}'s transactions center on a \textbf{distinct layer}: \textbf{physical-store
projection consistency}---were all registered projection owners updated and
independently verified, under one hash-checkable completion manifest with an honest
\textsc{degraded} state? These layers compose: \slm{}'s manifest could sit beneath a
MemTX belief transaction. We claim no priority over this concurrent work.

\textbf{Self-organizing memory operating systems.}
EverMemOS~\citep{evermemos} presents a \emph{self-organizing memory operating
system}: dialogue is converted into memory cells carrying episodic traces, atomic
facts and time-bounded foresight signals, consolidated into thematic scenes, and
retrieved by scene-guided agentic recollection. The two systems self-organize different
things. EverMemOS self-organizes
\emph{content}: it decides what to keep, how to group it, and what to reconstruct,
and it is evaluated on reasoning quality. \slm{} governs \emph{operations}: which
writes are admitted, whether every projection of an admitted write was applied and
verified, what may be erased and provably was, and---in this version---what is
permitted to change the agent's behaviour. A system could adopt both, and the
comparison is not competitive: we report no retrieval-quality result for V4. It
shares vocabulary with this work---atomic facts, foresight, scenes---and the
operating-system framing that MemOS~\citep{memos} established before either of us.
MindMemOS~\citep{mindmemos} and AgentMemBench~\citep{agentmembench} appeared after
the first version of this paper; the former on portable self-evolving memory
layers, the latter benchmarking memory-management strategies.

\textbf{Governed memory as a framework.}
Two lines of work use the term. A production governed-memory architecture for
multi-agent workflows has been described~\citep{governedmemory}, and SSGM~\citep{ssgm}
proposes a conceptual governance framework for evolving agent memory built on
consistency verification, temporal decay modelling, and access control applied
before consolidation. Both are close in intent to this work, and SSGM's decay and
pre-consolidation-control mechanisms are conceptually adjacent to the
trust-modulated retention and write admission described here. The difference we
claim is not the concept but its state: the mechanisms in this paper are
implemented, exercised, and---where they do not work---\emph{measured and reported
as not working} (\Cref{sec:learning-layer}).

\textbf{Diagnosing memory failure.}
MemFail~\citep{memfail} is the nearest prior art to \Cref{sec:eval:taxonomy}. It
formalises a memory system as the composition of summarization, storage and
retrieval, derives the failure modes each operation admits, and builds five
adversarially designed datasets to attribute a wrong answer to a specific
operation across four systems treated as black boxes. It is a diagnostic benchmark
applied from outside, and it is the right instrument for the question it asks.

Two further works bound the same space and must be named. \citet{silentnarratives}
is the closest of all: a longitudinal taxonomy of silent failures in a production
LLM agent runtime, eight weeks of incidents with postmortems, five mechanism
classes, and the finding that most were caught by human observation rather than by
tests or audits. Its organising pattern is \emph{an error signal that never reaches
a human in actionable form}---an error occurs and is swallowed, diluted, or, in its
most striking class, narrated into fluent prose by the model itself.
\citet{memcircuits} attacks the same word from the opposite end, tracing feature
circuits inside the model to localise a silent memory failure to the responsible
stage.

\paragraph{The systems lineage we sit in.} Before positioning our class we should
say plainly where the idea comes from, because it is not ours.
\citet{endtoend} established forty years ago that a guarantee cannot be
established by a lower layer alone; it has to be checked at the level where it is
supposed to hold. \citet{grayfailure} named \emph{differential observability}: a
system's own failure detectors report health while the clients that depend on it
observe failure, which is the same asymmetry we describe when a learner's counters
rise while its posterior means do not move. \citet{panorama} answered it by turning
callers into observers of the components they call---an independent observer, placed
outside the failing component---and detected every one of fifteen reproduced gray
failures. \citet{oathkeeper} pushed further into our territory, inferring semantic
rules from a system's own regression tests and enforcing them at runtime across
ZooKeeper, HDFS and Kafka to catch violations that raise nothing;
\citet{flycatcher} continues that line with learned checker inference.

So the principle is old and well developed: \textbf{effectiveness is an end-to-end
semantic property, and establishing it requires an oracle that is independent of the
mechanism under test.} We claim no part of that. What we contribute is its
application to a place this literature has not looked---the learning and
schema-guarded paths of an agent memory system---and two concrete checkers for that
setting. It is also worth saying what is different about our setting: the
distributed-systems work above detects failures that some client eventually
experiences, whereas a memory system's learner can be inert for months with no client
anywhere observing anything wrong, because a uniformly-selecting retriever still
returns plausible memories. There is no downstream observer to recruit. The oracle
has to be constructed from the mechanism's own persisted state, and constructed so
that it does not depend on the mechanism's own account of itself.

Our class sits between the two agent-memory works above and overlaps neither.
\textbf{In our silent instances there is no error event at all.} A Thompson posterior receiving a valid neutral reward
forever has nothing to swallow; a schema guard returning \code{False} is behaving
exactly as designed; a latency delta between two arms that were never
comparable is a correct subtraction of the wrong quantities. There is no exception to lose, no log line
suppressed, and no stage to localise---the computation runs, returns, and means
nothing. Where \citet{silentnarratives} asks why a failure's signal did not reach a
human, we ask why there was no signal to begin with, and where
\citet{memcircuits} instruments the model, we query the memory system's own
persisted state. The three are complementary, and a deployment would want all
three.

Our failures are equally not on MemFail's axis, which is why we cite it as a
boundary rather than as a baseline. None of summarization, storage or retrieval is where a
learner that recorded five thousand plays and moved no posterior fails; nor a
conditional path that never executed because its guard reads a neighbouring
table's column; nor a liveness probe that infers health from the output it gates.
These are failures of \emph{wiring}, not of answer quality, and they are invisible
to a black-box benchmark because the system returns a plausible answer throughout.
Attributing them requires operating the system and querying its stores, which is
what \Cref{sec:eval:taxonomy} reports.

The second of our two invariants has a lineage we should name. Asserting that a
conditional path has actually executed in production is well established in
software engineering as dead-conditional and flag-controlled dead-code analysis,
where a branch guarded by a long-pinned flag is syntactically reachable, appears in
coverage, and has been dead for months. We claim no novelty for that idea. What we
add is narrow: applying it to \emph{schema-presence} guards in a memory system's
learning path, where the flag is the existence of a database column, the
consequence is a Bayesian learner that is arithmetically inert rather than merely
switched off, and the check reports where the required data actually resides.

\section{Platform at a Glance}
\label{sec:platform}

\slm{} is one control plane organized as seven operating layers, exposing a full
arsenal of capabilities over a single managed store set. This section gives one
honest paragraph per capability and the cross-system capability coverage map.

\subsection{Seven Operating Layers}

\begin{enumerate}[topsep=4pt, itemsep=2pt, label=\textbf{L\arabic*.}]
  \item \textbf{Admission.} Identity resolution, scope enforcement, idempotency
    check, and raw-evidence durability. A canonical write is admitted only after the
    \code{ActorContext} is derived from server-side state and the generation fence
    confirms the idempotency key is not replayed from a superseded profile-binding epoch.
  \item \textbf{Queryable Core.} An atomic SQLite transaction writes the canonical
    fact row, the operation receipt, and the transactional obligation ledger entries
    for every required projection owner---all or none.
  \item \textbf{Enrichment.} Fact extraction, entity resolution, scene construction,
    temporal provenance derivation, graph edge writing, and optional embedding
    (mode-dependent).
  \item \textbf{Memory Brain.} Behavioral pattern accumulation, feedback and reward
    recording, consolidation, soft-prompt generation, and guarded skill-evolution
    workflows.
  \item \textbf{Multi-channel Retrieval.} Five candidate producers followed by RRF
    fusion, optional reranking, and entity-graph score enhancement.
  \item \textbf{Context Safety.} Policy evaluation, provenance attachment, redaction
    of recognized secrets, boundary-marker neutralization, and context budget
    enforcement.
  \item \textbf{Operations.} Lifecycle management, hash-chained audit, exact-match
    cache and reversible compression, Mesh peer-coordination, backup/restore, and
    daemon health surfaces.
\end{enumerate}

\subsection{Capability Inventory}

\paragraph{Multi-channel retrieval.}
Five parallel candidate producers---dense semantic (with variance-weighted
re-scoring), BM25
lexical, temporal (bi-temporal validity filter), Hopfield-associative, and
spreading-activation---feed a single-pass Weighted RRF step ($k=15$) with
strategy-adaptive channel weights. An optional cross-encoder reranker runs
post-fusion; the entity-graph channel scores the fused candidate set but does not
generate independent candidates. Healthy channels participate; degraded channels are
excluded with provenance recorded in the response. \textbf{What ships is the simplified mode, not the geodesic.} The configuration
default selects a variance-weighted Mahalanobis-style re-scoring; the closed-form
geodesic on the Gaussian manifold from our prior work is implemented but is not the
default path, so a reader inspecting the repository will find the simplified form in
production. The contribution that survives is the one that is load-bearing: facts
whose observed embeddings have narrower variance---better-confirmed facts---earn a
scoring advantage proportional to inverse variance. Calling that a geodesic
overstates what executes, and we do not.

Variance-weighted scoring is wired
(\code{retrieval/semantic\_channel.py}, \code{core/engine\_wiring.py}).

\paragraph{Information-geometric substrate.}
The semantic channel applies variance-weighted re-scoring when per-memory variance
data is available, falling back to cosine similarity otherwise. Sheaf-cohomology
consistency checks and the Riemannian--Langevin lifecycle model were introduced in
prior work~\citep{slmv3prior,slmv3}; the \code{EbbinghausLangevinCoupling} class is
experimental and has no production caller in V4.

\paragraph{Entity knowledge graph.}
An AST-based code knowledge graph (blast-radius analysis, community detection,
execution-flow tracing, incremental indexing with git-hook and file-watcher updates)
and an entity store with typed edges and traversal API run inside the managed store.
The spreading-activation channel uses this graph to surface multi-hop associations.

\paragraph{Bi-temporal time-awareness.}
Each fact carries a three-date bi-temporal model (referenced / observation / interval).
The \code{as\_of} parameter is threaded across all primary recall surfaces---Python
recall engine, HTTP \code{/recall}, MCP tool, CLI \code{--as-of}---and drives a
temporal validity filter: facts are demoted in ranking under half-open event-validity
semantics and transaction-time supersession (\emph{point-in-time demotion},
not snapshot deletion). References are UTC-normalized.

\paragraph{Learning and behavior brain.}
A zero-LLM behavioral pattern miner captures tool-sequence patterns, correction events,
temporal clusters, and cross-project assertions, which a soft-prompt generator distills
into a compact context prefix prepended during session-context and auto-recall assembly
(ordinary \code{recall()} ranking is unchanged; initialization failure disables injection
fail-soft). Feedback loops (thumbs-up/down,
explicit reward signals), outcome recording, a LightGBM ensemble reranker, and a
cognitive consolidation queue are all wired and active by default.

\paragraph{Governed skill evolution.}
\slm{} governs mutation of live agent skill configurations under a
\texttt{screen\,$\to$\,confirm\,$\to$\,mutate\,$\to$\,blind-verify\,$\to$\,persist}
pipeline with candidate provenance, per-cycle budget controls (30-min wall / 10 LLM
calls / 3 cycles-per-day, fail-closed on exhaustion), and DB-enforced hash-linked
status transitions. \emph{Scope}: the default configuration stops at verified
quarantine; an explicit \code{auto\_approve} flag---a configuration gate, not human
or RBAC approval---can atomically promote a verified skill into the live loader with
rollback on failure. Faithful parent-version lineage and generation tracking are not
yet reported.

\paragraph{SLM-Mesh coordination.}
SLM-Mesh provides serverless, per-tenant-isolated coordination through a local SQLite
broker: authenticated peer messages, shared key/value state, and advisory locks with
TTL-based expiry and fencing tokens---available in all three operating modes (zero-cloud
Mode~A through cloud Mode~C). Cross-machine state/lock synchronization (opt-in)
extends this to LAN peers: leaderless, pull-based last-writer-wins state convergence
under $(\mathit{revision}, \mathit{node\_id})$ and advisory-lock view convergence
under fencing-token order, with a durable retrying outbox, optional mDNS discovery,
and TLS/pinning/HMAC-signed transport. \emph{Scope}: not linearizable consensus, not
quorum replication, not a CRDT database; ordinary storage writes do not invoke the
fence primitive; cross-machine coverage is currently the default profile, validated
on a two-node loopback.

\paragraph{Unified context optimization.}
An exact-match cache with tagged invalidation and a reversible compression stage
(byte-exact recovery of originals) are co-located in the memory control plane and
exposed on three surfaces (proxy, MCP, and skill), with savings telemetry.

\paragraph{Governance-native controls.}
A governance control plane runs locally and in-process (no network call): a
server-derived immutable \code{ActorContext}; an 18-kind declarative operation policy
registry; role-based access (OWNER / ADMIN / MEMBER / VIEWER lattice); multi-scope
tenant isolation (personal / shared / global) enforced at write time; GDPR
export/erasure with tamper-evident receipts; a hash-chained audit trail; and a
deployment-context EU~AI~Act checklist that abstains from legal classification
without intended-use evidence. Details appear in \Cref{sec:governance}.

\paragraph{Operational recovery and remediation.}
Auto-reconciliation redrives failed projection obligations and dead-letters exhausted
ingestions; a write-path stall watchdog fast-fails new writers (rather than freezing all
tenants) when the single writer stalls; and an OWNER/ADMIN remediation surface---one-click on
the dashboard, plus CLI (\code{slm ops}) and MCP---lists and resolves (retry / force-reconcile
/ cancel) stuck, dead-lettered, or \textsc{degraded} operations, with failure counts surfaced
on \code{/health}. Detailed in \Cref{sec:spine:recovery}.

\subsection{Capability Coverage Matrix}
\label{sec:platform:matrix}

\Cref{tab:capability} maps fifteen capability axes across \slm{} and six
representative systems, showing where coverage concentrates and where \slm{}'s
integration adds axes not individually claimed as novel. Cells are scored from each
system's \emph{public documentation} as of 2026-08-03, not from private testing: a
dash (---) means a capability was \emph{not found in public sources}, which is \emph{not}
proof of absence---a system may support more than its docs describe. \slm{}'s own
\checkmark\ cells are the ones backed by the code and tests in this paper.

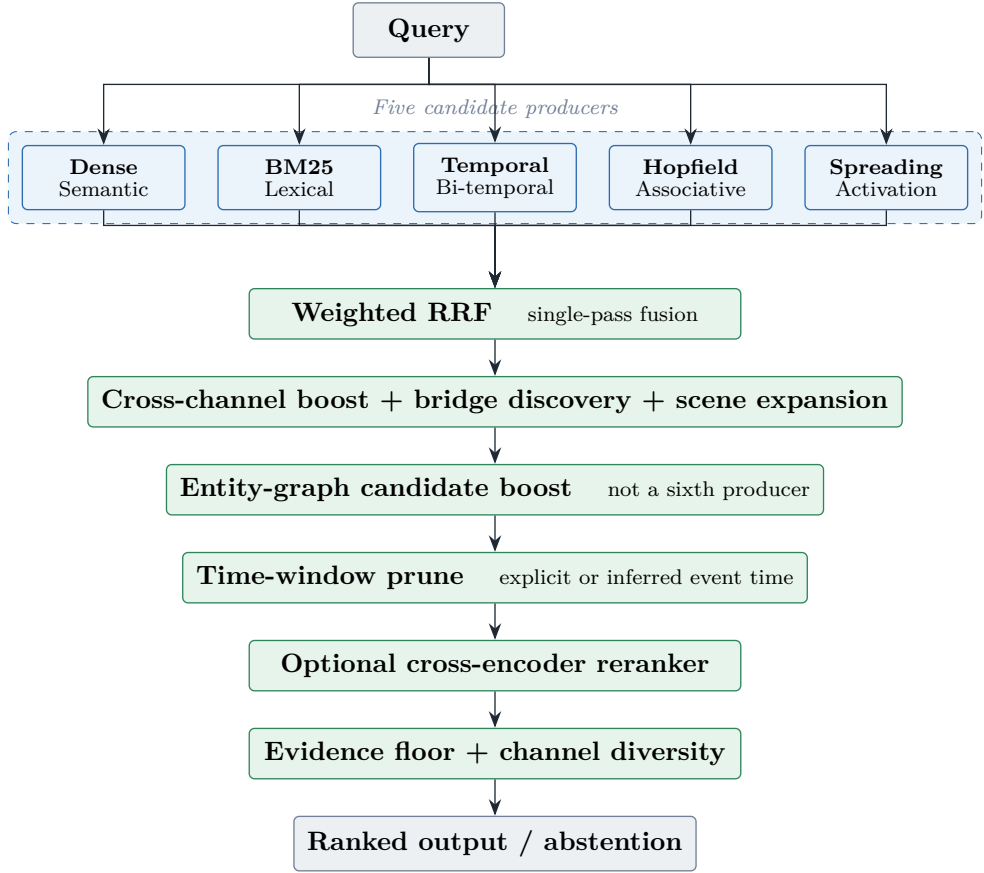
\begin{figure}[htbp]
  \centering
  \resizebox{\ifdim\width>\linewidth\linewidth\else\width\fi}{!}{%
\begin{tikzpicture}[
      node distance=0.48cm and 0.42cm,
      chan/.style={slmbox, fill=slmbluebg, draw=slmblue,
                   minimum width=2.15cm, minimum height=0.78cm, font=\scriptsize},
      proc/.style={slmbox, fill=slmgreenbg, draw=slmgreen,
                   minimum width=6.5cm, minimum height=0.66cm, font=\small},
      qbox/.style={slmbox, fill=slmgraybg, draw=slmgray,
                   minimum width=2.0cm, minimum height=0.72cm, font=\small},
    ]

    \node[qbox] (query) {\textbf{Query}};
    \node[chan, below=1.15cm of query, xshift=-4.3cm] (sem)
      {\textbf{Dense}\\[-1pt]Semantic};
    \node[chan, right=of sem] (bm25) {\textbf{BM25}\\[-1pt]Lexical};
    \node[chan, right=of bm25] (temp) {\textbf{Temporal}\\[-1pt]Bi-temporal};
    \node[chan, right=of temp] (hopf) {\textbf{Hopfield}\\[-1pt]Associative};
    \node[chan, right=of hopf] (sa) {\textbf{Spreading}\\[-1pt]Activation};

    \begin{scope}[on background layer]
      \node[draw=slmblue, fill=slmbluebg, dashed, rounded corners=3pt,
            fit=(sem)(sa), inner sep=5pt,
            label={[slmlabel, yshift=1pt]above:\textit{Five candidate producers}}] {};
    \end{scope}

    \node[proc, below=1.0cm of temp] (rrf)
      {\textbf{Weighted RRF} \quad {\scriptsize single-pass fusion}};
    \node[proc, below=of rrf] (expand)
      {\textbf{Cross-channel boost + bridge discovery + scene expansion}};
    \node[proc, below=of expand] (entity)
      {\textbf{Entity-graph candidate boost} \quad {\scriptsize not a sixth producer}};
    \node[proc, below=of entity] (window)
      {\textbf{Time-window prune} \quad {\scriptsize explicit or inferred event time}};
    \node[proc, below=of window] (rerank)
      {\textbf{Optional cross-encoder reranker}};
    \node[proc, below=of rerank] (qualify)
      {\textbf{Evidence floor + channel diversity}};
    \node[qbox, below=of qualify] (out) {\textbf{Ranked output / abstention}};

    \foreach \ch in {sem,bm25,temp,hopf,sa} {
      \draw[slmflow] (query.south) -- ++(0,-0.35) -| (\ch.north);
      \draw[slmflow] (\ch.south) -- ++(0,-0.2) -| (rrf.north);
    }
    \draw[slmflow] (rrf) -- (expand);
    \draw[slmflow] (expand) -- (entity);
    \draw[slmflow] (entity) -- (window);
    \draw[slmflow] (window) -- (rerank);
    \draw[slmflow] (rerank) -- (qualify);
    \draw[slmflow] (qualify) -- (out);

  \end{tikzpicture}}
  \caption{Implemented retrieval order.  Five parallel producers feed weighted
    reciprocal-rank fusion.  Cross-channel intersection boosting, code-memory
    bridge discovery, and scene expansion enlarge or adjust the fused set before
    entity-graph scoring boosts existing candidates.  Event-time pruning precedes
    the optional cross-encoder.  Finally, the evidence floor removes candidates
    without qualifying primary evidence and channel diversity is enforced before
    returning ranked results (or an abstention).}
  \label{fig:retrieval_pipeline}
\end{figure}

\begin{figure}[htbp]
  \centering
  \resizebox{\ifdim\width>\linewidth\linewidth\else\width\fi}{!}{%
\begin{tikzpicture}[
      node distance=0.65cm and 1.0cm,
      trigger/.style={slmbox, fill=slmamberbg, draw=slmamber,
                      minimum width=2.6cm, minimum height=0.85cm, font=\small},
      stage/.style={slmbox, fill=slmbluebg, draw=slmblue,
                    minimum width=2.8cm, minimum height=0.95cm, font=\small},
      gate/.style={slmbox, fill=slmgreenbg, draw=slmgreen,
                   minimum width=2.8cm, minimum height=0.85cm, font=\small},
      reject/.style={slmbox, fill=slmredbg, draw=slmred,
                     minimum width=2.2cm, minimum height=0.75cm, font=\scriptsize},
      promote/.style={slmbox, fill=slmgreenbg, draw=slmgreen,
                      minimum width=2.6cm, minimum height=0.85cm, font=\small},
      budget/.style={slmbox, fill=slmgraybg, draw=slmgray,
                     minimum width=2.2cm, minimum height=0.7cm, font=\scriptsize},
    ]

    \node[trigger] (t1) {\textbf{T1:} Post-session\\[-1pt]{\scriptsize Stop hook}};
    \node[trigger, right=0.5cm of t1] (t2) {\textbf{T2:} Degradation\\[-1pt]{\scriptsize Consolidation}};
    \node[trigger, right=0.5cm of t2] (t3) {\textbf{T3:} Health-check\\[-1pt]{\scriptsize Every $N$-th cycle}};

    \begin{scope}[on background layer]
      \node[draw=slmamber, fill=slmamberbg, dashed, rounded corners=3pt,
            fit=(t1)(t3), inner sep=5pt,
            label={[slmlabel, yshift=1pt]above:\textit{Triggers}}] {};
    \end{scope}


    \node[stage, below=1.2cm of t2] (screen)
      {\textbf{1. Screen}\\[-1pt]{\scriptsize Anti-loop guards}\\[-1pt]
       {\scriptsize (hash dedup + attempt cap)}};

    \node[reject, right=1.4cm of screen] (r1)
      {Skip\\[-1pt]{\scriptsize (already addressed)}};

    \node[stage, below=0.7cm of screen] (confirm)
      {\textbf{2. Confirm}\\[-1pt]{\scriptsize configured/lowest-cost model}\\[-1pt]
       {\scriptsize Evidence warrants change?}};

    \node[reject, right=1.4cm of confirm] (r2)
      {Reject\\[-1pt]{\scriptsize (logged)}};

    \node[stage, below=0.7cm of confirm] (mutate)
      {\textbf{3. Mutate}\\[-1pt]{\scriptsize configured model}\\[-1pt]
       {\scriptsize Evolve skill content}};

    \node[reject, right=1.4cm of mutate] (r3)
      {Failed\\[-1pt]{\scriptsize (mutation empty)}};

    \node[gate, below=0.7cm of mutate] (verify)
      {\textbf{4. Blind Verify}\\[-1pt]{\scriptsize information-isolated; distinct verifier}\\[-1pt]
       {\scriptsize when available, else independence logged}};

    \node[reject, right=1.4cm of verify] (r4)
      {Reject\\[-1pt]{\scriptsize (verifier reasoning logged)}};

    \node[promote, below=0.7cm of verify] (persist)
      {\textbf{5. Persist}\\[-1pt]{\scriptsize Quarantine dir + audit record}\\[-1pt]
       {\scriptsize transition metadata: generation = parent + 1}};

    \begin{scope}[on background layer]
      \node[draw=slmgray, fill=slmgraybg, dashed, rounded corners=4pt,
            fit=(screen)(persist), inner sep=8pt,
            label={[slmlabel, yshift=1pt]above:\textit{Evolution budget: 30\,min\,/\,10\,LLM calls\,/\,3\,cycles per day}}] {};
    \end{scope}

    \foreach \t in {t1, t2, t3} {
      \draw[slmflow] (\t.south) -- ++(0,-0.45) -| (screen.north);
    }

    \draw[slmflow] (screen.south) -- (confirm.north);
    \draw[slmflow] (confirm.south) -- (mutate.north);
    \draw[slmflow] (mutate.south) -- (verify.north);
    \draw[slmflow] (verify.south) -- (persist.north);

    \draw[slmflow, draw=slmred] (screen.east) -- (r1.west);
    \draw[slmflow, draw=slmred] (confirm.east) -- (r2.west);
    \draw[slmflow, draw=slmred] (mutate.east) -- (r3.west);
    \draw[slmflow, draw=slmred] (verify.east) -- (r4.west);

    \node[slmlabel, above right=-2pt and 2pt of r1] {\textsc{hash match}};
    \node[slmlabel, above right=-2pt and 2pt of r2] {\textsc{not warranted}};
    \node[slmlabel, above right=-2pt and 2pt of r3] {\textsc{llm fail}};
    \node[slmlabel, above right=-2pt and 2pt of r4] {\textsc{verification failed}};

  \end{tikzpicture}}
  \caption{Governed skill evolution pipeline in \slm{} 4.0.
    Three trigger conditions (post-session Stop hook; degradation scan;
    health-check scan) initiate evaluation within a hard budget envelope.
    Five guarded stages run sequentially: anti-loop screening, confirmation by
    the configured/lowest-cost model, mutation by the configured model, blind
    verification (information-isolated; the verifier cannot see the generator's
    rationale or the original skill; a distinct verifier model is used when
    available, otherwise reduced independence is logged), and persistence
    to a quarantine directory with transition metadata setting the generation to
    the latest promoted parent generation plus one.  Red exits persist their
    state-machine transition; promotion audit logging occurs only after blind
    verification succeeds.
    Budget exhaustion aborts the cycle before any LLM call is charged beyond
    the cap.}
  \label{fig:skill_evolution}
\end{figure}
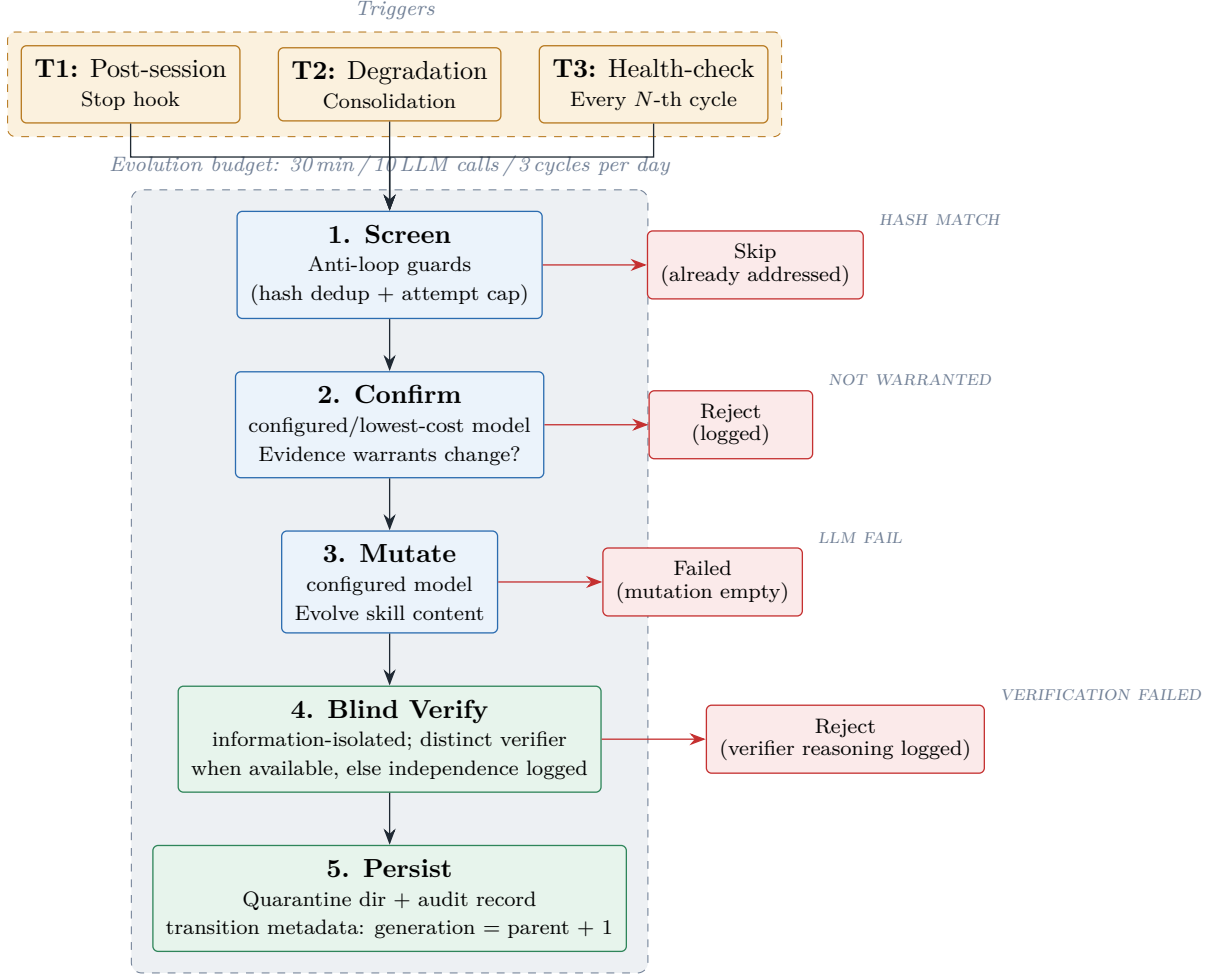

\begin{table*}[tp]
  \centering
  \caption{Capability coverage across representative agent-memory systems, scored from
           \emph{public documentation} as of 2026-08-03.
           \checkmark~=~documented/supported (for \slm{}, code-verified in this paper);
           $\circ$~=~partial or scoped in public docs;
           ---~=~not found in public sources (\emph{not} proof of absence). Each system
           excels on its target axis; this reflects public-documentation focus, not
           deficiency.}
  \label{tab:capability}
  \setlength{\tabcolsep}{5pt}
  {\small
  \begin{tabular}{@{}p{4.6cm}ccccccc@{}}
    \toprule
    \textbf{Capability axis}
      & \textbf{SLM~V4}
      & \textbf{Mem0}
      & \shortstack{\textbf{Zep/}\\\textbf{Graphiti}}
      & \textbf{Letta}
      & \shortstack{\textbf{Lang-}\\\textbf{Mem}}
      & \shortstack{\textbf{Super-}\\\textbf{memory}}
      & \shortstack{\textbf{Memo-}\\\textbf{base}} \\
    \midrule
    Local-first, no cloud in write path (Modes A/B)
      & \checkmark$^{*}$ & $\circ$ & $\circ$ & $\circ$ & $\circ$ & $\circ$ & $\circ$ \\[2pt]
    Multi-scope isolation + RBAC roles
      & $\circ$ & $\circ$ & $\circ$ & $\circ$ & ---     & $\circ$ & $\circ$ \\[2pt]
    GDPR erasure + hash audit + EU AI Act
      & $\circ$ & $\circ$ & $\circ$ & $\circ$ & ---     & $\circ$ & ---     \\[2pt]
    Bi-temporal / time-aware storage
      & \checkmark & $\circ$ & \checkmark & --- & ---   & ---     & $\circ$ \\[2pt]
    Multi-channel retrieval + RRF fusion
      & \checkmark & $\circ$ & $\circ$ & $\circ$ & ---   & $\circ$ & ---     \\[2pt]
    Knowledge graph + entities
      & \checkmark & $\circ$ & \checkmark & $\circ$ & --- & ---    & ---     \\[2pt]
    Learning + behavioral brain (soft prompts)
      & $\varnothing^{w}$ & ---  & ---     & ---     & ---     & ---     & ---     \\[2pt]
    Governed skill evolution (guarded pipeline)
      & $\varnothing^{y}$ & ---   & ---     & ---     & ---     & ---     & ---     \\[2pt]
    Verifiable transactions + manifest
      & \checkmark & ---     & ---     & ---     & ---     & ---     & ---     \\[2pt]
    Verified cross-store erasure receipts
      & $\circ^{x}$ & ---   & ---     & ---     & ---     & ---     & ---     \\[2pt]
    Exact cache + reversible compression
      & \checkmark & ---     & ---     & ---     & ---     & ---     & ---     \\[2pt]
    Trusted-peer mesh (serverless, isolated)
      & $\circ^{z}$ & ---   & ---     & ---     & ---     & ---     & ---     \\[2pt]
    Bounded agentic loops (independent gate)
      & \checkmark$^{\dagger}$ & --- & ---   & ---     & ---     & ---     & ---     \\[2pt]
    Operational recovery + admin remediation (dashboard/CLI)
      & \checkmark$^{\dagger}$ & --- & ---   & ---     & ---     & ---     & ---     \\[2pt]
    Framework adapters + multi-surface API
      & \checkmark & \checkmark & \checkmark & \checkmark & \checkmark & \checkmark & \checkmark \\
    \bottomrule
  \end{tabular}
  }
  \vspace{4pt}
  \begin{minipage}{\linewidth}
    {\footnotesize
    \textit{Notes.}
    Per-cell scoring rationale and primary-source citations are maintained in the full
    report's capability ledger; compound axes (e.g.\ GDPR erasure + hash audit + EU AI Act)
    are scored on their weakest conjunct.
    \textbf{SLM~V4 local-first (\checkmark$^{*}$):} local-first by default---full in
    Modes~A/B; Mode~C is opt-in cloud.
    \textbf{Evidence tier.} An unmarked \checkmark{} is verified by an experiment in
    \Cref{tab:results}. A \checkmark$^{\dagger}$ is verified by code inspection and
    unit tests only---present and reachable, with no fault-injection scenario behind
    it. A $\varnothing$ means implemented and \emph{measured ineffective}: the
    mechanism exists, is reached, and has not changed an outcome
    (\Cref{sec:learning-layer}). Distinguishing these three is the point of the
    paper, so the matrix should not flatten them.
    \textbf{($w$) Learning brain ($\varnothing$):} instrumented; posteriors at their
    priors across three stores, so no behaviour has been altered.
    \textbf{($y$) Skill evolution ($\varnothing$):} pipeline implemented and gated;
    zero cycles have completed, so there is nothing to score. Live-loader activation
    would be configuration-gated, not human- or RBAC-approved.
    \textbf{($x$) Erasure ($\circ$):} verified over the canonical store and the
    registered projection owners. Backup artifacts and any remote peer holding state
    are outside the completeness gate, and the audit chain is same-host tamper
    evidence rather than a remote attestation.
    \textbf{($z$) Mesh ($\circ$):} opt-in, default-profile leaderless LWW convergence
    with fencing-token advisory locks and TLS/pinned transport, two-node-loopback
    validated; not consensus/quorum/CRDT. All entries scored from public documentation
    reviewed through 2026-08-03; a~--- for a competitor reflects its public-doc scope,
    not a deficiency, and competitors may support more than their docs describe.
    }
  \end{minipage}
\end{table*}

\section{Platform Architecture}
\label{sec:architecture}

\slm{} is one control plane over many surfaces and stores, governed by the
admission invariant stated in \Cref{sec:introduction}. It is realised through an
immutable server-derived \code{ActorContext}, a declarative
\code{OperationPolicyRegistry}, and a transaction spine whose projection owners
each prove their own completion.

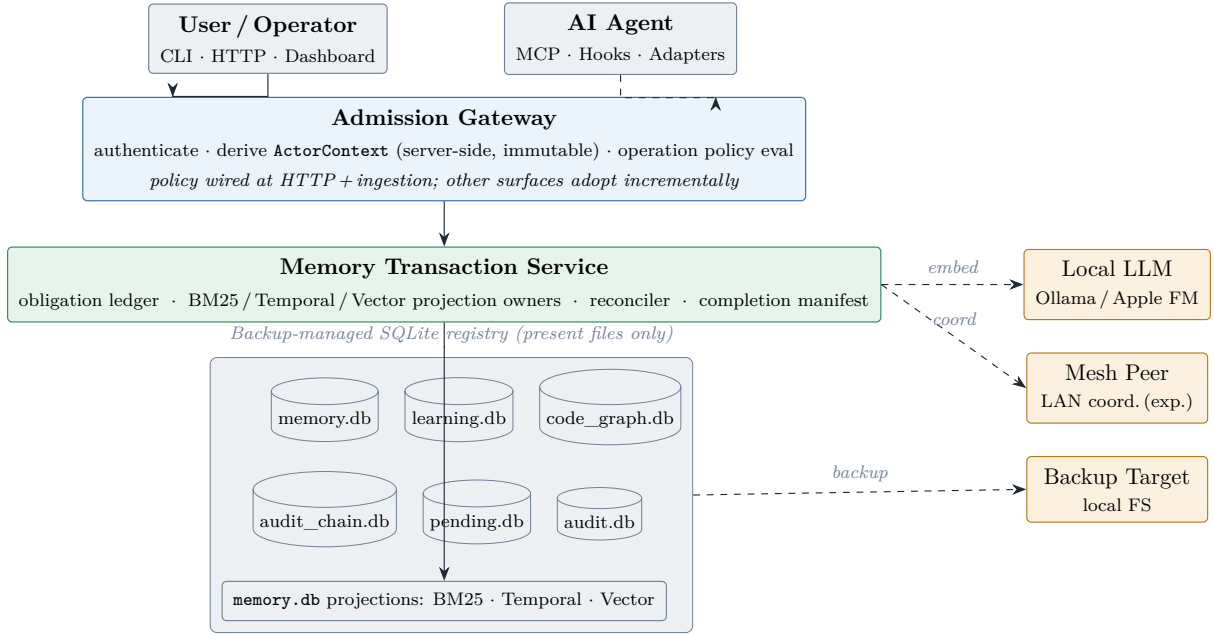
\begin{figure}[htbp]
  \centering
  \resizebox{\ifdim\width>\linewidth\linewidth\else\width\fi}{!}{%
\begin{tikzpicture}[
      node distance=0.75cm and 1.2cm,
      actor/.style={slmbox, fill=slmgraybg, draw=slmgray, minimum width=3.2cm,
                    minimum height=0.9cm},
      ext/.style={slmwarn, minimum width=2.8cm, minimum height=0.8cm},
    ]

    \node[actor] (user) {%
      \textbf{User\,/\,Operator}\\[1pt]
      {\scriptsize CLI \(\cdot\) HTTP \(\cdot\) Dashboard}};
    \node[actor, right=1.8cm of user] (agent) {%
      \textbf{AI Agent}\\[1pt]
      {\scriptsize MCP \(\cdot\) Hooks \(\cdot\) Adapters}};

    \coordinate (mid) at ($(user)!0.5!(agent)$);
    \node[slmgate, minimum width=7.2cm, below=0.9cm of mid] (gw) {%
      \textbf{Admission Gateway}\\[2pt]
      {\scriptsize authenticate \(\cdot\) derive \texttt{ActorContext}
        (server-side, immutable) \(\cdot\) operation policy eval}\\[1pt]
      {\scriptsize\itshape policy wired at HTTP\,+\,ingestion;
        other surfaces adopt incrementally}};

    \node[slmspine, minimum width=7.2cm, below=0.7cm of gw] (mts) {%
      \textbf{Memory Transaction Service}\\[2pt]
      {\scriptsize obligation ledger \(\;\cdot\;\)
        BM25\,/\,Temporal\,/\,Vector projection owners \(\;\cdot\;\)
        reconciler \(\;\cdot\;\) completion manifest}};

    \node[slmstore, below=0.85cm of mts, xshift=-1.85cm] (s1) {memory.db};
    \node[slmstore, right=0.4cm of s1]                   (s2) {learning.db};
    \node[slmstore, right=0.4cm of s2]                   (s3) {code\_graph.db};
    \node[slmstore, below=0.45cm of s1] (s4) {audit\_chain.db};
    \node[slmstore, right=0.4cm of s4]  (s5) {pending.db};
    \node[slmstore, right=0.4cm of s5]  (s6) {audit.db};

    \node[slmbox, draw=slmgray, fill=slmgraybg, font=\scriptsize,
          minimum width=5.2cm, minimum height=0.5cm,
          below=0.52cm of s4, xshift=1.85cm] (projections) {%
      \texttt{memory.db} projections: BM25 \(\cdot\) Temporal \(\cdot\) Vector};

    \begin{scope}[on background layer]
      \node[draw=slmgray, fill=slmgraybg, rounded corners=3pt,
            fit=(s1)(s4)(s3)(s6)(projections), inner sep=5pt,
            label={[slmlabel, yshift=2pt]above:%
              Backup-managed SQLite registry (present files only)}]
            (storegroup) {};
    \end{scope}

    \node[ext, right=2.2cm of mts] (llm) {%
      Local LLM\\{\scriptsize Ollama\,/\,Apple FM}};
    \node[ext, below=0.5cm of llm] (mesh) {%
      Mesh Peer\\{\scriptsize LAN coord.\,(exp.)}};
    \node[ext, below=0.5cm of mesh] (bkup) {%
      Backup Target\\{\scriptsize local FS}};

    \draw[slmflow] (user.south) -- ++(0,-0.35) -|
      ($(gw.north west)!0.25!(gw.north)$);
    \draw[slmflowdash] (agent.south) -- ++(0,-0.35) -|
      ($(gw.north east)!0.25!(gw.north)$);

    \draw[slmflow] (gw) -- (mts);

    \draw[slmflow] (mts.south) -- ++(0,-0.38) -| (projections.north);

    \draw[slmflowdash] (mts.east) -- (llm.west)
      node[slmlabel, midway, above=1pt] {embed};
    \draw[slmflowdash] (mts.east) -- (mesh.west)
      node[slmlabel, midway, above=1pt] {coord};
    \draw[slmflowdash] (storegroup.east) -- (bkup.west)
      node[slmlabel, midway, above=1pt] {backup};

  \end{tikzpicture}}
  \caption{SLM~4.0 system context.  The HTTP\,/\,internal-ingestion write path
    reaches the engine through the Admission Gateway (solid arrow from
    User\,/\,Operator); AI agents via MCP\,/\,Hooks\,/\,Adapters adopt the
    gateway policy incrementally (dashed arrow); CLI is likewise incremental.
    The gateway derives an immutable \texttt{ActorContext} from the authenticated
    session and evaluates the operation policy registry.  The Memory Transaction
    Service owns the obligation/reconciliation protocol, not the six database
    files.  The backup registry covers the present subset of
    \texttt{memory.db}, \texttt{learning.db}, \texttt{audit\_chain.db},
    \texttt{code\_graph.db}, \texttt{pending.db}, and \texttt{audit.db}.
    Its three registered owners (BM25, Temporal, Vector) maintain projections
    rooted in the canonical \texttt{memory.db}.  Dashed arrows indicate optional
    runtime connections (external systems) or incremental gateway adoption
    (AI agents).}
  \label{fig:context}
\end{figure}

\textbf{System context and boundaries.} External actors---human operators via
CLI, HTTP, or Dashboard and AI agents via MCP, hooks, or adapters---reach the
engine through the daemon bound on port 8765 (\code{server/unified\_daemon.py}).
The Admission Gateway derives an immutable \code{ActorContext} from the authenticated
session and evaluates the operation policy registry. In V4 this \emph{full} gateway
envelope is wired on the HTTP \code{/remember} write path and at the internal ingestion
admission slot; the other transport surfaces adopt it incrementally---declared MCP tools
and selected CLI mutations obtain a shared policy \emph{decision} but not yet the full
envelope, and direct legacy, Hook, and Adapter paths remain incremental. \Cref{fig:context}
is drawn from this canonical-path perspective.

\begin{figure}[htbp]
  \centering
  \resizebox{\ifdim\width>\linewidth\linewidth\else\width\fi}{!}{%
\begin{tikzpicture}[
      node distance=0.7cm and 0.9cm,
      tport/.style={slmbox, minimum width=2.5cm, minimum height=0.75cm,
                    fill=slmgraybg, draw=slmgray},
      fstep/.style={slmspine, minimum width=6.8cm, minimum height=0.8cm},
      owner/.style={slmspine, minimum width=1.9cm, minimum height=0.75cm},
      ownerX/.style={slmbox, minimum width=1.9cm, minimum height=0.75cm,
                     draw=slmgray, fill=slmgraybg},
    ]

    \node[tport] (http) {%
      HTTP / WebSocket\\{\scriptsize\itshape policy wired here}};
    \node[tport, below=0.45cm of http] (mcp) {MCP server};
    \node[tport, below=0.45cm of mcp] (cli) {CLI};
    \node[tport, below=0.45cm of cli] (hooks) {Hooks / Adapters};

    \node[slmlabel, above=0.3cm of http, xshift=-0.2cm] {Transport surfaces};

    \draw[decorate, decoration={brace, amplitude=4pt, mirror},
          draw=slmgray, line width=0.4pt]
      (mcp.south west) -- (hooks.south west)
      node[slmlabel, midway, left=6pt] {\rotatebox{90}{incremental}};


    \node[slmgate, minimum width=6.8cm, minimum height=4.5cm,
          right=1.6cm of http, yshift=-1.8cm] (gw) {%
      \textbf{Admission Gateway}\\[2pt]
      {\scriptsize derive \texttt{ActorContext} (server-side, frozen)
        \(\cdot\) assemble admission inputs}\\[1pt]
      {\scriptsize\itshape generation epoch captured here
        \(\rightarrow\) fence.record\_admission\_epoch}};

    \node[slmgate, minimum width=6.8cm, minimum height=0.9cm,
          below=0.65cm of gw] (pol) {%
      \textbf{Operation Policy Registry}\\[2pt]
      {\scriptsize keyed by \texttt{OperationKind} only \(\cdot\)
        auth checked before role eval \(\cdot\)
        fail-open (local) / fail-closed (company)}};

    \node[fstep, below=0.65cm of pol] (scrub) {%
      \textbf{1.\ Path-specific admission preparation}\\[1pt]
      {\scriptsize HTTP: trust/validation first, scrub after journal \(\cdot\)
        internal ingestion: scrub before admission}};

    \node[fstep, below=0.5cm of scrub] (journal) {%
      \textbf{2.\ \texttt{journal.prepare}}\\[1pt]
      {\scriptsize HTTP stores the encrypted original request before scrub}};

    \node[fstep, below=0.5cm of journal] (wc) {%
      \textbf{3.\ WriteCoordinator epoch check}\\[1pt]
      {\scriptsize compare captured epoch \(G\) vs \texttt{runtime.current\_generation}
        \(\;\Rightarrow\;\) stale: \texttt{WriteCoordinatorError} (no projection write)
        \(\;\cdot\;\) fresh: proceed}};

    \node[fstep, below=0.5cm of wc] (commit) {%
      \textbf{4.\ Atomic canonical commit}\\[1pt]
      {\scriptsize canonical row \(\cdot\) durable write receipt \(\cdot\)
        projection obligations --- one transaction, all or none}};

    \node[fstep, below=0.5cm of commit] (proj) {%
      \textbf{5.\ Reconcile projection obligations} (verify/apply/verify)};

    \node[owner, below=0.55cm of proj, xshift=-2.3cm] (bm25) {%
      Bm25Owner\\{\scriptsize lexical / FTS5}};
    \node[owner, right=0.4cm of bm25] (temp) {%
      TemporalOwner\\{\scriptsize validity}};
    \node[owner, right=0.4cm of temp] (vec) {%
      VectorOwner\\{\scriptsize embed.\ metadata}};

    \node[ownerX, right=0.8cm of vec, minimum width=2.5cm] (nonown) {%
      {\scriptsize graph / cache / mesh}\\[1pt]
      {\scriptsize\itshape not registered}\\
      {\scriptsize\itshape owners in V4}};

    \node[fstep, below=0.55cm of temp] (recon) {%
      \textbf{6.\ Reconciler} \(\rightarrow\)
      \textbf{CompletionManifest}\\[1pt]
      {\scriptsize HMAC-SHA256 v2 (M037; downgrade refused) \(\cdot\)
        \texttt{all\_met=True}: COMPLETE \(\cdot\)
        \texttt{all\_met=False}: DEGRADED (retried) \(\cdot\)
        canonical fail: FAILED}\\[1pt]
      {\scriptsize\itshape legacy: unkeyed SHA-256 v1 (self-consistency only)}};


    \draw[slmflow] (http.east) -- (gw.west|-http.east);
    \draw[slmflow] (mcp.east)  -- (gw.west|-mcp.east);
    \draw[slmflow] (cli.east)  -- (gw.west|-cli.east);
    \draw[slmflow] (hooks.east) -- (gw.west|-hooks.east);

    \draw[slmflow] (gw)    -- (pol);
    \draw[slmflow] (pol)   -- (scrub);
    \draw[slmflow] (scrub) -- (journal);
    \draw[slmflow] (journal) -- (wc);
    \draw[slmflow] (wc)    -- (commit);
    \draw[slmflow] (commit) -- (proj);

    \draw[slmflow] (proj.south) -- ++(0,-0.28) -|
      (bm25.north);
    \draw[slmflow] (proj.south) -- ++(0,-0.28) -|
      (temp.north);
    \draw[slmflow] (proj.south) -- ++(0,-0.28) -|
      (vec.north);

    \draw[slmflow] (bm25.south) -- (recon.north|-bm25.south);
    \draw[slmflow] (temp.south) -- (recon.north|-temp.south);
    \draw[slmflow] (vec.south)  -- (recon.north|-vec.south);

  \end{tikzpicture}}
  \caption{V4 transaction and policy spine.  Transport surfaces reach the
    Admission Gateway; in V4 the operation
    policy is evaluated on the HTTP write path and the internal ingestion admission
    slot --- other surfaces adopt incrementally.  The gateway derives an immutable
    \texttt{ActorContext} (server-side; caller-supplied fields are data, not
    authority) and captures the profile generation epoch in the process-local
    generation fence.  HTTP admission journals the encrypted original before
    the ingestion writer scrubs it; internal ingestion scrubs before admission.
    The WriteCoordinator checks the captured epoch before the atomic
    canonical-row, durable-receipt, and obligation-ledger commit.  The Operation Policy Registry is keyed by
    \texttt{OperationKind} only; authentication is checked before role evaluation.
    Three projection owners are registered in V4 (\texttt{Bm25Owner},
    \texttt{TemporalOwner}, \texttt{VectorOwner}); graph, cache, and mesh stores
    exist but are \emph{not} transaction owners.  The Reconciler emits a
    hash-checkable \texttt{CompletionManifest} (COMPLETE / DEGRADED / FAILED).}
  \label{fig:spine}
\end{figure}
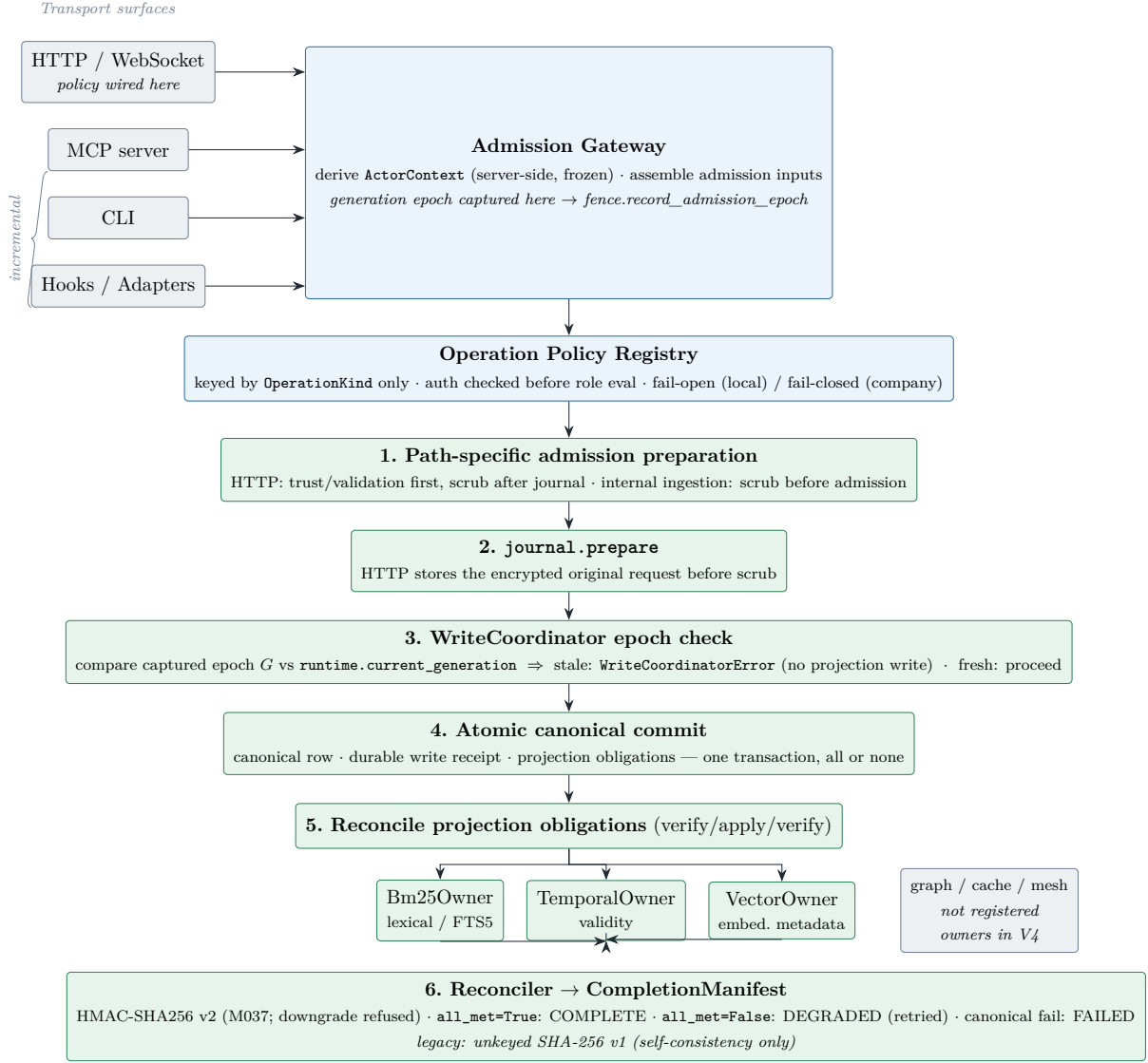

\Cref{fig:context} shows the overall entry path. \Cref{fig:spine} shows the
transaction spine: the transactional obligation ledger, projection owners, and
completion manifest that together enforce projection consistency.


\textbf{Store catalog.} \slm{} manages six primary SQLite files at a configured
data root: \code{memory.db} (canonical facts, FTS5 indexes, entity/graph tables,
vec0 virtual table, mesh coordination tables, RBAC tables);
\code{learning.db} (behavioral signals, reward and outcome tables, LightGBM ranker
state, soft prompts, skill-evolution lineage); \code{audit\_chain.db}
(append-only hash-chained audit trail); \code{pending.db} (pending operations
queue); \code{llmcache.db} (exact-match LLM cache); and \code{admission\_journal.db}
(encrypted admission journal). The Scale Engine manages two optional derived
projections---a CozoDB graph store and a LanceDB vector store---through a
\textbf{prepare} $\to$ \textbf{verify} $\to$ \textbf{promote} $\to$
\textbf{rollback} lifecycle, with SQLite~+~sqlite-vec remaining the canonical
source of truth and active retrieval backend at all times.

\textbf{Admission gateway.} On the canonical HTTP \code{/remember} and
internal-ingestion paths, the gateway derives an immutable \code{ActorContext}
(principal\_id, roles, allowed profiles, active profile\_id, transport, and a
session-token hash---from the authenticated session, never from the request body) and
assembles the operation's admission inputs (a stable idempotency key, operation kind,
resource IDs, scope, deadline, required schema capability, and a redacted payload hash)
on the path's request record---\code{RememberRequest} for HTTP,
\code{IngestionRequest} for internal ingestion---together with a durable receipt. (An
\code{OperationRequest} dataclass in \code{core/operation\_request.py} defines this
envelope as the \emph{target} unified shape; the shipping canonical paths use the split
request records above rather than constructing it.) The \code{ActorContext} is a frozen
dataclass (\code{@dataclass(frozen=True, slots=True)}); no downstream path can mutate it.

\begin{figure}[htbp]
  \centering
  \resizebox{\ifdim\width>\linewidth\linewidth\else\width\fi}{!}{%
\begin{tikzpicture}[
      node distance=0.42cm and 1.5cm,
      adm/.style={slmgate, minimum width=7.4cm, minimum height=0.66cm},
      txn/.style={slmspine, minimum width=7.4cm, minimum height=0.66cm},
      rej/.style={slmreject, minimum width=3.2cm, minimum height=1.05cm},
      dec/.style={diamond, aspect=3.7, draw=slmamber, fill=slmamberbg,
                  align=center, inner sep=2pt, font=\scriptsize,
                  minimum width=7.4cm, minimum height=0.8cm},
    ]

    \node[adm] (auth) {\textbf{1. HTTP \texttt{/remember}: authenticate + RBAC WRITE}};
    \node[adm, below=of auth] (valid) {\textbf{2. Deterministic request validation}};
    \node[adm, below=of valid] (trust) {\textbf{3. Trust hook evaluation}};
    \node[adm, below=of trust] (policy) {%
      \textbf{4. Derive immutable \texttt{ActorContext} + evaluate operation policy}};
    \node[adm, below=of policy] (request) {%
      \textbf{5. Build \texttt{RememberRequest}}\\[-1pt]
      {\scriptsize assign \texttt{idempotency\_key} once, outside retry}};
    \node[adm, below=of request] (capture) {%
      \textbf{6. Capture generation epoch \(G\)}};
    \node[txn, below=of capture] (journal) {%
      \textbf{7. \texttt{journal.prepare}}\\[-1pt]
      {\scriptsize encrypted original request is durable before ingestion scrub}};
    \node[dec, below=0.48cm of journal] (epoch) {%
      \textbf{8. WriteCoordinator:} \(G=\texttt{current\_generation}\,?\)};
    \node[rej, right=1.25cm of epoch] (stale) {%
      \textbf{STALE}\\\texttt{WriteCoordinatorError}\\
      {\scriptsize retryable; no commit}};
    \node[txn, below=0.52cm of epoch] (commit) {%
      \textbf{9. Atomic commit}\\[-1pt]
      {\scriptsize canonical row \(\cdot\) write receipt \(\cdot\) projection obligations}};
    \node[txn, below=of commit] (receipt) {%
      \textbf{10. Return \texttt{status=queryable}}\\[-1pt]
      {\scriptsize canonical fact and SQLite FTS are immediately readable}};
    \node[txn, below=of receipt] (owners) {%
      \textbf{11. Projection reconciliation}\\[-1pt]
      {\scriptsize obligation states: APPLIED \(\cdot\) VERIFIED \(\cdot\) FAILED}};
    \node[txn, below=of owners] (manifest) {%
      \textbf{12. Derive \texttt{CompletionManifest}}\\[-1pt]
      {\scriptsize COMPLETE \(\cdot\) DEGRADED \(\cdot\) FAILED}};

    \draw[slmflow] (auth) -- (valid);
    \draw[slmflow] (valid) -- (trust);
    \draw[slmflow] (trust) -- (policy);
    \draw[slmflow] (policy) -- (request);
    \draw[slmflow] (request) -- (capture);
    \draw[slmflow] (capture) -- (journal);
    \draw[slmflow] (journal) -- (epoch);
    \draw[slmflow] (epoch.east) -- (stale.west)
      node[slmlabel, midway, above=1pt] {No};
    \draw[slmflow] (epoch.south) -- (commit.north)
      node[slmlabel, midway, right=2pt] {Yes};
    \draw[slmflow] (commit) -- (receipt);
    \draw[slmflow] (receipt) -- (owners);
    \draw[slmflow] (owners) -- (manifest);

  \end{tikzpicture}}
  \caption{HTTP \texttt{/remember} admission sequence.  Authentication and
    WRITE authorization precede deterministic validation and the trust hook.
    The server then derives authority in an immutable \texttt{ActorContext},
    evaluates the operation policy, builds one idempotent \texttt{RememberRequest},
    and captures generation \(G\).  The encrypted original request is journaled
    before the ingestion writer scrubs it.  The writer checks \(G\) before one
    transaction commits the canonical row, durable receipt, and projection
    obligations.  The caller receives \texttt{status=queryable}; projection
    reconciliation subsequently records APPLIED, VERIFIED, or FAILED evidence.}
  \label{fig:admission}
\end{figure}
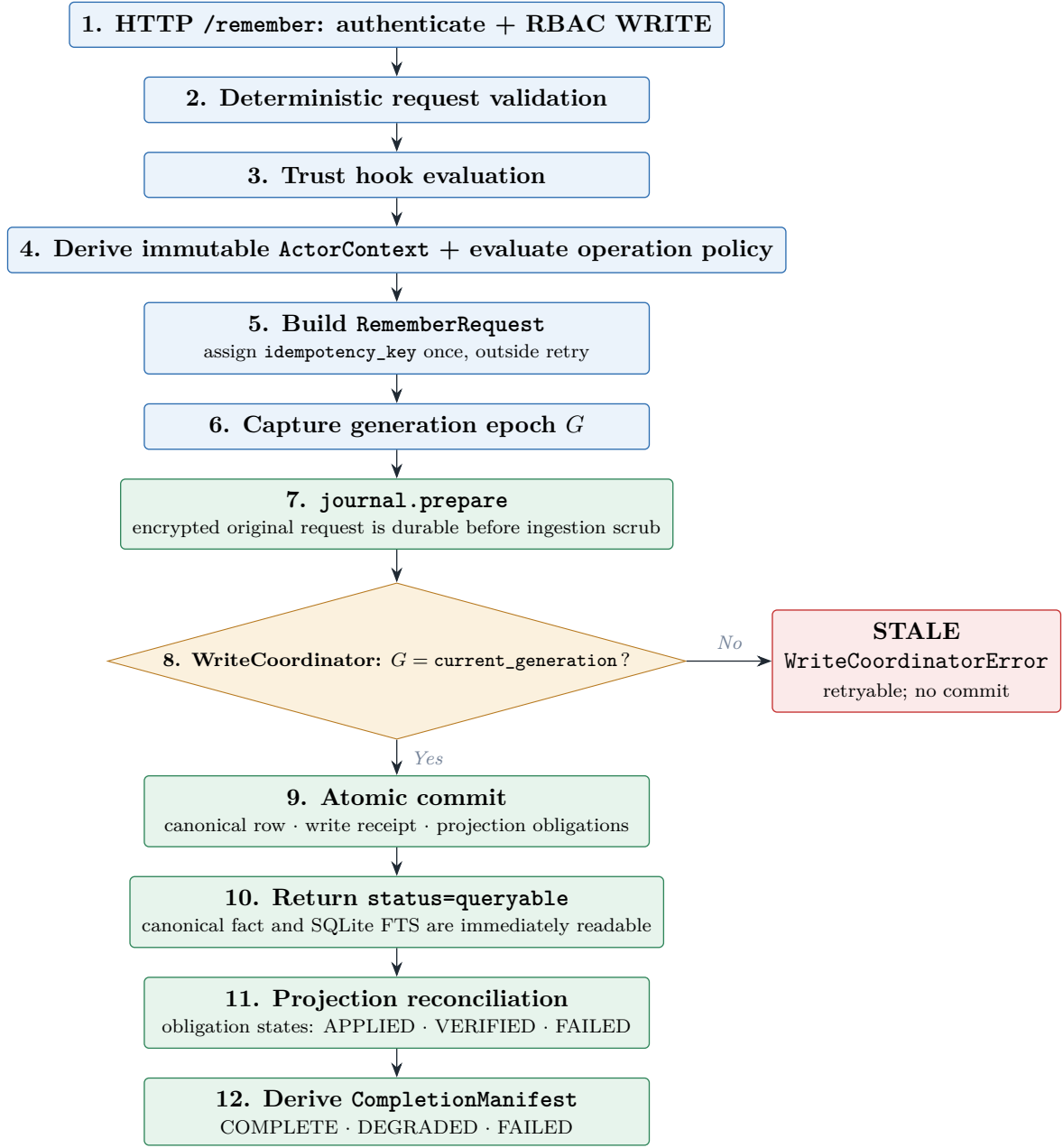

\Cref{fig:admission} shows the admission and fence protocol for a canonical
\code{/remember} call.

\textbf{Delivery surfaces.} The daemon hosts the HTTP REST API, dashboard, and HTTP
MCP endpoint in a single FastAPI application. The MCP server serves both HTTP and
stdio transports; the CLI runs as a separate process with canonical \code{remember}
proxying to the daemon. Nine separately packaged framework adapters implement each
framework's native memory interface backed by the local SLM engine: LangGraph
(\code{BaseStore}), Semantic Kernel (\code{VectorStore}), Microsoft Agent Framework
(\code{ContextProvider}), LangChain (\code{BaseChatMessageHistory}), LlamaIndex
(\code{BaseChatStore}), CrewAI (\code{StorageBackend}), AutoGen (\code{Memory}),
Google ADK (\code{BaseMemoryService}), and OpenAI Agents (\code{SessionABC}).

\textbf{Three operating modes.} Mode~A (fully local, zero cloud:
nomic-embed-text-v1.5 768-d, deterministic rule extraction, local PyTorch
cross-encoder reranker). Mode~B (local Ollama LLM: Phi-3 / Llama 3.2, local
embeddings, local PyTorch reranker). Mode~C (cloud text-embedding-3-large
3072-d, configured cloud LLM, Cohere reranker; content is sent to the configured
provider). Every mode reports \code{eu\_ai\_act\_compliant=None}; intended use and
deployment context are required for legal classification. Mode~A
is the default at installation; ``no cloud in the memory path'' applies to Modes~A
and B, not universally.

\section{Governance Control Plane}
\label{sec:governance}

The strategic premise of \slm{} V4 is \emph{rent the LLM, own the memory}.
Governance is not retrofitted---it is the substrate on which canonical operations run.
The controls in this section execute locally and in-process, without a network call.
Memory \emph{processing} is fully local in Modes~A and B; the opt-in Mode~C may send
configured embedding, reranking, extraction, or answer payloads to a provider, while
governance state and decisions remain local.

\textbf{Governing invariant (scope).} V4 realises the full envelope (one authenticated
actor, one profile generation, one policy decision, one durable receipt, one verifiable
completion state) on the canonical HTTP \code{/remember} path and internal ingestion,
over the three registered projection owners; declared MCP tools and selected CLI
mutations additionally obtain a shared policy \emph{decision} but do not yet share
that operation envelope or receipt; direct legacy mutations and HOOK/ADAPTER surfaces
remain incremental. We present it as the enforced design of the canonical write path
and the target for the platform, not as a property already enforced on every surface.

\subsection{Server-Derived Actor Identity}

\code{ActorContext} (\code{core/actor\_context.py}) is a frozen dataclass that carries
the complete, server-resolved identity for a single operation. Its fields are
populated exclusively from server-authenticated state---session tokens, daemon
descriptors, the profile runtime, and RBAC results---\textbf{never from the request
body}. Key fields: \code{principal\_id} (stable user or operator identifier, never a
raw session token); \code{roles} (a \code{FrozenSet} drawn from the lattice
$\mathrm{OWNER} \geq \mathrm{ADMIN} \geq \mathrm{MEMBER} \geq \mathrm{VIEWER}$,
plus SYSTEM and ANONYMOUS); \code{session\_token\_hash} (first 16 hex chars of the
SHA-256 of the raw token, for audit attribution only; no raw session material is
retained on any path); \code{transport} (HTTP / MCP / CLI / MESH / HOOK / INTERNAL /
DASHBOARD / ADAPTER).

\subsection{Operation Policy Registry}

\code{OperationPolicyRegistry} (\code{core/operation\_policy\_registry.py}) is the
declarative admission layer. \code{evaluate()} is \emph{pure and CPU-only}---no
file, no network, no database access---completing in microseconds. The internal policy
table is a \code{types.MappingProxyType}, preventing external mutation after
construction. One registry makes ``the same evaluator and policy table on each
participating surface'' a checkable property---any caller that invokes it obtains
identical evaluation semantics---rather than universal surface coverage, which the
registry alone does not establish.

The \code{evaluate(kind, actor, mode)} chain runs an ordered 8-step decision sequence:
resolve operation kind; check the explicit deny list; look up the registered
\code{OperationPolicy} (fail-open in local modes, fail-closed in company/remote modes
for unknown kinds); empty-roles check (deny all); \textbf{authentication before
authorisation}; role disjoint check; transport check; payload size annotation (never
a rejection). The registry contains 20 operation kinds; \Cref{tab:policy} shows a
representative subset.

\begin{figure}[htbp]
\centering
\resizebox{\ifdim\width>\linewidth\linewidth\else\width\fi}{!}{%
\begin{tikzpicture}[
    node distance=0.42cm and 1.1cm,
    box/.style={rectangle, rounded corners=2pt, draw=black!70, fill=white,
                font=\footnotesize, text width=3.4cm, align=center,
                minimum height=0.48cm, inner sep=3pt},
    dec/.style={diamond, draw=black!70, fill=gray!10,
                font=\footnotesize\itshape, text width=2.25cm, align=center,
                inner sep=1.2pt, aspect=2.5},
    deny/.style={rectangle, draw=black!50, fill=red!12,
                 font=\scriptsize, text width=2.5cm, align=center,
                 minimum height=0.45cm, inner sep=3pt},
    allow/.style={rectangle, rounded corners=2pt, draw=black!50,
                  fill=green!15, font=\footnotesize\bfseries,
                  minimum height=0.48cm, inner sep=3pt},
    arr/.style={->, >=stealth, thin},
]

\node[box] (req) {Request: raw \texttt{kind}, server-derived actor, mode};
\node[box, below=of req] (resolve) {1. Resolve raw kind to \texttt{OperationKind}};
\node[dec, below=of resolve] (unknown) {Unknown raw kind?};
\node[box, right=1.35cm of unknown, fill=yellow!15] (unknownDecision)
  {Unknown-kind decision\\{\scriptsize local: allow; company/remote: deny}};

\node[dec, below=of unknown] (xdeny) {2. Explicit deny?};
\node[deny, right=1.35cm of xdeny] (d1) {\texttt{explicit\_deny}};
\node[box, below=of xdeny] (lookup) {3. Policy lookup by kind};
\node[dec, below=of lookup] (found) {Policy found?};
\node[dec, below=of found] (roleset) {4. Required roles empty?};
\node[deny, right=1.35cm of roleset] (d2) {\texttt{policy\_denies\_}\\\texttt{all\_roles}};
\node[dec, below=of roleset] (auth) {5. Auth required and missing?};
\node[deny, right=1.35cm of auth] (d3) {\texttt{authentication\_}\\\texttt{required}};
\node[dec, below=of auth] (role) {6. Any required role?};
\node[deny, right=1.35cm of role] (d4) {\texttt{insufficient\_}\\\texttt{roles}};
\node[dec, below=of role] (xport) {7. Transport allowed?};
\node[deny, right=1.35cm of xport] (d5) {\texttt{transport\_}\\\texttt{not\_allowed}};
\node[box, below=of xport] (payload) {8. Annotate oversized payload (never deny)};
\node[allow, below=of payload] (allow) {ALLOW};

\draw[arr] (req) -- (resolve);
\draw[arr] (resolve) -- (unknown);
\draw[arr] (unknown) -- node[above,font=\tiny]{yes} (unknownDecision);
\draw[arr] (unknown) -- node[left,font=\tiny]{no} (xdeny);
\draw[arr] (xdeny) -- node[above,font=\tiny]{yes} (d1);
\draw[arr] (xdeny) -- node[left,font=\tiny]{no} (lookup);
\draw[arr] (lookup) -- (found);
\draw[arr] (found.east) -| node[pos=0.2,above,font=\tiny]{no} (unknownDecision.south);
\draw[arr] (found) -- node[left,font=\tiny]{yes} (roleset);
\draw[arr] (roleset) -- node[above,font=\tiny]{yes} (d2);
\draw[arr] (roleset) -- node[left,font=\tiny]{no} (auth);
\draw[arr] (auth) -- node[above,font=\tiny]{yes} (d3);
\draw[arr] (auth) -- node[left,font=\tiny]{no} (role);
\draw[arr] (role) -- node[above,font=\tiny]{no} (d4);
\draw[arr] (role) -- node[left,font=\tiny]{yes} (xport);
\draw[arr] (xport) -- node[above,font=\tiny]{no} (d5);
\draw[arr] (xport) -- node[left,font=\tiny]{yes} (payload);
\draw[arr] (payload) -- (allow);

\end{tikzpicture}}
\caption{\textbf{RBAC decision flow} in
  \code{OperationPolicyRegistry.evaluate()}.  Kind resolution and the
  mode-dependent unknown-kind decision precede explicit deny and policy lookup.
  A registered policy with no required roles denies all actors.  Authentication
  is tested before role authorization, followed by transport.  Payload size is
  advisory: oversized input annotates an otherwise allowed decision.}
\label{fig:rbac_flow}
\end{figure}
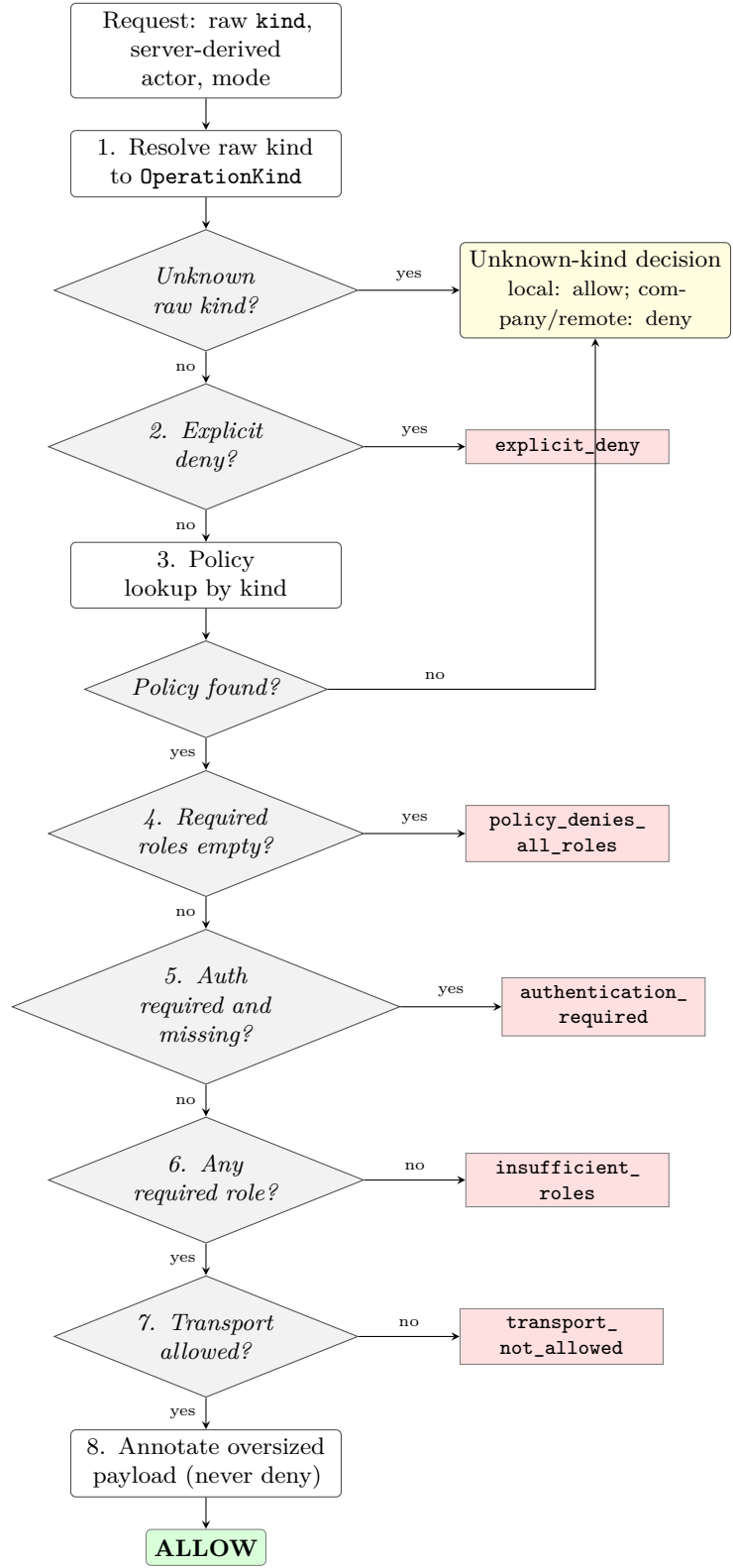

\Cref{fig:rbac_flow} illustrates the evaluation chain.

\begin{table}[h]
\centering
\small
\begin{tabular}{@{}llll@{}}
\toprule
Operation & Required roles & Allowed transports & Audit \\
\midrule
\code{REMEMBER}        & OWNER, ADMIN, MEMBER     & all          & standard \\
\code{RECALL}          & all (incl.\ VIEWER)      & all          & none     \\
\code{FORGET} / \code{CORRECT}  & OWNER, ADMIN  & all          & full     \\
\code{ERASE}           & OWNER only               & all          & full$^*$ \\
\code{BACKUP}          & OWNER only               & all          & standard \\
\code{RESTORE\_BACKUP} & OWNER only               & all          & full     \\
\code{MODE\_CHANGE}    & OWNER only               & admin transports & full \\
\code{SCHEMA\_MIGRATE} & OWNER only               & CLI / INTERNAL only & full \\
\code{MESH\_SEND}      & OWNER, ADMIN, MEMBER     & mesh transports    & standard \\
\code{EVOLVE\_SKILL}   & OWNER, ADMIN             & all                & full \\
\bottomrule
\end{tabular}
\caption{Representative policy table from \code{OperationPolicyRegistry.default()}.
  The full table contains 20 operation kinds.
  $^*$\code{ERASE} additionally sets \code{resource\_ownership\_check=True}
  and \code{redaction\_policy="full"}.}
\label{tab:policy}
\end{table}

\subsection{Multi-Scope Isolation and Profile Generation Fence}

\slm{} separates two orthogonal isolation axes: the \emph{profile}
(individual / team / company namespace) and the \emph{scope}
(\code{personal}, \code{shared}, or \code{global}) within and across profiles.
Every tenant-owned memory table carries a \code{profile\_id} column as a partition
key; cross-scope access is \emph{default-deny}. A fact in \code{personal} scope is
not visible to a request scoped to \code{shared} unless the writer granted it
explicitly---enforced at write time (scope label committed with the fact) via
profile-scoped predicates across all five retrieval channels.

The generation fence rejects a stale, superseded-epoch admitted \emph{write} on the
canonical remember path; it is not the retrieval-channel isolation mechanism. Scope of
isolation claim: authorization-layer control, not cryptographic multi-tenancy; a
single OS user with direct disk access is outside the threat model.

\subsection{GDPR Export and Erasure}

\code{GDPRCompliance} (\code{compliance/gdpr.py}) implements three GDPR data rights
locally---no request leaves the device. The Right to Erasure (Art.\ 17)
\code{forget\_profile} proceeds in three ordered phases:

\begin{enumerate}[topsep=4pt, itemsep=2pt]
  \item \textbf{Audit before deletion (Art.\ 5(2) accountability).} The runtime
    writes an erasure-request record to the hash-chained \code{audit\_chain.db}
    \emph{before} any row is deleted. This is a \emph{hard precondition}: if the
    audit write fails, the erasure fails \emph{closed}---it aborts and deletes
    nothing, so no erasure ever occurs without an accountability record.
  \item \textbf{Cache and vector purge.} The brain cache DB is purged before main-DB
    deletions; the vector store's \code{delete()} is called per fact ID.
  \item \textbf{Schema-driven wipe.} Every discovered profile-scoped table is deleted
    in a single pass. After all deletions, the runtime re-counts residual rows and
    distinguishes \emph{deletion} from \emph{provable deletion}, and carries the two
    as separate flags. Most systems report the first. A deletion that succeeded but
    cannot be demonstrated afterwards is a different object, for a regulator, from one
    that can, and conflating them is what makes an erasure claim unfalsifiable. The
    operation is also fail-closed on its own audit precondition: if the tamper-evident
    chain cannot be written, the erasure does not proceed, so there is no state in
    which data is destroyed and the record of destroying it is missing---which is the
    state Article~5(2) accountability cannot survive. The wipe is driven from the
    schema rather than an enumerated table list, so a table added later is covered
    without anyone remembering to add it; compliance drift by omission is designed out.
    It returns an explicit \code{erasure\_complete} flag, set only when no residue
    remains and no per-table delete, residue re-count, context-cache purge, learning-DB
    reset, vector-purge, or owner-proof failure occurred---any such error, including a
    top-level purge exception or an uncountable table, forces \code{erasure\_complete=0}
    (fail-closed reporting). Backup artifacts and remote peers are outside the erasure
    scope (future work).
\end{enumerate}

\subsection{Hash-Chained Audit Trail and EU AI Act Deployment Checklist}

\code{AuditChain} (\code{compliance/audit.py}) is a tamper-evident hash-chain audit
log in its own SQLite file (\code{audit\_chain.db}), completely independent of the
main store. The chain hash is SHA-256 over the prior hash and each entry's fields;
verification walks every entry in insertion order and returns \code{False} on the
first mismatch. A sibling sidecar (an out-of-table anchor recording chain length and
head hash) detects DB-only suffix or whole-chain truncation \emph{when the sidecar was
successfully written and remains intact}; it is a same-host integrity aid, not a remote
trust anchor, and an anchor-write failure is logged rather than blocking the audited write.

\code{EUAIActChecker.check\_compliance(mode)} produces a frozen assessment record
describing observable mode capabilities and missing deployment context. For every
mode it returns \code{compliant=None}, \code{risk\_category="undetermined"}, and
unset transparency and human-oversight status until intended use, actor roles, and
deployment evidence are supplied. Local versus provider-assisted processing alone
cannot determine an EU AI Act risk category. Two dates matter for a reader
positioning a deployment against it, and both fell before this paper's first
version: obligations for general-purpose model providers took effect on
2025-08-02, and the Commission's enforcement powers over those providers---requests
for information, model access, and recall---became applicable on 2026-08-02,
alongside the Article~50 transparency duties, whose enforcement sits with national
market-surveillance authorities. The high-risk obligations originally scheduled for
that same date were deferred to 2027-12-02 and did not take effect. The checklist
records deployment context and does not track this calendar; a reader must confirm
the current position independently. This is an engineering checklist, not
a conformity assessment or legal certification.

\section{Reliability Spine}
\label{sec:spine}

The governance layer answers \emph{who may act and on which scope}. This section
answers \emph{how we guarantee that the action succeeds consistently}. Four mechanisms
form the V4 reliability spine: generation-fenced admission, verifiable memory
transactions, cross-store verified erasure, and bi-temporal as-of. SLM-Mesh provides
the coordination layer. Together they realise the governing design goal on the write
path.

\subsection{Generation-Fenced Admission}
\label{sec:spine:fence}

A profile deleted and recreated with the same ID must not allow a cached context to
write into the new profile's data. The generation fence uses a process-local,
\code{threading.RLock}-protected, TTL-bound map of
$\code{(profile\_id, idempotency\_key)} \to \code{(epoch, timestamp)}$
in \code{storage/generation\_fence.py}, with entries expiring after 300 seconds.

At admission, \code{record\_admission\_epoch} stores the current profile generation
for the idempotency key. If two concurrent admissions record different epochs for the
same key, the fence sets the entry to \code{\_CONFLICT\_EPOCH = -1}---a sentinel that
can never equal a real generation ($\geq 0$)---so both requests are subsequently
rejected. When \code{CanonicalRememberRuntime.\_handle\_admission} processes the write,
it calls \code{admitted\_epoch()} and compares against the runtime's current
generation; a mismatch raises \code{ValueError("epoch is stale")}, which the
\code{WriteCoordinator} wraps as \code{WriteCoordinatorError} \emph{before} any
projection writer is called. The generation epoch is server-derived and cannot be
supplied by a caller.

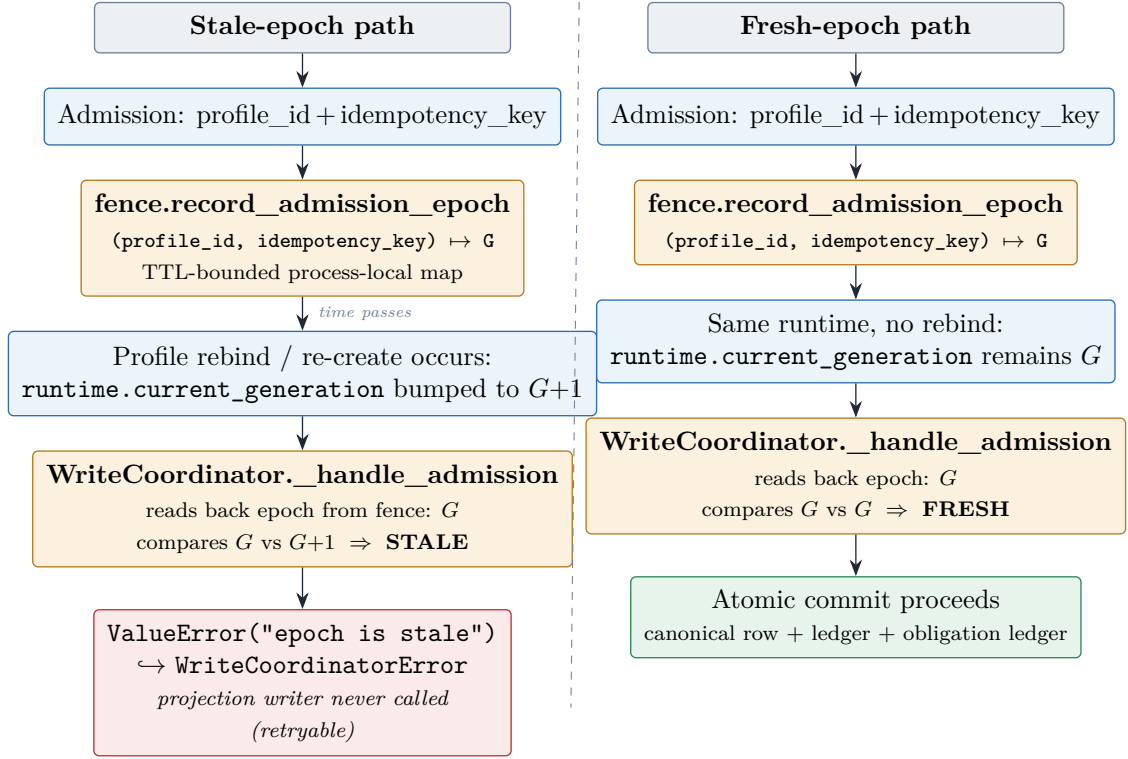
\begin{figure}[htbp]
  \centering
  \resizebox{\ifdim\width>\linewidth\linewidth\else\width\fi}{!}{%
\begin{tikzpicture}[
      node distance=0.55cm and 2.2cm,
      fstep/.style={slmgate, minimum width=5.5cm, minimum height=0.75cm},
      okstep/.style={slmspine, minimum width=5.5cm, minimum height=0.75cm},
      failstep/.style={slmreject, minimum width=3.5cm, minimum height=0.85cm},
      fencebox/.style={slmwarn, minimum width=5.5cm, minimum height=0.75cm},
      note/.style={slmlabel, text width=3.0cm, align=left},
      el/.style={slmlabel, font=\tiny\itshape},
      >=Stealth,
    ]

    \node[slmbox, fill=slmgraybg, draw=slmgray,
          minimum width=5.5cm, font=\small\bfseries] (hd_stale)
      {Stale-epoch path};

    \node[fstep, below=0.45cm of hd_stale] (a1) {%
      Admission: profile\_id\,+\,idempotency\_key};
    \node[fencebox, below=0.45cm of a1] (a2) {%
      \textbf{fence.record\_admission\_epoch}\\[1pt]
      {\scriptsize \texttt{(profile\_id, idempotency\_key) $\mapsto$ G}}\\
      {\scriptsize TTL-bounded process-local map}};
    \node[fstep, below=0.45cm of a2] (a3) {%
      Profile rebind / re-create occurs:\\
      \texttt{runtime.current\_generation} bumped to \(G{+}1\)};
    \node[fencebox, below=0.45cm of a3] (a4) {%
      \textbf{WriteCoordinator.\_handle\_admission}\\[1pt]
      {\scriptsize reads back epoch from fence: \(G\)}\\
      {\scriptsize compares \(G\) vs \(G{+}1\) \(\;\Rightarrow\;\) \textbf{STALE}}};
    \node[failstep, below=0.55cm of a4] (a5) {%
      \texttt{ValueError(``epoch is stale'')}\\
      $\hookrightarrow$ \texttt{WriteCoordinatorError}\\
      {\scriptsize\itshape projection writer never called}\\
      {\scriptsize\itshape (retryable)}};

    \draw[slmflow] (hd_stale)--(a1);
    \draw[slmflow] (a1)--(a2);
    \draw[slmflow] (a2)--(a3) node[el, right=2pt, midway] {time passes};
    \draw[slmflow] (a3)--(a4);
    \draw[slmflow] (a4)--(a5);

    \node[slmbox, fill=slmgraybg, draw=slmgray,
          minimum width=5.5cm, font=\small\bfseries,
          right=1.8cm of hd_stale] (hd_fresh)
      {Fresh-epoch path};

    \node[fstep, below=0.45cm of hd_fresh] (b1) {%
      Admission: profile\_id\,+\,idempotency\_key};
    \node[fencebox, below=0.45cm of b1] (b2) {%
      \textbf{fence.record\_admission\_epoch}\\[1pt]
      {\scriptsize \texttt{(profile\_id, idempotency\_key) $\mapsto$ G}}};
    \node[fstep, below=0.45cm of b2] (b3) {%
      Same runtime, no rebind:\\
      \texttt{runtime.current\_generation} remains \(G\)};
    \node[fencebox, below=0.45cm of b3] (b4) {%
      \textbf{WriteCoordinator.\_handle\_admission}\\[1pt]
      {\scriptsize reads back epoch: \(G\)}\\
      {\scriptsize compares \(G\) vs \(G\) \(\;\Rightarrow\;\) \textbf{FRESH}}};
    \node[okstep, below=0.55cm of b4] (b5) {%
      Atomic commit proceeds\\
      {\scriptsize canonical row + ledger + obligation ledger}};

    \draw[slmflow] (hd_fresh)--(b1);
    \draw[slmflow] (b1)--(b2);
    \draw[slmflow] (b2)--(b3);
    \draw[slmflow] (b3)--(b4);
    \draw[slmflow] (b4)--(b5);

    \draw[draw=slmgray, dashed, line width=0.4pt]
      ($(hd_stale.north east)!0.5!(hd_fresh.north west)$) --
      ($(a5.south east)!0.5!(b5.south west)$);

    \node[slmlabel, text width=12.8cm, align=center,
          below=0.4cm of a5, xshift=3.7cm] {%
      The generation epoch is \emph{server-derived} and cannot be supplied by a caller.
      \texttt{ActorContext.active\_profile\_generation} defaults to~0 for HTTP actors;
      the fence is independent of that field and relies solely on the
      \texttt{(profile\_id, idempotency\_key) $\to$ epoch} map
      (\texttt{storage/generation\_fence.py}).};

  \end{tikzpicture}}
  \caption{Profile generation fence (exp7).  The fence is a process-local,
    TTL-bounded map of \texttt{(profile\_id, idempotency\_key) $\to$ epoch}.
    At admission, \texttt{record\_admission\_epoch()} stores the current
    generation~\(G\).  If the profile is deleted and recreated (or otherwise
    rebound) before the write coordinator runs, the runtime's
    \texttt{current\_generation} advances to \(G{+}1\).
    \texttt{\_handle\_admission} reads back~\(G\) from the fence and compares it
    to the current generation: a mismatch raises
    \texttt{WriteCoordinatorError} before any projection writer is called
    (retryable).  On a matching generation the atomic commit proceeds normally.
    The fence does \emph{not} perform per-owner revision CAS; it rejects
    stale-epoch admissions at the coordinator level.}
  \label{fig:fence}
\end{figure}

\Cref{fig:fence} shows the fence protocol. \emph{Scope}: in-process, TTL-bounded,
not SQLite-backed; not a revision CAS; does not survive process restart. The canonical
HTTP \code{/remember} and internal-ingestion boundaries synthesize a non-empty
idempotency key (a UUID) when the caller omits one, so omission at those public
boundaries does not bypass the fence.

\subsection{Verifiable Memory Transactions}
\label{sec:spine:transactions}

Every canonical remembered-fact write must produce consistent state across three
physical representations: BM25 token index, temporal store, and vector projection.
V4 routes the canonical remember path through a transactional obligation ledger
with per-projection ownership and a hash-sealed completion manifest; registry-gated
MCP/CLI operations and direct legacy mutations do not record these obligations.

\textbf{Obligation ledger.} On every canonical remembered write,
\code{\_record\_projection\_obligations} creates obligation records for the three
registered owners (\code{"bm25"}, \code{"temporal"}, \code{"vector"}) in
\code{projection\_obligations} with \code{state=PENDING}. This is the V4
transactional obligation ledger---distinct from the Mesh remote-delivery outbox.

\textbf{Projection owners.} Each of the three owners (\code{Bm25Owner},
\code{TemporalOwner}, \code{VectorOwner} in \code{core/transactions/concrete\_owners.py})
implements the \code{ProjectionOwner} protocol:
\code{apply} / \code{verify} / \code{compensate} / \code{erase} / \code{prove\_erased}
/ \code{health}. Their checksums are SHA-256 over owner name, sorted fact IDs, and
per-fact fingerprints.

\textbf{Completion manifest.} After all owners have been applied and verified,
\code{CompletionManifest} assembles an \code{OwnerEvidence} record per obligation and
seals the canonical JSON serialisation: on \mbox{M037}-capable databases\footnote{\mbox{M037}
denotes schema migration~037 (schema version~37), which added the per-installation key column
from which the manifest seal's \mbox{HMAC-SHA256} key is derived.} with an
installation-key-derived \mbox{HMAC-SHA256} (version~2, refusing a version-1 downgrade
at verification); on older schemas with a deterministic unkeyed \mbox{SHA-256}
(accidental-corruption detection only, not tamper-evidence against a malicious DB
writer). State derives as: \textsc{complete} (canonical write succeeded, all owners in
terminal success); \textsc{degraded} (canonical write succeeded, at least one owner
failed, compensation ran); \textsc{failed} (canonical write itself failed). A
\textsc{degraded} manifest is never silently promoted to \textsc{complete}.

The reconciler maintenance loop redrives pending or missing-manifest obligations
(capped at 10 attempts per obligation), bounding eventual drift from transient
write failures.

\Cref{fig:transaction_spine} shows the transaction spine. \Cref{fig:manifest-states}
shows the manifest state machine.

\begin{figure}[htbp]
\centering
\resizebox{\ifdim\width>\linewidth\linewidth\else\width\fi}{!}{%
\begin{tikzpicture}[
    node distance=0.48cm and 1.0cm,
    box/.style={rectangle, rounded corners=2pt, draw=black!70, fill=white,
                font=\footnotesize, text width=3.2cm, align=center,
                minimum height=0.55cm, inner sep=3pt},
    phase/.style={rectangle, draw=black!50, fill=blue!8,
                  font=\footnotesize\bfseries, text width=3.2cm, align=center,
                  minimum height=0.55cm, inner sep=3pt},
    dec/.style={diamond, draw=black!70, fill=gray!10,
                font=\footnotesize\itshape, text width=2.1cm, align=center,
                inner sep=1.5pt, aspect=2.4},
    state/.style={rectangle, draw=black!40, fill=gray!10,
                  font=\scriptsize, text width=2.0cm, align=center,
                  minimum height=0.46cm, inner sep=2pt},
    manifest/.style={rectangle, rounded corners=2pt, draw=black!60,
                     fill=yellow!20, font=\footnotesize, text width=3.5cm,
                     align=center, minimum height=0.55cm, inner sep=3pt},
    arr/.style={->, >=stealth, thin},
    explicit/.style={->, >=stealth, thin, dashed, draw=red!65},
]

\node[box] (write) {Atomic canonical write + PENDING obligations\\
  {\scriptsize bm25 \(\cdot\) temporal \(\cdot\) vector}};
\node[phase, below=of write] (verify1) {1. Per-owner \texttt{verify()} first};
\node[dec, below=of verify1] (valid1) {Projection valid?};
\node[state, right=1.2cm of valid1] (verified1) {VERIFIED};
\node[phase, below=of valid1] (apply) {2. If invalid: \texttt{apply()}};
\node[dec, below=of apply] (appliedok) {Apply succeeded?};
\node[state, right=1.2cm of appliedok] (applyfailed) {FAILED};
\node[phase, below=of appliedok] (verify2) {3. If applied: \texttt{verify()} again};
\node[dec, below=of verify2] (valid2) {Now valid?};
\node[state, right=1.2cm of valid2] (terminal) {VERIFIED or FAILED};
\node[manifest, below=of valid2] (mfst)
  {Derive \texttt{CompletionManifest}\\
   \textbf{COMPLETE} / \textbf{DEGRADED} / \textbf{FAILED}};
\node[box, left=1.25cm of apply, fill=orange!10] (retry)
  {Maintenance reconciler\\{\scriptsize retry apply attempts, maximum 10}};
\node[box, right=1.25cm of verify2, fill=red!8] (comp)
  {Explicit caller action:\\\texttt{compensate()}\\{\scriptsize not automatic}};

\draw[arr] (write) -- (verify1);
\draw[arr] (verify1) -- (valid1);
\draw[arr] (valid1) -- node[above,font=\tiny]{yes} (verified1);
\draw[arr] (valid1) -- node[left,font=\tiny]{no} (apply);
\draw[arr] (apply) -- (appliedok);
\draw[arr] (appliedok) -- node[above,font=\tiny]{no} (applyfailed);
\draw[arr] (appliedok) -- node[left,font=\tiny]{yes: APPLIED} (verify2);
\draw[arr] (verify2) -- (valid2);
\draw[arr] (valid2) -- (terminal);
\draw[arr] (valid2) -- node[left,font=\tiny]{derive} (mfst);
\draw[arr] (verified1.east) -- ++(0.25,0) |- (mfst.east);
\draw[arr] (applyfailed.east) -- ++(0.25,0) |- (mfst.east);
\draw[arr] (terminal.east) -- ++(0.25,0) |- (mfst.east);
\draw[arr] (retry) |- (verify1);
\draw[explicit] (terminal) -- (comp)
  node[midway, above, font=\tiny, red!70]{explicit only};

\end{tikzpicture}}
\caption{\textbf{Verifiable memory transaction spine.}  Each reconciliation
  attempt verifies the existing projection first.  Only an invalid projection
  is applied, and a successful apply is verified again before the obligation is
  marked VERIFIED; failed apply or verification records FAILED.  Maintenance
  retries are idempotent and apply attempts are bounded at ten.  Compensation is
  a separate explicit service call, not an automatic consequence of failure.
  The manifest is derived from canonical presence and current obligation evidence.}
\label{fig:transaction_spine}
\end{figure}
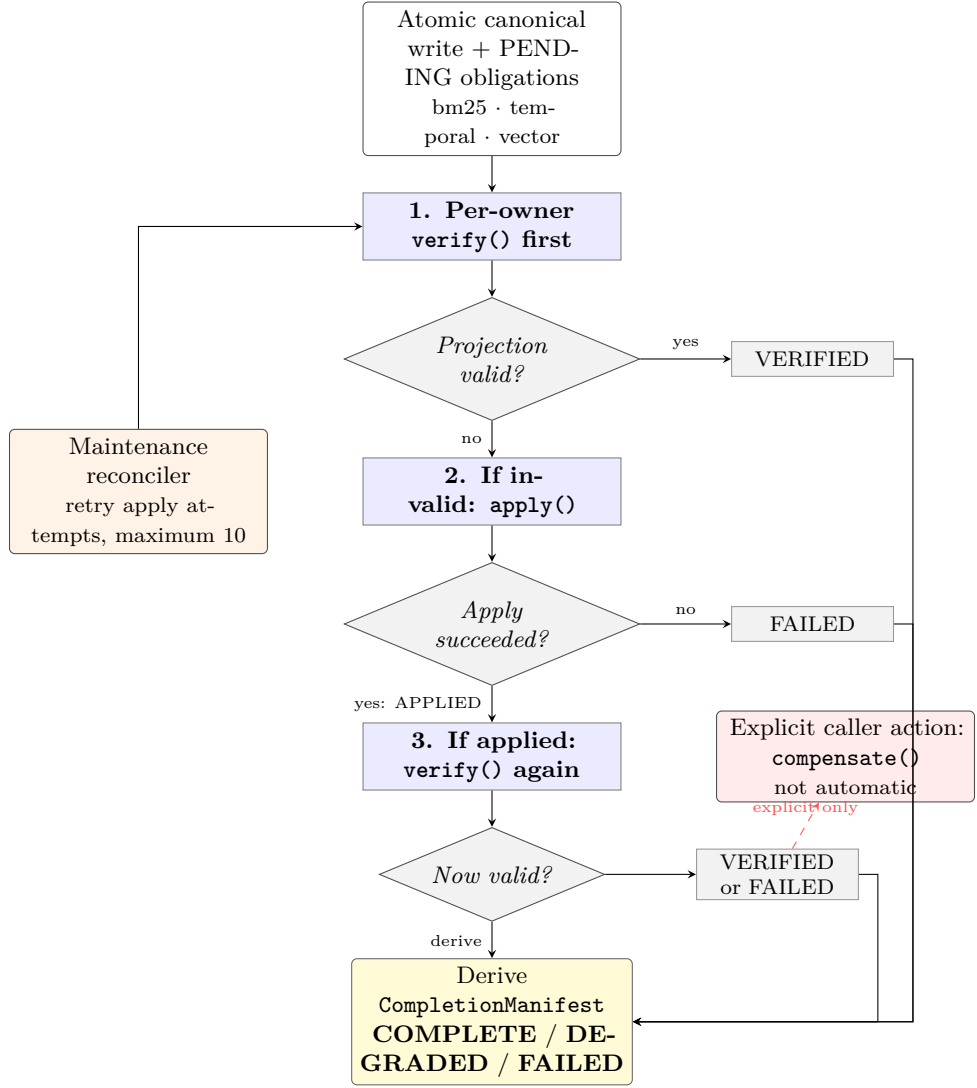
\begin{figure}[htbp]
\centering
\resizebox{\ifdim\width>\linewidth\linewidth\else\width\fi}{!}{%
\begin{tikzpicture}[
    node distance=0.55cm and 0.35cm,
    source/.style={draw, rounded corners=4pt, fill=gray!8,
                   minimum width=4.6cm, minimum height=0.9cm,
                   font=\small, align=center},
    state/.style={draw, double, rounded corners=4pt, fill=gray!16,
                  minimum width=2.5cm, minimum height=0.8cm,
                  font=\small\bfseries, align=center},
    condition/.style={draw, rounded corners=2pt, fill=gray!5,
                      text width=2.65cm, minimum height=0.72cm,
                      font=\scriptsize, align=center, inner sep=2pt},
    arr/.style={->, thick, >=stealth},
    lbl/.style={font=\scriptsize, align=center, fill=white, inner sep=1pt},
]

\node[source] (derive) {\texttt{derive\_state}\\[-1pt]
  {\scriptsize canonical presence + current obligation evidence}};
\node[condition, below=0.75cm of derive, xshift=-3.05cm] (cfail)
  {canonical absent\\or no obligations};
\node[condition, right=of cfail] (ccomplete)
  {all terminal-success\\(VERIFIED / ERASED)};
\node[condition, right=of ccomplete] (cdegraded)
  {canonical present;\\some obligation not met};
\node[state, below=of cfail] (failed) {FAILED};
\node[state, below=of ccomplete] (complete) {COMPLETE};
\node[state, below=of cdegraded] (degraded) {DEGRADED};

\draw[arr] (derive.south) -- ++(0,-0.32) -| (cfail.north);
\draw[arr] (derive.south) -- (ccomplete.north);
\draw[arr] (derive.south) -- ++(0,-0.32) -| (cdegraded.north);
\draw[arr] (cfail) -- (failed);
\draw[arr] (ccomplete) -- (complete);
\draw[arr] (cdegraded) -- (degraded);

\end{tikzpicture}}
\caption{\textbf{Completion-manifest state derivation.}  A reconcile cycle
  recomputes exactly one output from canonical presence and the current
  obligation set: FAILED when the canonical commit is absent or the set is
  empty, COMPLETE when every obligation is terminal-success (VERIFIED or
  ERASED), and DEGRADED otherwise.  The fan-out depicts mutually exclusive
  derivation conditions; there are no state-to-state transitions.}
\label{fig:manifest-states}
\end{figure}
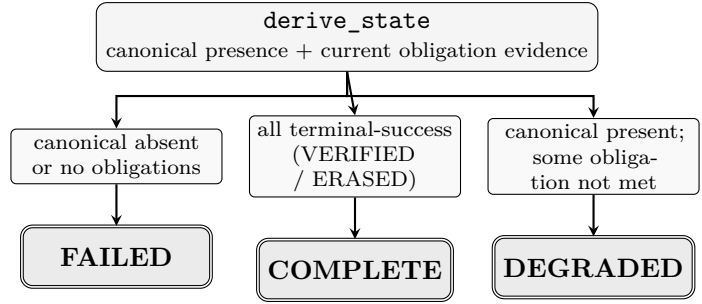

\subsection{Cross-Store Verified Erasure}
\label{sec:spine:erasure}

\code{ErasureService} (\code{core/transactions/erasure.py}) coordinates verified
deletion across all registered projection owners:

\begin{enumerate}[topsep=4pt, itemsep=2pt]
  \item \textbf{Obligation recording.} One \code{ERASE} obligation per owner enters
    \code{projection\_obligations} in \code{PENDING} state.
  \item \textbf{Tombstones.} Before per-owner deletion, a tombstone row is inserted
    per fact ID. A provenance conflict (two erasure operations recording different
    \code{memory\_id} values for the same fact) fails the operation \emph{closed}:
    \code{remove()} aborts before any owner deletion and the receipt is marked
    \textsc{failed}.
  \item \textbf{Per-owner erase.} Each owner's \code{erase(context)} removes its
    physical entries and returns an \code{OwnerErasureProof} with an erasure
    checksum.
  \item \textbf{Proof verification.} \code{ErasureService.finalize()} calls each
    owner's \code{prove\_erased(context)}---a live physical re-query of the
    underlying tables, not a re-hash of the envelope---to re-confirm absence, then
    seals the full receipt with an installation-key-derived \mbox{HMAC-SHA256} on
    \mbox{M037}-capable databases (version~2, refusing a version-1 downgrade) or
    unkeyed \mbox{SHA-256} on older schemas.
\end{enumerate}

The receipt carries \code{all\_erased=True} only when every owner's proof reports
\code{erased=True}; partial residue sets \code{all\_erased=False}. The three
registered owners in V4 are \code{Bm25Owner} (BM25 tokens), \code{TemporalOwner}
(temporal validity rows), and \code{VectorOwner} (embedding metadata).
Backup artifacts and remote peers are not erased and are outside the current scope.

\subsection{SLM-Mesh: Coordination Across All Three Modes}
\label{sec:spine:mesh}

SLM-Mesh provides serverless, per-tenant-isolated coordination through a local SQLite
broker (in \code{memory.db} via \code{mesh\_*}-prefixed tables): authenticated peer
messages, shared key/value state, and advisory locks with TTL-based expiry and fencing
tokens. Eight mesh operations are exposed as MCP tools. The broker is available in all
three operating modes, including fully offline Mode~A.

\textbf{Per-tenant isolation.} Server-assigned principals are bound to a
\code{profile\_id}; \code{check\_cross\_profile\_sender()} rejects any peer ID that
belongs to a different profile. Cross-machine state/lock synchronization currently
covers the default profile only.

\textbf{Authentication and fencing.} When the broker binds to a non-loopback host it
requires \code{SLM\_MESH\_SHARED\_SECRET}---the broker raises \code{RuntimeError} at
construction time if the host is not localhost and the secret is absent. Fencing-token
order resolves advisory lock \emph{views} to a single holder; ordinary
storage-mutation paths do not currently call that primitive, so this is advisory
coordination, not end-to-end single-writer enforcement over storage writes.

\textbf{Cross-machine coordination (opt-in).} Each node periodically pulls a peer's
deltas and merges them deterministically: shared state converges by last-writer-wins
under the total order $(\mathit{revision}, \mathit{node\_id})$; advisory-lock views
resolve under $(\mathit{fencing\_token}, \mathit{node\_id})$. A durable remote outbox
provides bounded durable retry with duplicate possibility (jittered-backoff retry,
48-hour TTL, per-peer cap, dead-letter path); delivery is \emph{not} guaranteed after
cap eviction, enqueue failure, retry exhaustion, or 48-hour expiry. The transport
rejects production plaintext and
supports verified TLS, optional pinned certificate, bearer authentication, and
HMAC-signed messages; opt-in mDNS discovery is off by default.

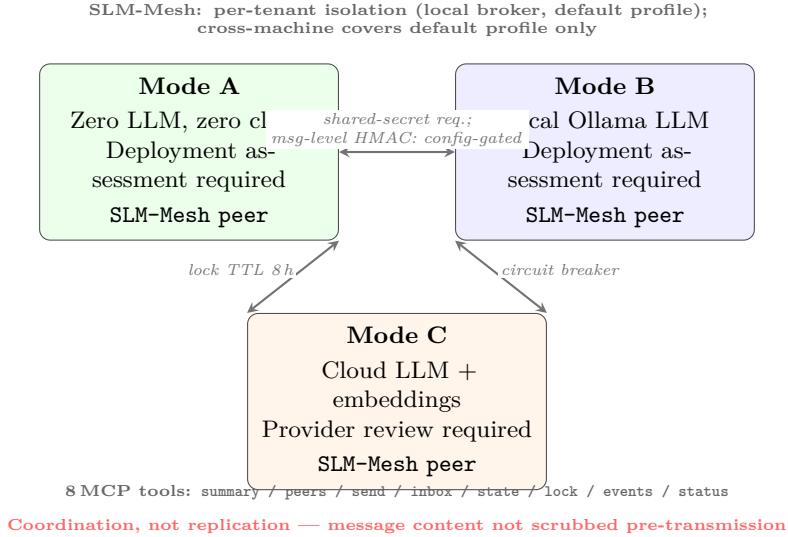
\begin{figure}[htbp]
\centering
\begin{tikzpicture}[
    machine/.style={rectangle, rounded corners=4pt, draw=black!70,
                    fill=white, font=\footnotesize, text width=3.6cm,
                    align=center, minimum height=1.8cm, inner sep=5pt},
    label/.style={font=\tiny\bfseries, text=black!60},
    arr/.style={<->, >=stealth, thick, draw=black!55},
    authlabel/.style={font=\tiny\itshape, text=black!55, fill=white,
                      inner sep=1pt},
]

\node[machine, fill=green!8] (A) at (0,0) {
  \textbf{Mode~A}\\[2pt]
  Zero LLM, zero cloud\\
  Deployment assessment required\\[2pt]
  \texttt{SLM-Mesh peer}
};

\node[machine, fill=blue!7] (B) at (5.5,0) {
  \textbf{Mode~B}\\[2pt]
  Local Ollama LLM\\
  Deployment assessment required\\[2pt]
  \texttt{SLM-Mesh peer}
};

\node[machine, fill=orange!8] (C) at (2.75,-3.3) {
  \textbf{Mode~C}\\[2pt]
  Cloud LLM + embeddings\\
  Provider review required\\[2pt]
  \texttt{SLM-Mesh peer}
};

\draw[arr] (A.east) -- node[authlabel, above=1pt, text width=3.4cm, align=center]{shared-secret req.;\\ msg-level HMAC: config-gated} (B.west);
\draw[arr] (A.south east) -- node[authlabel, left=2pt, pos=0.4]{lock TTL 8\,h} (C.north west);
\draw[arr] (B.south west) -- node[authlabel, right=2pt, pos=0.4]{circuit breaker} (C.north east);

\node[label, text width=11cm, align=center] at (2.75, 1.75)
  {SLM-Mesh: per-tenant isolation (local broker, default profile);\\
   cross-machine covers default profile only};
\node[label] at (2.75, -4.5) {%
  8\,MCP tools: \texttt{summary / peers / send / inbox / state / lock / events / status}};
\node[label, text=red!60] at (2.75, -4.95) {%
  Coordination, not replication --- message content not scrubbed pre-transmission};

\end{tikzpicture}
\caption{%
  \textbf{SLM-Mesh topology across all three operating modes.}
  A mesh instance runs alongside any SLM daemon regardless of mode:
  Mode~A (zero cloud), Mode~B (local LLM), or Mode~C (cloud).
  Non-loopback binding requires a shared secret
  (\texttt{SLM\_MESH\_SHARED\_SECRET}) to start; message-level HMAC
  signing/verification is optional and configuration-dependent (off by
  default for compatibility).  Distributed locks auto-expire after
  8\,hours so a crashed peer cannot deadlock a resource.
  A circuit breaker on \texttt{mesh\_send} fast-fails after three consecutive
  daemon-unreachable errors and re-probes after a 60\,s cooldown.
  Per-tenant isolation---server-assigned principals bound to
  \texttt{profile\_id}---is inherited by every local-broker mesh primitive;
  cross-machine state and lock synchronisation currently covers the default
  profile only.
  Scope: coordination (signals, shared state, locks), not a
  conflict-resolving replicated database.%
}
\label{fig:mesh_topology}
\end{figure}

\Cref{fig:mesh_topology} shows the mesh topology. \emph{Scope}: cross-machine features
are opt-in; WAN-scale operation, per-device revocation, and fault-injection at scale
are future work. SLM-Mesh provides leaderless LWW convergence with fencing-token
advisory locking---\textbf{not} linearizable consensus, not quorum replication, and
not a general CRDT merge.

\subsection{Operational Recovery and Admin Remediation}
\label{sec:spine:recovery}

A shared deployment must ensure that a stuck or failed operation for one tenant neither
corrupts nor silently blocks the others, and that every failure is either auto-recovered or
resolvable by an operator. V4 layers three mechanisms on the transaction spine.

\textbf{Auto-recovery.} The maintenance reconciler redrives pending and failed projection
obligations (capped at ten attempts per obligation); the ingestion path retries a
materialization up to ten times and then records the operation in a durable
\code{dead\_letter\_operations} table with its content, error, and attempt count. Transient
failures heal without human action, and a permanently failed operation is retained rather
than lost.

\textbf{Stall isolation (write-path watchdog).} The single-writer \code{WriteCoordinator}
serializes canonical writes for correctness. To keep one pathological operation from freezing
every tenant's writes, each in-flight item is watched: if the worker exceeds a stall threshold
(default 30\,s, far above the 1\,s per-item deadline) on a single item, the coordinator trips
a circuit breaker so that \emph{new} submissions fail fast with an explanatory
\code{WriterStalledError} (``write subsystem stalled; admin remediation required'') instead of
queuing behind a dead worker; the breaker clears automatically once the item completes.
Callers receive an immediate, actionable error rather than a silent hang. \emph{Scope}: the
watchdog surfaces and fast-fails a stall but does not forcibly terminate the worker thread---a
hung native handler still requires a restart (future work).

\textbf{Surfacing and one-click remediation.} Failures are made visible rather than buried in
logs: the daemon \code{/health} and \code{get\_status} surfaces report
\code{dead\_letter\_count}, \code{degraded\_operations}, and \code{writer\_stalled}. An
OWNER/ADMIN operator can enumerate and resolve them through one remediation surface exposed
uniformly on the dashboard, the CLI (\code{slm ops list} / \code{slm ops resolve}), and MCP
(\code{list\_failed\_operations}, \code{resolve\_operation}), gated by the
\code{OPS\_INSPECT}/\code{OPS\_RESOLVE} policy kinds with a full audit record on resolve.
\code{resolve} supports three actions on a specific operation---\emph{retry} (re-enqueue a
dead-lettered ingestion), \emph{force-reconcile} (redrive its projection obligations
immediately, bypassing the periodic throttle), and \emph{cancel} (mark it terminally
handled). Because a large share of \slm{} operators are non-technical, the dashboard presents
this in plain language---a health banner and a ``needs attention'' table with per-row Retry /
Re-sync / Cancel buttons---so recovery requires neither the CLI nor raw SQL. These paths are
covered by dedicated tests (stall-watchdog fast-fail and recovery, RBAC-gated listing, and
resolve semantics).

\section{The Learning Layer: Measuring Whether a Learner Learns}
\label{sec:learning-layer}

\subsection{An asymmetry between the write path and the behaviour path}

\Cref{sec:spine} states the invariant the write path enforces: one
authenticated actor, one profile generation, one policy decision, one durable
receipt, one verifiable completion state. On that path, a fact cannot enter the canonical
store without satisfying all five. Other transports adopt the gateway incrementally
(\Cref{sec:architecture}), so this is a property of the primary write path and not yet
of every route into the store.

Nothing comparable governed the path that changes the agent's \emph{behaviour}. A
reward could move a channel-selection posterior, a posterior could reweight
retrieval, and a soft prompt could enter a session prefix, with no admission
decision, no receipt, and no completion state anywhere in that chain. We
described the result in the previous version of this paper as a governed learning
and behaviour brain. It was neither governed nor learning, and the second half is
measurable.

\subsection{Three Bayesian learners, none of which has learned anything}
\label{subsec:three-learners}

Each of the three mechanisms below is implemented, is reached on a live call path,
and has never acquired a preference between the options it exists to choose among. The numbers were derived directly from the stores
rather than from any monitoring surface.

\textbf{These figures come from the authors' own operating stores, which a reader
does not have.} They are not harness output and are not reproducible from the
repository; they are a frozen reading taken at a stated instant
(2026-08-23T18:46:39Z) from stores that continue to accumulate, so a later reading
of the same stores will differ---the released evidence bundle contains an earlier
snapshot taken the same evening at 17:55Z whose counts are correspondingly lower,
and the gap between them is ninety minutes of ordinary operation rather than a
discrepancy. The re-derivation queries are released with the
evidence bundle so the method is checkable even where the data is not. What
\emph{is} reproducible by anyone is the mechanism: both invariants of
\Cref{subsec:invariants} ship in the package, run against any store, and will
report the same verdicts on any deployment in the same condition.

\paragraph{The channel-selection bandit.}
Thompson sampling over Beta posteriors, one arm per retrieval-weight
configuration, updated as $\alpha \mathrel{+}= r$, $\beta \mathrel{+}= (1-r)$.
Across three independent stores---a live working store, an archived export taken a
month earlier, and an unrelated instrumented store---every arm satisfies
\begin{equation}
\alpha - \alpha_0 \;=\; \beta - \beta_0 \;=\; n/2
\label{eq:neutral-identity}
\end{equation}
\emph{exactly}, on \textbf{459 of 459 arms} over \textbf{5{,}657} recorded plays.

What \Cref{eq:neutral-identity} establishes needs stating precisely, because it is
weaker than it first looks and still sufficient. Since $\alpha-\alpha_0 = \sum_i
r_i$ and $\beta-\beta_0 = n - \sum_i r_i$, the identity says only that
$\sum_i r_i = n/2$: the rewards \emph{average} exactly one half. It does
\emph{not} follow that each individual reward was $0.5$---an alternating stream of
zeros and ones satisfies it identically, and so does any symmetric mixture. So the
identity alone does not prove the reward channel was emitting a constant.

It proves something sufficient anyway, though not what we first said it proved.
Every arm's posterior mean is exactly $0.5$, the prior mean, so no arm has acquired a
preference over any other. We previously wrote that a Thompson sampler in this state
``selects uniformly at random forever''. That is wrong, and we keep the error here
because it has the same shape as everything else in this section.

A neutral update does not leave a posterior where it started. Both parameters grow,
so $\mathrm{Beta}(1{+}n/2,\,1{+}n/2)$ holds the mean at one half while the variance
falls. Arms are played different numbers of times, so their posteriors concentrate by
different amounts, and a sampler drawing one value from each does not choose among
them evenly. Drawing $10^{6}$ samples from $\mathrm{Beta}(1,1)$,
$\mathrm{Beta}(10,10)$ and $\mathrm{Beta}(100,100)$---which share a mean of exactly
one half---the argmax falls to them in proportion $0.454 / 0.291 / 0.255$.

So the learner is not standing still. It is converging, with increasing confidence,
on the conclusion that every arm is average, and what remains of its exploration is
steered by how often an arm happened to be played rather than by anything it
observed. We think that is worse than a frozen learner, and we did not see it until
someone checked the arithmetic rather than the claim.

And the exactness is itself the evidence. A genuine reward stream would have to
land on $\sum_i r_i = n/2$ to the last representable digit on all 459 arms
independently, across three stores with different play counts. Where per-observation
data exists we can close the gap directly rather than by inference: the source-trust
learner's observation table records \textbf{629} rows and its distinct reward set is
exactly $\{0.5\}$---every single observation is the neutral fallback. The bandit has
no comparable per-play record, and that absence is itself the second finding below.

The settlement ledger holds 4 rows against those 5{,}657 arm-plays, and none is
settled. The two counters are not inconsistent: a play increments its arm and
separately writes a ledger row that a later authenticated outcome resolves. The
row-writing was disabled at the only caller on 2026-07-27, together with the
per-exposure signal enqueue it shared a flag with---the same date on which the
arms' last-played timestamps stop. Nothing was written for the reward path to
settle, so every play applied the neutral prior and none was ever revised.

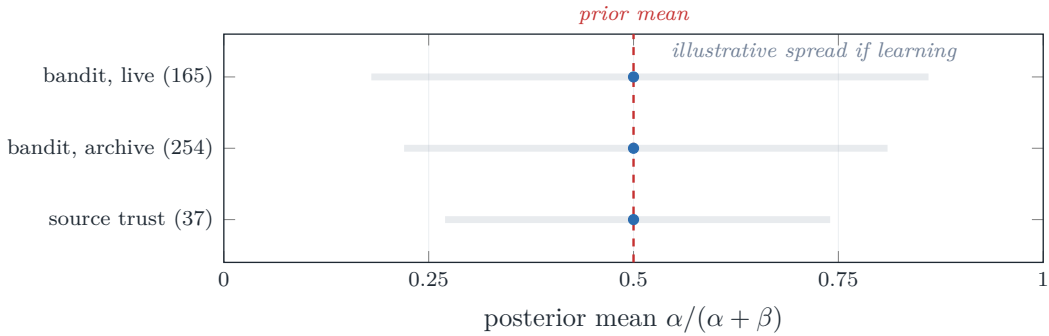
\begin{figure}[htbp]
\centering
\begin{tikzpicture}
\begin{axis}[
    width=0.78\linewidth, height=4.6cm,
    xmin=0, xmax=1,
    ymin=-0.6, ymax=2.6,
    xtick={0,0.25,0.5,0.75,1},
    xticklabels={$0$,$0.25$,$0.5$,$0.75$,$1$},
    ytick={0,1,2},
    yticklabels={%
      {\scriptsize source trust (37)},
      {\scriptsize bandit, archive (254)},
      {\scriptsize bandit, live (165)}},
    xlabel={\small posterior mean $\alpha/(\alpha+\beta)$},
    axis line style={draw=slmink, line width=0.5pt},
    tick label style={font=\scriptsize, text=slmink},
    label style={font=\small, text=slmink},
    ymajorgrids=false, xmajorgrids=true,
    grid style={draw=slmgraybg, line width=0.4pt},
    clip=false,
]
\addplot[draw=slmred, line width=0.9pt, dashed] coordinates {(0.5,-0.6) (0.5,2.6)};
\node[anchor=south, font=\scriptsize\itshape, text=slmred] at (axis cs:0.5,2.6)
  {prior mean};
\addplot[only marks, mark=*, mark size=1.9pt, draw=slmblue, fill=slmblue]
  coordinates {(0.5,2) (0.5,1) (0.5,0)};
\addplot[draw=slmgray, line width=2.6pt, opacity=0.16]
  coordinates {(0.18,2) (0.86,2)};
\addplot[draw=slmgray, line width=2.6pt, opacity=0.16]
  coordinates {(0.22,1) (0.81,1)};
\addplot[draw=slmgray, line width=2.6pt, opacity=0.16]
  coordinates {(0.27,0) (0.74,0)};
\node[anchor=south, font=\scriptsize\itshape, text=slmgray]
  at (axis cs:0.72,2.08) {illustrative spread if learning};
\end{axis}
\end{tikzpicture}
\caption{Three Beta populations, and the entire observed range of each. Every unit
in all three sits at posterior mean exactly $0.5$---the prior mean---so each
population's range is a single point rather than an interval. Live bandit: 165 arms
over 1{,}405 plays. Archived bandit from a store captured a month earlier: 254 arms
over 1{,}394 plays. Source trust: 37 sources over 629 observations whose recorded
reward is $0.5$ in every row. The shaded bands show, for scale only, where a
population receiving differential signal would spread; no measurement is plotted
there. Frozen reading, 2026-08-23T18:46:39Z, from the authors' own operating stores
(\Cref{subsec:three-learners}); these stores continue to accumulate and a later
reading will differ.}
\label{fig:learners-at-prior}
\end{figure}

\paragraph{The source-trust model.}
A second Beta posterior, one unit per evidence source. Identical signature: 37 of
37 sources at posterior mean $0.5$, across \textbf{629} observations whose recorded
reward is $0.5$ in every single row. This learner was still writing observations during
the preparation of this section.

\paragraph{Trust-weighted forgetting.}
Our prior work~\citep{slmv3} presented a trust-modulated retention rate,
$\lambda_{\mathrm{eff}} = \lambda\,(1 + \kappa\,(1-\tau))$, as delivered. The
formula is implemented, and it is called for every fact on the retention path. It
has never had an effect. Its enclosing scheduler asks the schema a question before
performing the per-fact trust lookup---does the trust table exist, and does the
facts table carry an author column---and falls back to $\tau = 1.0$ when the
answer is no. With $\tau = 1.0$ the expression reduces to $\lambda_{\mathrm{eff}}
= \lambda$: not merely disabled, but \emph{arithmetically identical} to a system
that never had the feature.

The falling back is correct behaviour; it is what keeps an older store openable.
What makes this worth reporting is where the missing data turned out to be. The
author column is absent from the table the guard tests, and present---populated on
every one of \textbf{4{,}340} rows---on a neighbouring provenance table that a
different consumer in the same package already reads successfully. The same data,
two consumers, one finds it and one does not.

That is the finding, and the obvious remedy is smaller than it looks. Those 4{,}340
rows cover \textbf{3{,}766 of 5{,}392 facts, or 69.8\%}: re-keying the join would
make the path effective for roughly seven facts in ten and leave the rest on the
same fallback. Of the authors recorded, \textbf{3{,}997} are opaque capability
digests and \textbf{237} are literally \code{unknown}, leaving 106 that name a
readable principal. So this is a partial repair and a backfill decision, not a free
fix, and the identity it would restore is machine-scoped rather than person-scoped.

We report that distinction because our own check got it wrong first. The
join-liveness invariant of \Cref{subsec:invariants} initially reported the
populated row count and concluded ``no migration and no backfill''---true about the
column, misleading about the join. It now measures coverage over the guarded table
and says so. A diagnostic that overstates its own remedy is an instance of the class
it exists to detect, which is not a comfortable thing to publish and is the reason
to publish it.

\subsection{Why the learners' own metrics cannot see this}

All three failures are invisible from inside the mechanism, for the same reason in
each case: the signal that would reveal the failure is derived from the mechanism
that failed.

A stalled bandit still records plays, still advances its timestamps, and still
reports a posterior. Its play count is not merely uninformative about learning; it
is the thing that keeps rising while the posterior concentrates on a preference it
never acquired. A guard that
never passes raises nothing, appears in coverage, and is reachable by any
call-graph trace. And a neutral fallback is the worst possible default for
observability, because it composes into an identity: a decay multiplier that falls
back to $1$ and a reward that falls back to $\tfrac{1}{2}$ both produce output
indistinguishable from correct operation.

This generalises past this system. \emph{Implemented}, \emph{reachable} and
\emph{effective} are three different questions. A grep answers the first. A
call-graph trace answers the second. Neither answers the third, and only querying
the store against real data does.

\subsection{Two invariants that answer the third question}
\label{subsec:invariants}

Both are read-only and on no hot path, and both are released in 4.1.5, reachable
from the command line as a diagnostics subcommand. They are \emph{not} in the 4.1.3
artifact whose measurements this paper reports---that release predates them---so the
version of record for the invariants is one release later than the version of record
for the measurements. We would rather state that than blur it: the numbers in
\Cref{subsec:three-learners} were taken with the checks run against 4.1.3 stores from
a working tree, and anyone re-running them today does so with 4.1.5 installed. They are deliberately mechanical: each is a few lines, each returns a
verdict rather than a score, and neither requires a model.

\paragraph{Prior distance.} For each Bayesian learner, assert that the posterior
has moved away from its prior \emph{mean} after $n$ observations. The check reports
\textsc{stalled} when \Cref{eq:neutral-identity} holds on every unit,
\textsc{moving} when any unit has diverged, and \textsc{insufficient\_data} below
an observation floor---because a cold learner sitting at its prior is correct, and
a diagnostic that fires on correct behaviour is how a real warning stops being
read. Motion off the prior is necessary for learning, not sufficient; the check
distinguishes \emph{receiving signal} from \emph{receiving nothing}, which is
precisely the distinction the learner's own counters cannot make.

\paragraph{What the prior-distance check gets wrong.} Disclosing a diagnostic's
boundaries is part of shipping one, and this one has three.

It reports \textsc{stalled} on \emph{symmetric} signal, not only on absent signal:
a learner receiving genuinely balanced outcomes---an even split of good and bad per
arm, which a 50/50 comparison produces---satisfies \Cref{eq:neutral-identity} and
is reported as receiving nothing. The check separates \emph{a posterior that has
not moved} from one that has, which is not the same as separating signal from
silence, and an operator seeing \textsc{stalled} must still ask which they have.
Second, the aggregate verdict loses per-arm detail in the direction that matters: it
returns \textsc{moving} if \emph{any} unit has diverged, so 164 stalled arms beside
one live arm reports as moving. Third, the update applies a ceiling
(\code{MIN(cap, alpha + r)}), and an arm pinned at that ceiling no longer satisfies
the identity even if every reward it ever received was neutral---a false negative.
A fourth mode was found by this paper's own second-round review: an
observation floor applied in aggregate lets a store with many units and few
observations report \textsc{stalled} on units that were simply never played, so the
floor now binds per unit as well. A fifth is the one the name invites: the check
compares each unit's posterior \emph{mean} against the prior mean, to within
$10^{-9}$. It does not compare distributions. Two units at the same mean and
different observation counts are different posteriors and will be sampled
differently, and this check cannot tell them apart. ``Prior distance'' is therefore a
convenient name for a narrower test than it sounds like, and the verdict it earns is
\emph{no learned preference}, not \emph{unchanged}. Its registry is also explicit rather than
exhaustive: it inspects the two Beta learners named in it, not every table in the
store that might be learning something.

\paragraph{Join liveness.} For each schema-guarded path, assert that the guard's
requirements are present in this store, and when one is not, search the remaining
tables
for the required column and report the row count found. The verdicts are
\textsc{live}, \textsc{dead}, and---the one that matters---%
\textsc{satisfied\_elsewhere}. The evidence is asymmetric and the check should be
read that way: a satisfied guard means the path is \emph{reachable}, not that it has
ever run, whereas an absent requirement on a store that has been in service does
establish that the path cannot have run against it. The findings rest only on the
second direction. A dead path reported alongside the populated table
that already holds its data is a one-line fix with no migration and no backfill; a
dead path reported alone is a bug report. The check also records what the feature
computes when the guard fails, because ``switched off'' and ``computing exactly
what being switched off would compute'' are different facts for an operator.

\subsection{What they found}

Pointed at the authors' own store, with no configuration and no hints:

\begin{table}[h]
\centering
\caption{Both invariants, first run, against a live store.}
\small
\begin{tabular}{llr}
\toprule
\textbf{Check} & \textbf{Verdict} & \textbf{Evidence} \\
\midrule
prior distance, channel bandit & \textsc{stalled} & 165/165 units, 1{,}405 obs \\
prior distance, source trust & \textsc{stalled} & 37/37 units, 629 obs \\
join liveness, trust forgetting & \textsc{satisfied\_elsewhere} & 4{,}340 rows, 69.8\% coverage \\
\bottomrule
\end{tabular}
\end{table}

The third row is the result we did not anticipate. The check was written to report
that a guarded path had never run; it also identified the table holding the data,
the number of populated rows, and the absence of any migration requirement. The
remedy was produced by the diagnostic, not by the engineer reading it.

\subsection{From verdict to named cause to repair, and a test of the repair}
\label{subsec:repair}

A diagnostic earns its place by what follows it firing. The \textsc{stalled}
verdict prompted an investigation of the learning path, that investigation found
defects, and two of them were repaired in public releases. We want to state the
strength of that chain accurately, because it is weaker than a clean
detect-to-repair story and still worth reporting.

What we can show: a shipped check returned a concrete verdict on a real store; work
followed it; the defects that work found are named below and their fixes are
installable. What we \emph{cannot} show is a controlled attribution. The frozen state
of \Cref{subsec:three-learners} covers 459 arms over 5{,}657 plays whose last activity
is 2026-07-27, and this paper attributes that state to ledger writing being disabled
at its only caller. The session-identifier defect below was measured later on a
different population. The post-repair reading later still covers 165 arms. These are
three overlapping observations of one subsystem at three times, not one population
followed through a repair. We ran no replay and no single-factor ablation, so we
cannot say which defect produced which part of the historical state, and a reader
should not take the sequence below as that claim.

\paragraph{One defect the investigation found.} The bandit records a play and
separately writes a ledger row that a later authenticated outcome resolves. Outcome
matching keys on the conversation a recall belonged to. The component that mints
session identifiers and the component that matches them did not agree on which names
counted: recalls arriving through an agent front-end's subprocess were filed under
identifiers the matcher deliberately excludes, because those identifiers are the
placeholders a surface invents when no caller supplies one, and a placeholder must
never be mistaken for a real conversation. Both halves were correct in isolation.
Their intersection was empty, so no engagement signal could attach to any play.

The shape is what interests us, and it does generalise: \textbf{two components can
each enforce a correct rule and still compose into a mechanism that cannot fire.}
Neither component is wrong, so neither has a failing test, and the composition is
what no unit test and no coverage report examines. We report the shape as an
observation from one system, not as a measured prevalence.

\paragraph{Why a neutral default made it worse.} A play that ended with no
evidence was settled at the neutral reward rather than left unsettled. Under
$\alpha \mathrel{+}= r$, $\beta \mathrel{+}= (1-r)$, a reward of exactly
$\tfrac{1}{2}$ adds $\tfrac{1}{2}$ to each parameter: the posterior mean does not
move and the posterior \emph{variance falls}. Repeated neutral settlement
therefore does not leave a learner undisturbed---it makes the learner steadily
more confident that its arms are average, and \emph{quantifiably} harder for
genuine evidence to move. From $\mathrm{Beta}(1,1)$, one genuine reward of $1$
shifts the posterior mean by $0.167$; after two hundred neutral settlements the
same reward shifts it by $0.0025$. That is a \textbf{67-fold} loss of
responsiveness to real evidence, accumulated while the mean never moved and every
counter the subsystem keeps continued to rise---which is why no threshold on those
counters could have caught it. \textbf{Neutral is a commitment, not an abstention.} A learning
system with no way to express \emph{no evidence} will convert its own silence into
confidence, and will do so while every counter it keeps continues to rise.

\paragraph{The repairs.} Three changes, public in 4.1.6 and 4.1.7, with a fourth
withdrawn in 4.1.8 and discussed below. The host
variable the front-end actually sets is now among the names the resolver checks.
Every synthetic identifier is now minted by one function and registered as synthetic,
so the matcher recognises the ones that were previously minted ad hoc and silently
accepted as real. And a play that observed nothing is closed as \textsc{unobserved}
with its posterior left untouched. A fourth change turned adaptive ranking on by default,
where it had sat behind an unset variable that kept the selector out of the call path
entirely on a default install. \textbf{That fourth change was reverted one release
later}, and the reason is the mechanism this section has just described. With the
selector live and most arms holding no settled reward, the weighting sampled for a
query came from posteriors that had observed nothing, so two identical questions could
weigh semantic against lexical evidence differently for no reason drawn from anything
observed. The release notes withdrawing the default say exactly that. We record it
because it is the same finding arriving from the opposite direction: the paper argues
that an unsettled posterior is not a neutral one, and a shipped default had to be
withdrawn because it is not. Adaptive ranking is opt-in again, and turning it on
before the loop supplies real rewards trades determinism for nothing.

The synthetic namespace still exists, and should: an identifier a surface invented is
not a conversation, and the matcher is right to refuse it. What was wrong was that
some inventions were not declared as such.

\paragraph{What changed, measured.} Read against the authors' own store at
2026-08-25T01:23Z under 4.1.7: of \textbf{60} recorded plays, \textbf{40} closed
\textsc{unobserved} with posteriors untouched---a state the earlier build could
not produce, because it had no way to decline---\textbf{12} settled from a
reported outcome, and 8 remained open. The mechanism that guaranteed neutrality is
gone, and the abstention that replaced it is observable in the ledger.

\paragraph{The controlled result.} Field state cannot separate ``the repair worked''
from ``nobody used it''. A controlled ablation can, and this one changes exactly one
factor: the identifier namespace a recall is filed under. Everything else---workload,
corpus, simulated engagement, seed, and every production module in the path
(\code{ContextualBandit}, the \code{is\_conversation} predicate the hot path applies,
\code{record\_recall}, the engagement feature extractor, the reward model and the
settlement pass)---is held identical. \Cref{tab:exp12} reports three arms over 120
recalls each.

\begin{table}[htbp]
\centering
\caption{exp12, closed-loop ablation. One varied factor: the session identifier a
recall is filed under. \slm{} v4.1.9, seeded; repeated runs give identical
verdicts and an identical maximum shift, with the instantiated-arm count varying
73--82 because arm selection is stochastic.}
\label{tab:exp12}
\footnotesize
\begin{tabular}{@{} p{3.5cm} r r r r @{}}
\toprule
\textbf{Arm} & \textbf{Tickets} & \textbf{Settled} & \textbf{Arms moved} &
\textbf{max $|\bar\theta - \tfrac{1}{2}|$} \\
\midrule
Defect: synthetic id      & 0   & 0   & 0 of 0   & 0.000 \\
Repaired: conversation id & 120 & 120 & \textbf{82 of 82} & \textbf{0.343} \\
Control: no engagement    & 120 & 0   & 0 of 0   & 0.000 \\
\bottomrule
\end{tabular}
\end{table}

With the defect present, no outcome ticket is written, nothing settles, and no
posterior moves. With it absent and the agent acting on what it was shown, every
settlement is drawn from observed engagement (\textsc{artifact\_overlap}) and every
instantiated arm moves off its prior mean, the largest by $0.343$. \textbf{The loop
closes, and the defect is what was holding it open.}

The third arm is the one that makes the second mean anything. It writes all 120
outcome tickets and differs only in that the agent then acts on something unrelated.
Nothing settles and no posterior moves. A loop that reported learning there would be
manufacturing reward from the existence of a ticket rather than from evidence, which
is the failure this paper spent nine pages describing, and it would have been easy to
ship and easy to mistake for success.

\paragraph{The same repair, observed in deployment.} An ablation shows a mechanism
can work. It does not show that the shipped fix reaches a real installation. The
authors' own store answers that directly, because the quantity the repair governs is
observable in it: which namespace a recall's session identifier came from. Before the
repair, on 2026-08-23, 188 of 188 recalls were filed under synthetic identifiers that
the matcher is designed to refuse. On 2026-08-24, the day the repair shipped, 142
synthetic and 9 conversation identifiers---the changeover, mid-day. On 2026-08-25,
\textbf{0 synthetic and 7 conversation}. A cross-tabulation of the two tables the join
depends on shows the earlier state exactly: \textbf{zero} session identifiers appear in
both \code{pending\_outcomes} and the tool-event log across the whole history, which is
why no engagement signal had ever attached to a play. The defect and its repair are
both visible in the same column, and neither is inferred.

\paragraph{A fourth instance, found after this draft was frozen.} Applying the same
question to the other settlement path found the class again. The reward label is
computed as $0.5$ plus bonuses minus penalties, so an outcome carrying no engagement
signal evaluates to exactly one half---and that value was persisted, becoming a
mid-strength positive training label downstream and a Beta update that tightens a
posterior around its prior. On the authors' store, \textbf{492 of 492} recorded
outcome scores were exactly this value, so not one of them reflected an observation.
The batch settlement pass already declined to score an unsignalled row; the
single-outcome path did not, and the two had disagreed since the mechanism shipped.
Both abstain as of 4.1.9. We report this as an eleventh instance and deliberately do
\emph{not} renumber the ten of \Cref{sec:eval:taxonomy}: that count was frozen when
the taxonomy was written, and the fact that the method kept finding instances
afterwards is better evidence for the method than a larger number would be. It is
also the sharpest illustration of the shape, because here the neutral value did not
merely fail to inform the learner---it trained a failure as a moderate success.

\paragraph{What we still do not claim.} A controlled ablation is not a deployment
result. It shows the mechanism can transport a signal from an agent's behaviour to a
posterior, under engagement we generated; it does not show that the preferences a
real deployment would learn are ones an operator would endorse, and it does not
measure retrieval quality. All \textbf{165} arms in the authors' live store remain
at their prior. The twelve settled plays each carried a reported reward of exactly
$\tfrac{1}{2}$, so they moved no mean. \emph{The barrier to learning is identified,
named and removed; the learning is not demonstrated.} Posterior movement requires
sustained real engagement that a single-operator store over days does not supply,
and we would rather publish that distinction than a curve. What this section does
establish is narrower and, we think, more useful: an invariant that asks whether a
mechanism \emph{takes effect} located a defect that 459 healthy-looking arms,
5{,}657 recorded plays, and every counter in the subsystem reported as normal
operation---and the repair it led to is in a public release a reader can install
and inspect.

\subsection{What does hold}

Two admission boundaries on this path are enforced, and it is worth separating
them from the three that are not.

\paragraph{Completion is decided by an independent gate.} The bounded-loop engine
terminates on a verdict from a checker the agent cannot edit or grade. The agent's
own assertion of completion is recorded in the ledger and is \emph{never read}
when deciding to terminate---the field appears only in assignments, never in a
conditional. Each run leaves an append-only hash-chained ledger that can be
re-verified after the fact, which makes it a fourth verifiable artifact alongside
the completion manifest, the erasure receipt, and the audit chain. It is not part
of the write transaction and does not participate in the completion manifest;
execution verification and write verification are separate chains.

\paragraph{Unverifiable evidence is admitted for display and barred from
learning.} External gate-verified execution receipts enter through a versioned
contract. Where the producing tool cannot itself prove the receipt, the record is
persisted with an explicit trust level and an \code{eligible\_for\_learning} flag
set false: it may be displayed and observed, and it cannot promote a memory, alter
learning, or assert execution authority. The tier is set by the
\emph{verifiability} of the evidence rather than by its content or the reputation
of its source. The package deliberately neither imports the producing tool nor
reimplements its log grammar; the tool's own command is the versioned protocol
port. At the time of writing this boundary has been exercised once, by a
demonstration run, which is the honest extent of the claim.

\subsection{What is not governed}

\begin{itemize}[leftmargin=*,itemsep=1pt]
  \item The three mechanisms in \Cref{subsec:three-learners} were instrumented and
        inert at the version whose measurements this paper reports. Nothing here
        should be read as a claim that any of them works, before or after the
        repairs of \Cref{subsec:repair}.
  \item The skill-evolution pipeline is implemented, gated, and has never
        executed. Parent-version lineage and generation tracking are not reported.
  \item The invariants in \Cref{subsec:invariants} are \emph{admission} claims,
        not outcome claims. We present no evidence that gated learning produces
        better agent behaviour, because we have run no such experiment. A reader
        who takes this section as a retrieval-quality or task-success result has
        been misled.
  \item The loop ledger is verifiable but is not bound into the completion
        manifest.
  \item \textbf{Two of the ten shapes have a mechanical detector; eight do not.}
        The invariants cover Bayesian learners that have not moved and
        schema-guarded paths whose requirements are absent. A liveness probe reading
        its own output, an availability check on a re-violable invariant, a
        distribution channel forking its artifact, and a metric differencing two
        arms whose costs were never comparable are each named in \Cref{sec:eval:taxonomy} and each
        still require a human to notice. Mechanising more of them is the obvious
        continuation and we make no claim to have done it.
  \item Two of the three mechanisms have had their named cause repaired in
        4.1.6--4.1.9 (\Cref{subsec:repair}); \emph{repaired} here means the
        barrier is removed, not that either learner has been shown to learn. The
        third---re-keying the trust join onto the provenance table---is scoped
        work we have not done, and \Cref{sec:futurework} says so rather than
        implying otherwise.
\end{itemize}

\section{System Model and Verified Design Invariants}
\label{sec:formal}

The governing invariant states, as a design goal, that a memory operation yields one
authenticated actor, one profile generation, one policy decision, one durable receipt,
and one verifiable completion state across the registered projection owners. V4
realises this full envelope on the canonical HTTP \code{/remember} path and internal
ingestion; declared MCP tools and selected CLI mutations obtain a shared policy
decision without the full envelope. We label the five scoped correctness properties
below \emph{Verified Design Invariants} (VDIs). This section is an assurance case,
not a proof development, and the distinction is deliberate rather than apologetic.
The definitions are formal: they fix the state a claim quantifies over, so that a
reader can say precisely what would falsify it. The invariants are not deduced from
axioms; each is a property of the canonical write/read path established by an
implementation trace and measured directly. We chose that instrument because the
claims here are about what a running system does to a physical store, and a proof
about a model of that system would establish a different thing. Each justification
traces the shipped code, and measured evidence for each invariant appears in
\Cref{sec:evaluation}. A reader wanting deduced theorems should read this section as
the falsification conditions for the evaluation that follows, not as a substitute for
it.

\begin{definition}[Actor context]
An \emph{actor context} is the frozen tuple
$A = (p, R, P_{\mathrm{allow}}, \pi, g, \Sigma, \delta, \tau, h)$
where $p$ is the server-derived principal (never from the request body), $R$ is the
frozen role set, $P_{\mathrm{allow}}$ is the profile allowlist, $\pi$ is the active
profile ID, $g$ is the active generation, $\Sigma$ is the scope allowlist, $\delta$
is the transport channel, $\tau$ is the resolved client host, and $h$ is the SHA-256
prefix of the session token (first 16 hex chars; no raw session material retained).
$A$ is authenticated iff $p \neq \varepsilon$ and ANONYMOUS $\notin R$.
\end{definition}

\begin{definition}[Projection obligation and owner]
A \emph{projection obligation} records
$O = (\mathrm{op}, \rho, \omega, s, v, c)$ where
$\rho$ is the owner name, $\omega \in \{\mathrm{APPLY}, \mathrm{ERASE}\}$,
$s \in \{\mathrm{PENDING}, \mathrm{APPLIED}, \mathrm{VERIFIED}, \mathrm{FAILED},
\mathrm{COMPENSATED}, \mathrm{ERASED}\}$, and $c$ is an owner-specific SHA-256
checksum. Terminal successes: $\{\mathrm{VERIFIED}, \mathrm{ERASED}\}$.
The three registered admission owners are
$\Omega = \{\mathrm{bm25}, \mathrm{temporal}, \mathrm{vector}\}$.
\end{definition}

\begin{definition}[Completion manifest]
A \emph{completion manifest} is the frozen record
$M = (\mathrm{op}, \pi, \sigma, b, n, \mathcal{E}, h_M, t_c, t_u)$
where $\sigma \in \{\mathrm{COMPLETE}, \mathrm{DEGRADED}, \mathrm{FAILED}\}$,
$b$ is the \code{all\_met} flag, $\mathcal{E}$ is the sorted tuple of
\code{OwnerEvidence} dicts, and $h_M$ is the envelope hash: on \mbox{M037}-capable
databases an installation-key HMAC-SHA256 (version~2, refusing a version-1 downgrade);
on older schemas, an unkeyed SHA-256 self-consistency check.
\end{definition}

\begin{theorem}[Bounded In-Process Generation-Epoch Rejection]
\label{thm:fence}
\textbf{Claim.} If the binding epoch increments from $g_0$ to $g_1 > g_0$ after
an admission event $(k, \pi, g_0, t_0)$ is recorded but before
\code{\_handle\_admission} executes, the handler raises
\code{ValueError("epoch is stale")} \emph{before} any projection write occurs.
\textbf{Scope:} in-process, single-writer, 300-second TTL, not CAS, not
SQLite-backed, not persistent across restarts.\\
\textbf{Implementation trace.} Under \code{\_binding\_lock}, \code{admitted\_epoch} returns
$g_0 \neq g_1$, so the guard fires before \code{command\_impl.submit()} is reached;
conflict epoch $-1 \neq g_1 \geq 0$ likewise fires.
Measured: exp7, 199/200 (\Cref{sec:evaluation}); the single failure was a transient writer unavailability on the positive control, not a fence misjudgement, and it is analysed in \Cref{sec:eval:results}.
\end{theorem}

\begin{theorem}[Completion-Manifest Integrity and Honest State]
\label{thm:manifest}
\textbf{Claim.} (i) On \mbox{M037}-capable databases the manifest envelope is sealed with
an installation-key \mbox{HMAC-SHA256} (version~2, refusing a version-1 downgrade at
verification): a DB-only writer without the installation key cannot forge a seal that
\code{verify\_manifest} accepts (authenticity, not merely corruption detection). On older
schemas the seal is an unkeyed \mbox{SHA-256} that detects accidental corruption (bit
flips, partial writes) but is not tamper-evidence against a malicious writer.
(ii) \code{all\_met = True} iff every obligation is in terminal-success state; a
\textsc{degraded} manifest is never reported as success.\\
\textbf{Scope:} authenticity holds under installation-key secrecy on M037; legacy unkeyed
v1 is corruption-detection only; an attacker holding the installation key is outside the
model.\\
\textbf{Implementation trace.} \code{verify\_manifest} recomputes the versioned seal over the canonical
JSON fields and refuses a v1 downgrade on M037; HMAC unforgeability (resp.\ SHA-256
collision resistance) yields an invalid/different digest on any single-field mutation.
\code{derive\_state} maps \code{all\_met=True} iff all obligations are in
$\{\mathrm{VERIFIED}, \mathrm{ERASED}\}$, any other state routing to DEGRADED.
Measured: exp2, 200/200.
\end{theorem}

\begin{theorem}[Covered HTTP Read Profile Isolation]
\label{thm:isolation}
\textbf{Claim.} On covered HTTP read paths, an actor lacking READ permission for the
active profile $\pi_A$ is rejected before any query executes; an authorized actor
cannot retrieve private rows (\code{profile\_id = $\pi_B \neq \pi_A$}).\\
\textbf{Scope:} authorization-layer isolation via SQL predicates and RBAC on covered
HTTP paths; MCP stdio and CLI direct-call paths are not covered by this gate; direct
SQLite file access is outside the threat model.\\
\textbf{Implementation trace.} Two independent controls. (a) \emph{Storage predicate}---covered read
paths parameterize every query with the actor's \code{profile\_id} from the server-side
\code{ActorContext}, so a query for $\pi_A$ never returns $\pi_B$ rows; \emph{measured} by
exp5 with a positive control (200/200) on two read paths, \code{get\_facts\_by\_ids} and
\code{TemporalChannel}. (b) \emph{HTTP RBAC middleware}---the read gate rejects
unauthorized actors with 401/403 before a query runs on enumerated sensitive routes; this
is established by \emph{code inspection} of the middleware, not by exp5. An end-to-end
unauthorized/authorized HTTP probe is future work.
\end{theorem}

\begin{theorem}[Erasure Completeness over Registered Owners]
\label{thm:erasure}
\textbf{Claim (conditional).} If the returned receipt has \code{all\_erased=True},
then at \code{finalize()} time each owner $o \in \Omega$ returned \code{erased=True}
with empty residue. Partial residue sets \code{all\_erased=False}; the receipt is
honest about the gap.\\
\textbf{Scope:} limited to the three registered owners (bm25, temporal, vector)
present in \code{ErasureService.\_owners} at erasure time; projection stores not
wired as registered owners are not covered.\\
\textbf{Implementation trace.} \code{finalize()} calls \code{prove\_erased()} per owner---a live
physical re-query---and seals over all fields with HMAC-SHA256 (v2 on M037); refusing
a v1 downgrade makes the receipt unforgeable by a DB-only writer without the
installation key. Measured: exp1, 200/200.
\end{theorem}

\begin{theorem}[As-Of Temporal Demotion and Retention Non-Increase]
\label{thm:temporal}
\textbf{Claim (a).} A system-invalidated fact ($t_{\mathrm{si}} < +\infty$) receives
score factor 0.25; after per-channel re-sort it ranks strictly below any fact with the
same positive raw score and demotion factor 1.0.\\
\textbf{Claim (b).} The production Ebbinghaus function $R(t) = e^{-t/S}$ in
\code{ebbinghaus.py} is non-increasing in $t \geq 0$ for every $S \geq S_{\min} > 0$.\\
\textbf{Scope:} with \code{as\_of}$=q$, supersession demotes only when it is visible at
$q$ (\code{system\_expired\_at}$\le q$), and half-open event-validity
(\code{valid\_from}$>q$ or \code{valid\_until}$\le q$) applies a separate $0.5$ factor.
Demoted facts remain present and re-ranked---\code{as\_of} is point-in-time
\emph{demotion}, not snapshot isolation---and the filter fails open on error.\\
\textbf{Implementation trace.} (a) \code{temporal\_validity\_filter.py} multiplies system-invalidated
scores by 0.25 and re-sorts descending; $0.25s < 1.0s$ for any $s > 0$.
(b) \code{math.exp(-t/S)} is non-increasing in $t$; IEEE 754 underflow eventually
clamps to 0.0, making $R$ non-strictly constant there. Measured: exp6a, 200/200.
\end{theorem}

\section{Threat Model and Capability Coverage}
\label{sec:threat}

\slm{}'s threat model is bounded to adversaries operating above the OS boundary or
within the runtime's own failure modes. Isolation is an authorization-layer control;
a single OS user with direct access to the SQLite store files is outside this model.

\begin{figure}[htbp]
  \centering
  \resizebox{\ifdim\width>\linewidth\linewidth\else\width\fi}{!}{%
\begin{tikzpicture}[
      node distance=0.5cm and 0.3cm,
      ghdr/.style={rectangle, rounded corners=2pt, minimum width=2.45cm,
        minimum height=0.7cm, align=center, font=\scriptsize\bfseries,
        inner sep=4pt},
      gent/.style={rectangle, rounded corners=1pt, minimum width=2.45cm,
        align=center, font=\tiny, inner sep=3pt, fill=white, draw=slmgray,
        line width=0.3pt, text width=2.3cm},
      rchip/.style={rectangle, rounded corners=1pt, minimum width=2.45cm,
        minimum height=0.5cm, align=center, font=\tiny\bfseries, inner sep=2pt},
    ]

    \node[ghdr, fill=slmbluebg, draw=slmblue] (h1)
      {Governance\\[-1pt]\&\ Isolation};

    \node[gent, below=0.35cm of h1] (e1a)
      {\textbf{exp1} \\ erasure completeness \\ BM25+Temporal\\ +Vector \\ tombstone + receipt};
    \node[gent, below=0.25cm of e1a] (e1b)
      {\textbf{exp5} \\ cross-tenant zero-leak \\ positive control \\ tested on 2 read paths};
    \node[gent, below=0.25cm of e1b] (e1c)
      {\textbf{exp8} \\ policy registry \\ OWNER/ADMIN/\\ MEMBER allow \\ VIEWER deny; auth-first};

    \node[rchip, fill=slmbluebg, draw=slmblue, below=0.3cm of e1c] (rchip1)
      {3 exps $\cdot$ 600 trials \\ rate: 1.000};

    \node[ghdr, fill=slmgreenbg, draw=slmgreen, right=0.3cm of h1] (h2)
      {Transactional\\[-1pt]Integrity};

    \node[gent, below=0.35cm of h2] (e2a)
      {\textbf{exp2} \\ manifest COMPLETE \\ all owners applied};
    \node[gent, below=0.25cm of e2a] (e2b)
      {\textbf{exp2 (fault)} \\ manifest DEGRADED \\ real \texttt{compensate()} \\ cleans applied owner};

    \node[rchip, fill=slmgreenbg, draw=slmgreen, below=0.3cm of e2b]
      {1 exp $\cdot$ 200 trials \\ rate: 1.000};

    \node[ghdr, fill=slmamberbg, draw=slmamber, right=0.3cm of h2] (h3)
      {Safe Evolution\\[-1pt]\&\ Recovery};

    \node[gent, below=0.35cm of h3] (e3a)
      {\textbf{exp3} \\ downgrade refusal \\ reader\_ver $>$ code ver \\ zero mutation};
    \node[gent, below=0.25cm of e3a] (e3b)
      {\textbf{exp4} \\ partial-restore rollback \\ live $\leftarrow$ pre-restore bytes \\ staging cleaned};

    \node[rchip, fill=slmamberbg, draw=slmamber, below=0.3cm of e3b]
      {2 exps $\cdot$ 400 trials \\ rate: 1.000};

    \node[ghdr, fill=slmbluebg, draw=slmblue, right=0.3cm of h3] (h4)
      {Temporal\\[-1pt]Mechanisms};

    \node[gent, below=0.35cm of h4] (e4a)
      {\textbf{exp6a} \\ superseded-fact \\ demotion 0.25$\times$ \\ re-ranked below valid};
    \node[gent, below=0.25cm of e4a] (e4b)
      {\textbf{exp6b} \\ Ebbinghaus decay \\ non-increasing in age \\ bounded {[0,1]}};
    \node[gent, below=0.25cm of e4b] (e4c)
      {\textbf{exp6c} \\ date-proximate \\ outranks distant \\ horizon excludes far past};

    \node[rchip, fill=slmbluebg, draw=slmblue, below=0.3cm of e4c]
      {3 exps $\cdot$ 600 trials \\ rate: 1.000};

    \node[ghdr, fill=slmgreenbg, draw=slmgreen, right=0.3cm of h4] (h5)
      {Admission\\[-1pt]Control};

    \node[gent, below=0.35cm of h5] (e5a)
      {\textbf{exp7} \\ generation fence \\ stale-epoch rejected \\ writer never called};
    \node[gent, below=0.25cm of e5a] (e5b)
      {\textbf{exp7 (fresh)} \\ fresh-epoch admitted \\ \texttt{runtime.remember()} \\ commits ok};

    \node[rchip, fill=slmgreenbg, draw=slmgreen, below=0.3cm of e5b] (rchip5)
      {1 exp $\cdot$ 200 trials \\ rate: 1.000};

    \begin{scope}[on background layer]
      \node[draw=slmink, rounded corners=4pt, fill=white,
            fit=(h1)(rchip1)(h5)(rchip5), inner sep=6pt,
            label={[font=\small\bfseries, text=slmink]above:%
              SLM~4.0 Reliability Evaluation: 10 Deterministic Property Scenarios}]
            {};
    \end{scope}

    \node[rectangle, rounded corners=2pt, draw=slmink, fill=slmgraybg,
          minimum width=13.2cm, minimum height=0.55cm, align=center,
          font=\scriptsize\bfseries,
          below=1.6cm of e1c, xshift=5.24cm]
      {All 10 scoped properties held in 200 repeated executions each;
       repetitions are flakiness checks, not independent statistical samples};

  \end{tikzpicture}}
  \caption{Failure-class taxonomy from \Cref{sec:evaluation}: the ten measured guarantees grouped
    into five categories that map to SLM~4.0's five technical areas.
    \emph{Governance \& Isolation} covers projection-store erasure completeness
    (exp1), cross-tenant zero-leak with positive control (exp5), and operation
    policy correctness with authentication-before-role-check (exp8).
    \emph{Transactional Integrity} covers completion manifest correctness and
    real \texttt{compensate()} behavior (exp2).
    \emph{Safe Evolution \& Recovery} covers migration downgrade refusal (exp3)
    and backup-restore atomic rollback (exp4).
    \emph{Temporal Mechanisms} covers superseded-fact demotion (exp6a),
    Ebbinghaus decay monotonicity (exp6b), and time-window inference (exp6c).
    \emph{Admission Control} covers generation-fenced admission rejection and
    fresh-epoch commit (exp7). Each scenario ran 200 deterministic repetitions;
    every scoped property held.}
  \label{fig:taxonomy}
\end{figure}

\paragraph{T1 --- Cross-tenant read.}
Read paths apply a canonical scope predicate enforcing \code{profile\_id} equality and the
personal/shared/global scope algebra at query time; cross-scope access requires explicit
shared membership (recorded at write time) or the \code{global} scope flag, default-deny.
The all-channel statement rests on call-path inspection and per-channel unit tests;
\textit{measured evidence} is exp5 (200/200 repeated executions), which exercises the scope predicate on
two paths (\code{get\_facts\_by\_ids} and the temporal channel) with a positive
control---it is not an empirical proof for every channel.

\paragraph{T2 --- Replayed or stale-epoch write.}
The generation fence records the admitted epoch at ingestion; a mismatch raises
\code{ValueError("epoch is stale")} before any projection writer is called.
\textit{Evidence:} exp7 (199/200 repeated executions; see \Cref{sec:eval:results}). \emph{Scope:} in-process, 300-second TTL,
on the canonical remember path only.

\paragraph{T3 --- Projection inconsistency.}
Each canonical write registers an obligation for the three registered owners and
records a \code{CompletionManifest} sealed with an installation-key HMAC-SHA256 (v2
on M037). If any owner fails to verify, the manifest carries an honest
\textsc{degraded} status; the maintenance loop redrives pending obligations.
\textsc{degraded} is never silently discarded or promoted to \textsc{complete}.
\textit{Evidence:} exp2 (200/200 repeated executions).

\paragraph{T4 --- Memory-content secret leakage.}
The shared queryable-write chokepoint applies \code{scrub\_secrets\_for\_ingest}
unconditionally before canonical and projection persistence. \emph{Disclosure:} mesh
coordination messages are not subject to the full admission scrub (fail-open); sender-side
filtering remains required for sensitive payloads.

\paragraph{T5 --- Migration downgrade or backup corruption.}
Two measured controls: a newer-stamped database is refused on the deferred migration pass
with the table-set preserved (\textit{measured:} exp3, 200/200), and a partial-restore
failure rolls live data back to pre-restore bytes (\textit{measured:} exp4, 200/200). The
Scale Engine's projection parity-verify-before-promote and rollback path is established by
code inspection; a fault-injection experiment driving prepare/verify/promote/rollback is
future work.

\paragraph{T6 --- Unauthorized skill mutation.}
Skill evolution is gated by a per-cycle budget and a blind-verified pipeline to
quarantine. The MCP entrypoint is gated by \code{OperationPolicyRegistry}
(\code{EVOLVE\_SKILL}, roles $\{\mathrm{OWNER, ADMIN}\}$). \emph{Residual gap:} the
MCP actor is synthesized from deployment mode with no live session principal, so no
human-approval step is interposed; \code{auto\_approve} is a configuration gate,
not a live-RBAC control.

\paragraph{T7 --- Mesh relay content exposure.}
Mesh messages are redaction-scrubbed on receipt but not admission-scrubbed (fail-open).
Cross-machine coordination uses leaderless LWW---not consensus, quorum, or a
replicated database---with TLS/pinning and HMAC transport available and ordinary
storage writes not invoking the fence primitive.

\paragraph{Deferred controls (D1--D3).}
Per-device mesh key revocation (future work); cryptographic multi-tenancy (delegated
to OS-level disk encryption); inference-time injection gating (complementary to
MemGate~\citep{memgate}, layerable above the retrieval surface).

\section{Implementation}
\label{sec:implementation}

\slm{} spans more than 25 subsystems
(\code{access}, \code{compliance}, \code{core}, \code{mesh}, \code{retrieval},
\code{storage}, \code{temporal} components in \code{encoding}/\code{retrieval},
\code{optimize}, \code{code\_graph}, \code{learning}, \code{trust}, and others),
exercised by a release gate whose exact collected, selected, passed, skipped, and
deselected counts are bound to the final source SHA, command, and result artifact in
the release manifest. The exact commands and environment are documented with the
released harness.

We separate two kinds of evidence here, because the bundle accompanying this paper
does not treat them alike. The eleven experiments of \Cref{sec:evaluation} ship with
complete per-experiment result artifacts, and every number this paper quotes from
them is reconstructable from those files. The repository-wide suite logs in the same
bundle are \emph{not} of that standard: one is truncated before its terminal summary
and two end on failing tests, all of which are the concurrency-contention artifacts
described in \Cref{sec:limitations}. We include them because omitting them would be
worse, and we state plainly that they substantiate the experiment results and not the
repository-wide gate wording above. A reader should treat the gate as a release
process we describe and the experiments as the evidence we offer.

The system runs in three privacy modes (A, B, C as described in
\Cref{sec:architecture}) and is delivered through a CLI (\code{slm} with agent-native
\code{--json}), an MCP server (HTTP and stdio, with configurable tool profiles from
\code{core}/14 tools to \code{whole}/all registered), a dashboard, a Python SDK,
editor and proxy integrations, and \textbf{nine separately packaged framework
adapters}. The adapters are packaged as standalone distributions outside the base wheel; in this
release the repository's adapter package metadata has been updated to admit SLM~4.0, and
the small core surface they depend on (\code{SLMConfig}, \code{MemoryEngine}, \code{Mode})
is present and load-verified against 4.1.3. Published adapter releases and full
per-framework integration testing are pending.

The new architectural modules for the transaction spine, policy registry, admission
gateway, schema-capability record, backup coordinator, and erasure manifest are
additive: they sit beneath the existing product, and surfaces adopt them
incrementally behind transport tests.

\section{Reliability Evaluation}
\label{sec:evaluation}

We evaluate \slm{}'s durability, isolation, temporal, and admission guarantees by
direct measurement against an installed package reporting version 4.1.3 on
Python~3.14.5---a
fault-injection and mechanism evaluation, not an external conversational-accuracy
benchmark.

\textbf{Method.} Each experiment imports the installed \code{superlocalmemory}
package and drives production code paths---real SQLite databases built through
the production initialization sequence, the real backup coordinator, the real
migration runner, the real authorization/scope layer, the real temporal machinery,
the real \code{MemoryTransactionService}, the real \code{ErasureService} with
concrete projection owners from \code{concrete\_owners.py}, the real
\code{generation\_fence} module, and the real \code{\_DEFAULT\_REGISTRY}. Synthetic inputs
are limited to a deterministic vector where a vector is structurally required, plus the
synthetic projection owners of exp2 (superseded by the real owners in exp2b; see the
per-experiment disclosures below).

Every experiment is \textbf{bracketed}: it fails if the mechanism does nothing
\emph{or} does the wrong thing. Isolation (exp5) carries an explicit
\textbf{positive control}---the requester must still see its own data through the
same read paths. The harness is fail-loud: an infrastructure error crashes the run.
Per-experiment setup disclosures: exp2 uses synthetic \code{\_TrackingOwner}
stand-ins; exp7 builds the full production schema through the migration runner's
\code{apply\_all} over both stores, rather than an enumerated migration subset; exp8
is CPU-only with no
SQLite database. Among the rest, only exp1 exercises the full production projection
owner set.

\subsection{Results}
\label{sec:eval:results}

Ten of the eleven scoped component properties held in every one of their 200
deterministic repetitions (\Cref{tab:results}) at the disclosed scope of the table.
The eleventh---the generation fence---held \textbf{199 of 200}. The released
evidence bundle therefore reports \textbf{2{,}199/2{,}200}, where the previous
version of this paper reported 2{,}200/2{,}200 on the same eleven scenarios. The
real-owner replication of \Cref{sec:eval:realowner} (exp2b, 200 repetitions) is
included in that total, and every experiment is executed by the single runner
\code{benchmark/run\_all.py}.

We report the single failure rather than re-running until it disappears, and it is
worth being explicit about why: selecting a favourable run is the precise practice
this paper's evidence standard exists to forbid, and we have criticised it in our
own prior work elsewhere in this section.

The failing repetition was the \emph{positive} control---a fresh-epoch write that
should have been admitted---not the stale-epoch rejection. It failed because the
canonical writer was transiently unavailable, so the fence's judgement is not
implicated. Diagnosing it produced a further instance of the failure class in
\Cref{sec:eval:taxonomy}. The writer collapsed three unrelated exception types into
one message: an unavailable admission journal (contention, retryable), lost writer
ownership (another process holds the lease), and a write-coordinator error---which
is \emph{the same type the generation fence raises to reject a stale epoch}. A
spurious fence rejection and a disk stall therefore produced identical text, for
the operator reading a log and for this experiment recording a failure. The three
causes are now named in the message, so the next occurrence is attributable. The
underlying transient is unresolved. Re-running the same scenario five further times
produced 200/200 on each, so the observed rate is one occurrence in 1{,}200
repetitions rather than one in 200; a rate that low is characterised, not
diagnosed, and it is listed in \Cref{sec:limitations}. No single scenario exercises
the full end-to-end path (HTTP auth $\to$ \code{ActorContext} $\to$ journal $\to$
fence $\to$ owners $\to$ obligation ledger $\to$ manifest $\to$ ANN); transport,
multi-process, mesh, and long-lived deployment fault-injection are future work.

\begin{table}[htbp]
\centering
\caption{Reliability evaluation---200 deterministic repetitions per scenario, \slm{} v4.1.3 / Python~3.14.5 / macOS-26.4.1-arm64. Ten scenarios held 200/200; the generation fence held 199/200.}
\label{tab:results}
\footnotesize
\setlength{\tabcolsep}{4pt}
\begin{tabular}{@{} c p{7.6cm} l r r r @{}}
\toprule
\textbf{\#} &
\textbf{Guarantee (exact as measured)} &
\textbf{Metric} &
\textbf{Trials} &
\textbf{Held} &
\textbf{Rate} \\
\midrule
1 &
  \code{Bm25Owner} + \code{TemporalOwner} + \code{VectorOwner}
  erase all wipe-tenant projection rows from \code{bm25\_tokens},
  \code{fact\_temporal\_validity}, and \code{embedding\_metadata};
  tombstones and a verifiable receipt persisted; keep-tenant
  content-hash unchanged &
  complete-erasure & 200 & 200 & 1.000 \\[4pt]

2 &
  \code{MemoryTransactionService}: committed op $\to$ manifest COMPLETE
  with both owners applied; faulted op $\to$ manifest DEGRADED, successful
  owner's projection removed by real \code{service.compensate()},
  failed owner had no projection residue &
 manifest-correct & 200 & 200 & 1.000 \\[4pt]

2b &
  Real \code{Bm25Owner}, \code{TemporalOwner}, and \code{VectorOwner}:
  committed writes yield COMPLETE manifests with all projections present;
  injected Bm25-owner failure yields a DEGRADED manifest and removes the
  surviving temporal projection to zero residue &
  real-owner-correct & 200 & 200 & 1.000 \\[4pt]

3 &
  Newer-stamped DB raised \code{SchemaVersionError} on the deferred pass
  with no change to the table-name set &
  refuse+preserve & 200 & 200 & 1.000 \\[4pt]

4 &
  Partial-restore failure rolls live data back to pre-restore bytes &
  rollback+clean & 200 & 200 & 1.000 \\[4pt]

5 &
  Personal rows of one tenant never leak to another on the two tested
  read paths (\code{get\_facts\_by\_ids} and \code{TemporalChannel}),
  with positive control &
  zero-leak & 200 & 200 & 1.000 \\[4pt]

6a &
  Superseded facts kept but demoted $0.25{\times}$ and re-ranked below valid &
  correct-demotion & 200 & 200 & 1.000 \\[4pt]

6b &
  Ebbinghaus retention non-increasing in age, bounded $[0,1]$ &
  monotonic-decay & 200 & 200 & 1.000 \\[4pt]

6c &
  Date-proximate events outrank distant ones; horizon excludes far past &
  correct-window & 200 & 200 & 1.000 \\[4pt]

7 &
  Generation fence rejects a stale-epoch admission
  (\code{WriteCoordinatorError}, writer never called) and admits a
  fresh-epoch admission &
  fence-correct & 200 & \textbf{199} & \textbf{0.995} \\[4pt]

8 &
  \code{\_DEFAULT\_REGISTRY.evaluate()}: REMEMBER allowed for
  OWNER/ADMIN/MEMBER; VIEWER$\to$REMEMBER denied; unknown kind fail-open
  in local, fail-closed in company; unauthenticated actor denied with
  \code{authentication\_required} before role check &
  policy-correct & 200 & 200 & 1.000 \\

\midrule
 & \textbf{Execution total} & & \textbf{2,200} & \textbf{2,199} & \textbf{0.9995} \\
\bottomrule
\end{tabular}
\end{table}

These are deterministic property scenarios, so the repetitions test for flakiness
and environment sensitivity rather than sampling a defined random population. We
therefore report no confidence interval or inferred failure probability.

Experiments exp7, exp2, exp5, exp1, and exp6a correspond to the five Verified Design
Invariants of \Cref{sec:formal} (\Cref{thm:fence,thm:manifest,thm:isolation,thm:erasure,thm:temporal})---generation
fence, completion manifest, read isolation, erasure completeness, and as-of demotion---covering
1{,}000 of the 2{,}200 released repetitions; the remainder cover real-owner validation
(exp2b), policy-registry evaluation (exp8), and further correctness properties
(migration downgrade, backup rollback, and additional temporal mechanisms).

\phantomsection\label{sec:eval:realowner}
\textbf{Real-owner validation (exp2b).} Because exp2's manifest/compensate check uses
synthetic owner stand-ins, we additionally ran \textbf{exp2b} with the real
\code{Bm25Owner}, \code{TemporalOwner}, and \code{VectorOwner} driven through
\code{MemoryTransactionService}: all 200 repetitions held---100 committed writes yielded a
\textsc{complete} manifest with all three real projections physically present, and 100
with an injected Bm25-owner failure yielded a \textsc{degraded} manifest in which the real
\code{service.compensate()} removed the surviving temporal projection to zero residue and
the failed owner left no partial rows. \emph{Scope}: the Bm25 fault is an isolated
owner-verify failure (the live retrieval-engine heal path is not exercised), and the
\code{VectorOwner} runs in its \code{embedding\_metadata} metadata mode with the
\code{sqlite-vec} ANN index out of scope, as in exp1.

\subsection{Real-Scale Performance}
\label{sec:eval:perf}

\emph{Version of record for this subsection.} The figures below were produced against
an installed package reporting version \textbf{4.0.0}---the version this paper
presents---whereas the reliability evaluation of \Cref{sec:eval:results} was re-run on
4.1.3. We state the split rather than blur it, and the reason for it is the point of
this paper: the reliability harness was re-run because a later release \emph{changed}
its result, from 200 of 200 to 199 of 200 on one scenario, and a changed result must be
reported at the version that produced it. Nothing in the release notes between 4.0.0 and
4.1.3 touches the read, write or embedding paths measured here, so we did not manufacture
a version match we had no evidence for. A reader who wants these numbers at a later
release should re-run the released harness rather than read a re-labelled table.

Performance was measured against a retained memory-store database copy
(1{,}232\,MiB; 768-d embeddings), driving the real engine with warm caches, GC disabled
during timed loops, and embedder warm-up excluded from timing. \emph{Provider scope}:
these figures use the in-process \code{sentence-transformers} embedder
(\code{nomic-ai/nomic-embed-text-v1.5}), whose cold model load is 10{,}009.9\,ms and is
excluded; an earlier measurement of this suite against a warm Ollama endpoint recorded a
541.8\,ms warm-up. Embedding provider and database size both materially affect these
numbers, so both are stated rather than left implicit; figures produced under different
providers are not directly comparable.

\begin{figure}[htbp]
  \centering
  \begin{tikzpicture}
    \begin{axis}[
        width=0.86\linewidth, height=5.2cm,
        ybar, bar width=12pt,
        xlabel={Concurrent workers}, ylabel={Throughput (ops/s)},
        symbolic x coords={1,2,4,8,12,16}, xtick=data,
        ymin=0, ymax=12.5,
        nodes near coords, nodes near coords style={font=\scriptsize},
        enlarge x limits=0.12,
        tick label style={font=\small}, label style={font=\small},
      ]
      \addplot[fill=slmbluebg, draw=slmblue, line width=0.6pt]
        coordinates {(1,0.9)(2,1.9)(4,5.4)(8,9.0)(12,15.0)(16,18.4)};
    \end{axis}
  \end{tikzpicture}
  \caption{Mixed remember/recall throughput versus concurrency on a $\approx$1\,GB store
    (Mode~B, Ollama embeddings, \code{ThreadPoolExecutor}). Throughput scales from 1.0 to
    18.4\,ops/s across 1--16 workers while median per-operation latency rises from 74.9\,ms to 1{,}009.4\,ms
    (0.84--1.23\,s), and \textbf{zero writer-lock errors} occur at any level: the single-writer
    WAL model serialises writers without contention failures.}
  \label{fig:perf-concurrency}
\end{figure}
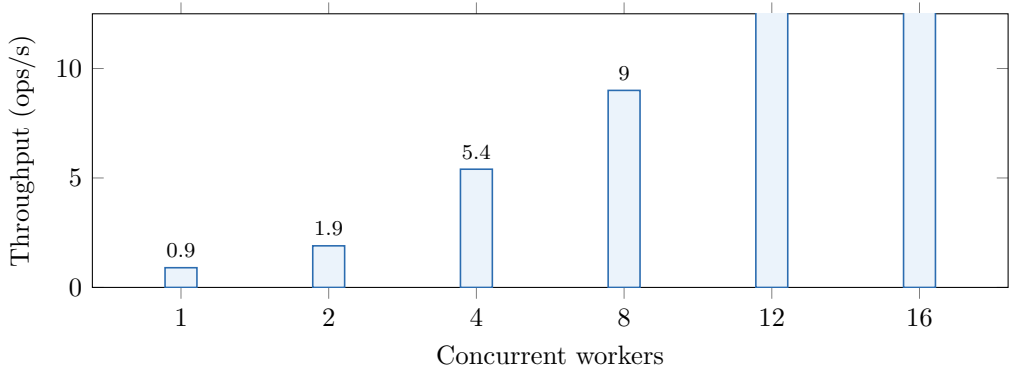

\textbf{Concurrency.} \Cref{fig:perf-concurrency} sweeps 1--16 GIL-shared threads
alternating \code{store\_fast} writes and direct \code{recall} (the legacy bypass path,
not the governed write path). Throughput rises from 0.9 to 18.4\,ops/s across 1--16 workers, while median
per-operation latency degrades from 74.9\,ms to 1{,}009.4\,ms as GIL and write-lock
contention grow, with \textbf{zero ``database is locked''
errors observed at any level} under the single-writer, WAL-journaled design; we report the
observation without attributing multi-process or transport causality.

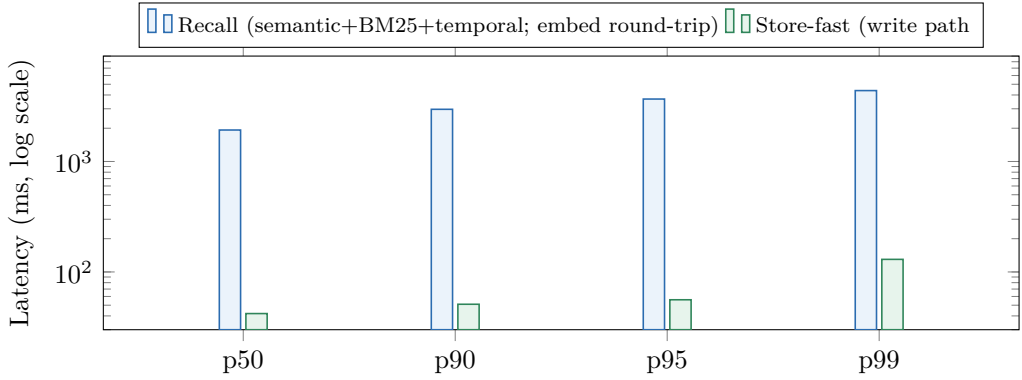
\begin{figure}[htbp]
  \centering
  \begin{tikzpicture}
    \begin{axis}[
        width=0.86\linewidth, height=5.2cm,
        ybar, bar width=8pt, ymode=log,
        log origin=infty,
        ymin=30, ymax=9000,
        ylabel={Latency (ms, log scale)},
        symbolic x coords={p50,p90,p95,p99}, xtick=data,
        enlarge x limits=0.22,
        legend style={font=\scriptsize, at={(0.5,1.03)}, anchor=south, legend columns=2},
        tick label style={font=\small}, label style={font=\small},
      ]
      \addplot[fill=slmbluebg, draw=slmblue, line width=0.6pt]
        coordinates {(p50,1927)(p90,2967)(p95,3675)(p99,4383)};
      \addplot[fill=slmgreenbg, draw=slmgreen, line width=0.6pt]
        coordinates {(p50,42)(p90,51)(p95,56)(p99,130)};
      \legend{Recall (semantic+BM25+temporal; embed round-trip), Store-fast (write path, no embed)}
    \end{axis}
  \end{tikzpicture}
  \caption{Operation-latency percentiles over $n{=}300$ warm operations on a 1{,}232\,MB store
    (Mode~B; log scale). Recall latency uses the in-process sentence-transformers provider
    (\code{nomic-ai/nomic-embed-text-v1.5}); the write path without embedding (store-fast) is 42.0\,ms at the median.
    Latency is therefore a property of the configured embedding provider, not of the control plane;
    embedder warm-up (10{,}009.9\,ms, in-process sentence-transformers) is excluded from timing and GC is disabled during each timed loop.}
  \label{fig:perf-latency}
\end{figure}

\textbf{Latency.} \Cref{fig:perf-latency} reports percentiles over $n{=}300$ warm
operations on the 1{,}232\,MiB store. Recall is embedding-bound: 1.93\,s at the median
and 4.38\,s at p99 with the in-process sentence-transformers provider. The 42.0\,ms median for the \emph{legacy
direct} \code{store\_fast} write path on that store bypasses canonical admission, the
policy registry, the encrypted journal, projection obligations, and the completion
manifest, so it does \emph{not} bound governed-path cost.

\textbf{Governed write-envelope overhead.} To bound the control-plane cost directly, we
measured the canonical governed write envelope in-process
(\code{CanonicalRememberRuntime.remember()}---the same path the HTTP handler calls after
request parsing: encrypted admission-journal prepare/dispatch/commit, the single-writer
\code{WriteCoordinator}, the canonical SQLite write, and the projection obligation-ledger
insert) against the ungoverned in-process write on a fresh store, $n{=}1{,}000$ warm calls each
after a 30-call warm-up burst.
The previous version of this paper \emph{reported} a governed envelope at p50/p95/p99
$=3.522/4.427/5.297$\,ms against $1.835/2.379/2.569$\,ms ungoverned, and derived from
that pair a control-plane overhead of $1.687$\,ms (p50) to $2.728$\,ms (p99). Those
five numbers are stated here in the past tense deliberately: they are the claim under
retraction, not a finding of this version, and neither the pair nor the difference
should be quoted from this paper as current.

\textbf{We withdraw that overhead figure.} Re-measuring it on the current release
returned a \emph{negative} p50 delta, which is impossible for a genuine overhead: a
governance envelope can only add work. The number was not mis-measured, it was
mis-defined---and our first account of \emph{why} was also wrong, which is worth
recording because a reader can check it.

We initially concluded that the two paths were not nested: that the comparand
performed three writes in one transaction where the governed writer performed one.
That is false. The governed path constructs the same
\code{IngestionCommand} around the same writer callback and calls
\code{submit()} on it, then adds the journal, the coordinator binding and the
projection-obligation records. The comparand is a logical \emph{subsequence} of the
governed path, so the governed path does strictly more work---and still measures
faster at the median. Structural nesting was never the problem.

We then offered a second explanation, and it is also wrong. We said that the
governed path holds one transaction across a run of writes while the comparand
commits once per call, so the subtraction measured commit batching. It does not.
The write coordinator issues \code{BEGIN IMMEDIATE} and commits once for
\emph{each} queued write. Both real arms perform one transaction and one commit per
write, and nothing is batched across calls in either.

We are not going to propose a third cause. Two explanations, each consistent with
the code we had read at the time, each checkable, each wrong. We know of two ways
the arms differ besides governance: the governed path reuses one long-lived
connection where the comparand is called standalone, and the governed path sets
SQLite's \code{synchronous} pragma per write lane while the comparand never sets it
at all. Either could produce the sign. We have not isolated which, and naming one
would repeat the error this paragraph exists to record.

Two further facts belong here. Interleaving the arms under a recorded seed did not
change the sign, so this is not an ordering artifact. And each arm writes to its own
temporary database, so they never contend for the same file; sharing one would trade
that for cross-arm interference. Neither arrangement isolates the envelope, and that
is the point: no choice of comparand does, which is why \Cref{sec:eval:envelope}
stops looking for one.

The released result artifact carries our \emph{first} explanation in its
interpretation field, and we withdraw it there as well. Its measurements are
unaffected: they are absolute per-arm costs, and the field it contradicts is a note,
not a number. Its \code{overhead\_well\_defined} flag already reads false.

\Cref{tab:latency-arms} therefore reports the three paths as absolute costs and
states no difference between them.

\subsubsection{Measuring the envelope instead of subtracting pipelines}
\label{sec:eval:envelope}

The failure above is not a measurement error, it is a method error: we asked what
governance costs by differencing two programs, and no amount of care in choosing the
comparand fixes a question posed that way. So we stopped subtracting. The envelope is
a known set of components on a known path, and each can be timed where it runs,
inside the same transaction on the same connection under the same pragmas as the work
it is being compared against. Nothing is bypassed, no comparand is constructed, and
there is consequently no comparability question left to get wrong.
\Cref{tab:envelope} reports $n{=}400$ warm governed writes at 4.1.9.

\begin{table}[htbp]
\centering
\caption{exp13, governance envelope cost by component, measured in place.
$n{=}400$ warm calls, \slm{} v4.1.9, in-process, temporary filesystem. Total
governed write p50 $=11.00$\,ms.}
\label{tab:envelope}
\footnotesize
\begin{tabular}{@{} p{5.6cm} r r @{}}
\toprule
\textbf{Component} & \textbf{p50 (ms)} & \textbf{Share of write} \\
\midrule
Admission journal --- commit (terminal row)           & 3.137 & 28.5\% \\
Admission journal --- prepare (encrypt + durable row) & 3.109 & 28.3\% \\
Admission journal --- dispatch transition             & 1.480 & 13.5\% \\
Projection-obligation ledger insert                   & 0.042 & 0.4\% \\
Generation fence (epoch check)                        & \textbf{0.0019} & 0.02\% \\
\midrule
\textbf{Envelope total}                               & \textbf{7.77} & \textbf{70.6\%} \\
\bottomrule
\end{tabular}
\end{table}

The distribution is the result, and it is not the one we expected to report. The
envelope is roughly seventy per cent of a governed write, and \emph{essentially all
of it is durability}. The two components that make the write \emph{governed} rather
than merely durable---the generation fence that decides whether an epoch may still
write, and the obligation ledger that makes cross-store completion checkable---cost
$0.044$\,ms together, four tenths of one per cent. The remaining $7.73$\,ms is three
transitions of an encrypted write-ahead journal.

\textbf{Governed memory is not expensive because it is governed. It is expensive
because it is durable.} The policy machinery is microseconds; the write-ahead record
is milliseconds. That distinction matters to anyone deciding whether they can afford
this design, because the two halves have entirely different engineering responses: a
deployment that needs lower write latency should reach for journal batching, group
commit or a faster device, and will find nothing worth removing in the governance
decisions themselves. Repeated runs gave envelope shares of $70.3$, $70.6$, $70.7$ and
$71.3$ per cent, with the fence between $0.0019$ and $0.0021$\,ms in every one; the
journal component tracks the device and ranged from $7.29$ to $8.47$\,ms, which is
the expected sensitivity and the reason the share is quoted rather than the absolute
alone.

\emph{Scope.} In-process; HTTP transport, request parsing and the trust hook are
excluded. Databases sit on a temporary filesystem, so a persistent disk raises the
journal components specifically. The sum of component medians is not the median of
the sum, so the share is indicative rather than an identity.

\begin{table}[htbp]
\centering
\caption{Write-path cost by arm, interleaved under a recorded seed,
$n{=}1{,}000$ per arm. The arms are not comparable in ways we have not isolated, so
no difference between them is a governance overhead. The retained artifact records
the platform and the module measured but no package version, and we do not assign
one here.}
\label{tab:latency-arms}
\footnotesize
\begin{tabular}{@{} p{6.4cm} r r @{}}
\toprule
\textbf{Path} & \textbf{p50 (ms)} & \textbf{p99 (ms)} \\
\midrule
Governed: journal, coordinator, writer & 17.2 & 130.2 \\
Ungoverned: \code{IngestionCommand.submit()}, own transaction per call & 51.1 & 99.8 \\
\emph{(artifact)} bare writer, no caller transaction & 173.7 & 343.5 \\
\bottomrule
\end{tabular}
\end{table}

The third row is included for completeness and is not a peer of the other two: the
innermost writer is built to run inside a caller-managed transaction, and calling it
standalone adds a per-call \code{fsync} that neither real path pays. It bounds
nothing.

The governed path is the faster of the two real paths. We report that and stop
there, because every account we have given of \emph{why} has failed on inspection.
What survives is the observation itself: adding a journal, a fence, an obligation
ledger and a single-writer queue did not make this write path slower than calling
its own inner command directly. That is worth knowing and it is not a measurement of
governance overhead. The defensible statement about cost is the absolute one: a
governed write completes in
$17.2$\,ms at the median. That figure is measured on freshly created temporary
stores, not on the $1{,}232$\,MiB copy used for the recall, concurrency and
resident-memory results above; it is a floor, and a larger store will not be
faster.
\emph{Scope}: in-process (HTTP transport and request parsing excluded; loopback HTTP
typically adds $0.5$--$2$\,ms); databases on a temporary filesystem, so persistent-disk
fsync on the first journal write may raise p50 toward the low tens of milliseconds---these
are floor numbers---and the comparand arm is \code{IngestionCommand.submit()} called
standalone. A full end-to-end HTTP benchmark under
multi-agent concurrency remains future work.

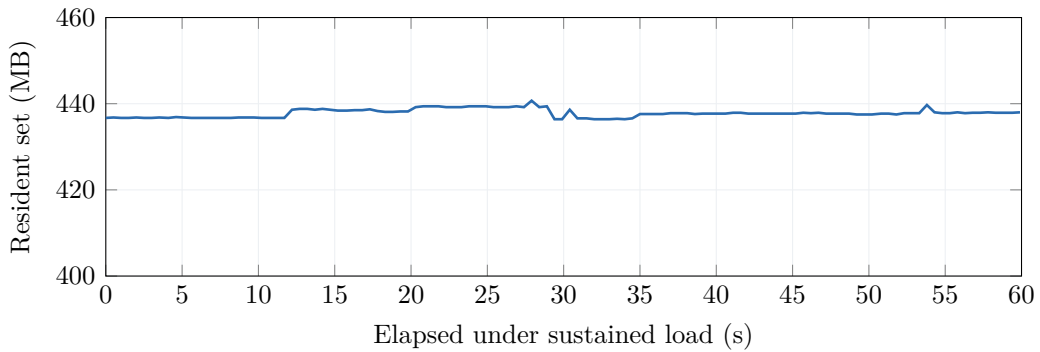
\begin{figure}[htbp]
  \centering
  \begin{tikzpicture}
    \begin{axis}[
        width=0.86\linewidth, height=5.0cm,
        xlabel={Elapsed under sustained load (s)}, ylabel={Resident set (MB)},
        xmin=0, xmax=60, ymin=400, ymax=460,
        tick label style={font=\small}, label style={font=\small},
        grid=major, grid style={slmgraybg},
      ]
      \addplot[draw=slmblue, line width=1pt, mark=none]
        coordinates {(0.0,436.7)(0.5,436.8)(1.0,436.7)(1.5,436.7)(2.0,436.8)(2.5,436.7)(3.0,436.7)(3.5,436.8)(4.1,436.7)(4.6,436.9)(5.1,436.8)(5.6,436.7)(6.1,436.7)(6.6,436.7)(7.1,436.7)(7.6,436.7)(8.2,436.7)(8.7,436.8)(9.2,436.8)(9.7,436.8)(10.2,436.7)(10.7,436.7)(11.2,436.7)(11.7,436.7)(12.2,438.6)(12.7,438.8)(13.2,438.8)(13.7,438.6)(14.2,438.8)(14.7,438.6)(15.2,438.4)(15.8,438.4)(16.3,438.5)(16.8,438.5)(17.3,438.7)(17.8,438.3)(18.3,438.1)(18.8,438.1)(19.3,438.2)(19.8,438.2)(20.3,439.2)(20.8,439.4)(21.3,439.4)(21.8,439.4)(22.3,439.2)(22.8,439.2)(23.3,439.2)(23.8,439.4)(24.4,439.4)(24.9,439.4)(25.4,439.2)(25.9,439.2)(26.4,439.2)(26.9,439.4)(27.4,439.2)(27.9,440.7)(28.4,439.2)(28.9,439.4)(29.4,436.4)(29.9,436.4)(30.4,438.6)(30.9,436.6)(31.5,436.6)(32.0,436.4)(32.5,436.4)(33.0,436.4)(33.5,436.5)(34.0,436.4)(34.5,436.6)(35.0,437.6)(35.5,437.6)(36.0,437.6)(36.5,437.6)(37.0,437.8)(37.5,437.8)(38.1,437.8)(38.6,437.6)(39.1,437.7)(39.6,437.7)(40.1,437.7)(40.6,437.7)(41.1,437.9)(41.6,437.9)(42.1,437.7)(42.6,437.7)(43.2,437.7)(43.7,437.7)(44.2,437.7)(44.7,437.7)(45.2,437.7)(45.7,437.9)(46.2,437.8)(46.7,437.9)(47.2,437.7)(47.7,437.7)(48.2,437.7)(48.7,437.7)(49.2,437.5)(49.8,437.5)(50.2,437.5)(50.8,437.7)(51.3,437.7)(51.8,437.5)(52.3,437.8)(52.8,437.8)(53.3,437.8)(53.8,439.7)(54.3,438.0)(54.8,437.8)(55.3,437.8)(55.8,438.0)(56.3,437.8)(56.8,437.9)(57.3,437.9)(57.8,438.0)(58.3,437.9)(58.8,437.9)(59.4,437.9)(59.9,438.0)};
    \end{axis}
  \end{tikzpicture}
  \caption{Process resident-set size over 60\,s of sustained mixed load on a 1{,}232\,MB store.
    RSS remains within 436.4--440.7\,MB (4\,MB rounded spread; final 438.0\,MB)
    with no monotonic upward trend---consistent with a bounded working
    set (canonical pages plus embedding buffers) rather than an unbounded leak. A longer soak with an
    explicit idle-recovery phase is noted as future validation (\Cref{sec:limitations}).}
  \label{fig:perf-rss}
\end{figure}

\textbf{Memory stability.} \Cref{fig:perf-rss} tracks RSS across 60\,s of sustained
load. After initial warm-up, RSS remains within 436.4--440.7\,MB with no monotonic
upward trend (4\,MB rounded spread; final 438.0\,MB). The harness classifies this
window as \texttt{NONE} for its bounded-spread leak indicator; a definitive leak verdict still requires a longer soak with
an explicit idle-recovery phase (future work).

\subsection{Retrieval Quality: Carried-Forward V3 LoCoMo Evidence}
\label{sec:eval:locomo}

The experiments above measure durability, isolation, admission, and temporal
\emph{mechanisms}; they are not a retrieval-accuracy benchmark. For retrieval quality
we carry forward the published LoCoMo~\citep{locomo} evidence for the V3 architecture
that \slm{} productionizes; we do \emph{not} rerun it against the V4 release
artifact, and every figure is protocol-scoped.

\begin{table}[htbp]
\centering
\caption{Carried-forward LoCoMo evidence for the V3 architecture (protocol-scoped;
\emph{not} a rerun V4-release benchmark). Source:~\citep{slmv3prior}.}
\label{tab:locomo}
\footnotesize
\setlength{\tabcolsep}{5pt}
\begin{tabular}{@{} l r p{7.0cm} @{}}
\toprule
\textbf{Configuration} & \textbf{LoCoMo} & \textbf{Protocol scope} \\
\midrule
Mode~A Raw & 60.4\% &
  10 conversations; 1{,}276 scored questions; local embeddings, local retrieval,
  zero-LLM answer construction. \\[3pt]
Mode~A Retrieval & 74.8\% &
  Same set; local retrieval followed by a single GPT-4.1-mini answer-formation pass---the
  standard LoCoMo answering protocol, distinct from the zero-LLM Raw row above. Published
  V3 result, not re-derived on the V4 artifact; a hash-pinned V4 re-run is future work. \\[3pt]
Mode~C & 87.7\% &
  Conv-30 only; 81 scored questions; text-embedding-3-large plus GPT-4.1-mini
  answer generation and judge. Not a full-dataset result. \\
\bottomrule
\end{tabular}
\end{table}

The V3 Mode~A Retrieval result breaks down by category as 72.0\% single-hop, 70.3\%
multi-hop, 80.0\% temporal, and 85.0\% open-domain. An information-geometric ablation on
six V3 LoCoMo conversations reported $+12.7$\,pp under the published protocol;
the displayed one-decimal endpoints (71.7\% with versus 58.9\% without the
geometric layers) imply 12.8 points because of rounding. Both are V3-architecture
results, not re-verified on V4; an end-to-end
V4 comparative benchmark against LoCoMo and LongMemEval is preregistered future work.

\subsection{A Failure Class Found by Operating the System}
\label{sec:eval:taxonomy}

\textbf{The harness in \Cref{sec:eval:results} did not find any of what follows.}
That is the first thing to say about it, because conflating the two would
misrepresent both. The scenarios in \Cref{tab:results} are adversarial fault
injections designed against hypothesised failures; the instances below were found
by running this system across two machines for a month, and every one of them was
present while all eleven scenarios passed.

The previous version of this paper closed by noting that transport, multi-process
and long-lived deployment fault injection were future work. This section is a
partial and unplanned discharge of that item.

\subsubsection{Ten instances of one shape}

\begin{table}[htbp]
\centering
\caption{Mechanisms that were implemented, reachable on a live call path, and
ineffective or unobservable at their final connection. Each was verified directly
against source and stores; the middle column names the signal an operator would
have consulted.}
\label{tab:failure-class}
\footnotesize
\setlength{\tabcolsep}{4pt}
\begin{tabular}{@{} p{4.5cm} p{4.3cm} p{4.6cm} @{}}
\toprule
\textbf{Mechanism} & \textbf{What the operator saw} & \textbf{What was true} \\
\midrule
Channel-selection bandit & 1{,}405 plays on this store, timestamps advancing &
posterior mean $0.5$ on 459/459 arms across three stores (5{,}657 plays); no arm
holds any preference \\
Source-trust model & 629 observations recorded & posterior mean $0.5$ on 37/37
sources; every recorded reward $0.5$ \\
Trust-modulated retention & feature configured and called per fact & guard reads a
column absent from the table it tests; $\tau=1.0$, so $\lambda_{\mathrm{eff}} =
\lambda$ \\
Recall-health monitor & healthy & semantic retrieval off for over an hour;
liveness was inferred from the probe output it gates \\
Availability gate on a data invariant & 503 on every route, indefinitely & the
store was serving correctly; one drifted row failed a re-checkable data condition
that ordinary use can re-violate \\
Schema bootstrap vs.\ deferred migration & \code{no such column} & a valid older
store; startup indexed a column a deferred migration adds \\
Editor-plugin distribution & correct version, everywhere & two versions serving
one store; a private data directory made a 611\,MB store present as 28\,KB \\
Writer-unavailability reporting & one message, three causes & a stale-epoch
rejection was indistinguishable from a disk stall \\
Governance-overhead figure & a cheap, measured control-plane cost & the quantity
was never an overhead: the two arms differenced are not comparable, and two attempts
to say why were both wrong \\
Benchmark entry point & documented command & the documented interpreter carries a
release three years old \\
\bottomrule
\end{tabular}
\end{table}

Two of the ten were introduced by the previous version of this paper rather than by
the system: the overhead figure, and the benchmark command that produced it. We
count them because they were the same failure---a wired artifact whose output does
not mean what its label says---and excluding the two we authored ourselves would be
the least defensible edit available.

A broader internal review of this codebase identified further instances beyond
these ten. We report only the ten we re-derived first-hand for this paper, and note
that at least one item from that wider set has since been repaired---code-graph
vector search, whose extension is now loaded per connection---which is a reason to
publish a verified count rather than an inherited one.

\subsubsection{What they share}

What the ten share is not silence. We said that in an earlier draft and the table
above refutes it: one returns 503 on every route, one raises \code{no such column},
and one is a documented command pointing at a three-year-old interpreter. Those are
loud. What they share is that \textbf{the signal, where there was one, did not name
the thing that was wrong.} A 503 says the store is unavailable when the store is
fine. \code{no such column} names a column when the fault is a migration that has not
run. Loudness and diagnosticity are independent, and only the second is useful.

It is worth separating the subclass that is genuinely silent, because it is the
harder one and it is where our two checks apply. In four of the ten there is no
signal at all: nothing raised, nothing logged at warning or above, every file present
and every version string correct. A learner sitting at its prior mean and a decay
multiplier reducing to one produce output identical to correct operation. Those four
would not appear in a coverage report as uncovered, and none is a crash, a leak, or a
wrong answer to a user's question. The remaining six fail loudly, or partially, or
outside the runtime altogether---two are artifacts of the previous version of this
paper rather than of the system---and we flatten them at our peril. Two properties
recur across the set.

\paragraph{Implemented, reachable and effective are three different questions.}
A grep answers the first. A call-graph trace answers the second. Neither answers the
third. Answering it requires an oracle over the \emph{effect} the mechanism is
supposed to produce, and that oracle has to be independent of the mechanism. For a
learner or a schema-guarded join, persisted state is the convenient place to find
one, which is what our two checks use. It is not the general answer: process
availability, package distribution and experimental validity each need a different
independent observer, and three of the ten could not have been settled by querying a
store at all. The principle is oracle independence. Querying the store is one
instance of it.

\paragraph{A signal derived from the mechanism it evaluates cannot detect that
mechanism's failure.} A liveness probe reading the output it gates, a play counter
standing in for learning, a version string reported by whichever copy started
first, a latency delta whose two arms were assumed to differ only in governance. In each case the
instrument and the subject were the same component, so the failure was outside the
instrument's range.

These two properties are why the usual instrumentation does not help. Adding
logging to a mechanism that believes it is working produces confident logs; adding
a metric derived from its own output produces a healthy metric.

\subsubsection{Threats to validity}

This is not a controlled experiment and we do not present it as one.

\begin{itemize}[leftmargin=*,itemsep=1pt]
  \item \textbf{One codebase, two machines, no control condition.} Whether the
        prevalence generalises is untested. What generalises, if anything, is the
        \emph{shape}---and the two invariants of \Cref{subsec:invariants} are the
        falsifiable part: point them at another system and they either find
        instances or they do not.
  \item \textbf{Found by operation, not by search.} There was no systematic sweep
        that would let us report a denominator. We cannot say what fraction of
        this system's mechanisms are affected, only that these ten are.
  \item \textbf{Selection is toward the visible.} An inert mechanism nobody
        happened to query is by construction absent from this table. The count is
        a floor.
  \item \textbf{Two authors of the finding are also the authors of the system},
        which is what makes the first-hand re-derivation and the shipped checks the
        substance of the claim rather than the narrative around it.
  \item \textbf{One unexplained failure is excluded.} A single segmentation fault
        was observed once during full-suite runs, in garbage collection inside the
        validation layer with a vector-store event loop live. It has not been
        root-caused and does not appear in \Cref{tab:failure-class}; a section
        about invisible failures that hid its own unexplained one would not be
        worth reading.
\end{itemize}

\section{Limitations and Threats to Validity}
\label{sec:limitations}

\textbf{Authorization-layer isolation only.} Tenant and scope isolation are enforced
by the authorization layer; a single OS user with direct filesystem access to the
SQLite stores is outside the threat model.

\textbf{Mesh validated only at small scale.} Per-device revocation and key rotation,
WAN-latency and network-partition fault injection, and multi-machine operation beyond
a two-node loopback remain future work. Mesh is coordination---not consensus, quorum,
or a CRDT database.

\textbf{Temporal accuracy not benchmarked end-to-end.} We measure temporal
\emph{mechanisms}; we do not report agent-task accuracy against an external benchmark.
The cross-surface \code{as\_of} demotion path is newly wired and not yet benchmarked.
The \code{EbbinghausLangevinCoupling} class is experimental with no production caller.

\textbf{Data-subject rights are workspace-scoped, not member-scoped.} Erasure and
export operate on a profile. A member of a shared workspace cannot exercise access
or erasure over \emph{only their own} contributions, because no member-scoped
erasure path exists and the erasure operation is owner-only. The obstacle is
identity, not plumbing: facts carry no author column, and the provenance table that
does carry one covers 69.8\% of facts with 3{,}997 of 4{,}340 authors recorded as
opaque capability digests and 237 as \code{unknown}. For a single-operator
deployment this is immaterial. For a shared workspace under GDPR it is a real
limitation on Article~15 access and Article~17 erasure at member granularity, and a
deployer should treat the profile, not the person, as the unit those rights attach
to until a member-scoped path exists.

\textbf{Verified erasure is scoped to live stores.} The completeness gate covers the
canonical store and the registered projection owners. Backup artifacts and any
remote peer that has received state are outside it, and the completeness flag is
withheld while backup obligations are pending rather than asserted over them. The
audit chain establishes same-host tamper evidence; it is not a remote trust anchor
and does not attest to another machine.

\textbf{Learning is shown under ablation, not in deployment.}
\Cref{sec:learning-layer} reports a channel-selection bandit, a source-trust model,
and a trust-modulated retention term that were wired and had never changed an outcome
at the version whose field measurements this paper reports. Two had their named cause
repaired afterwards, and \Cref{tab:exp12} shows under ablation that the repaired path
carries a signal from an agent's behaviour to a posterior. Three limits on that
result, all of which matter. The engagement it consumes is generated by the
experiment, not observed in the field, so it establishes transport and not preference
quality. It says nothing about whether the arms a real deployment converges on are
ones an operator would endorse, because we ran no such deployment. And in the authors'
own live store every arm was still at its prior at the last reading, which is what
a single-operator installation over days looks like whether the loop works or not.
The third mechanism has a scoped repair named in \Cref{sec:futurework} that was not
performed. No other result in this paper depends on any of them.

\textbf{The failure class is not a controlled study.} The ten instances of
\Cref{sec:eval:taxonomy} come from one codebase on two machines, found by operating
the system rather than by a systematic sweep, so we can report no denominator and no
prevalence. An inert mechanism nobody happened to query is by construction absent,
which makes the count a floor. The falsifiable part is the two invariants: pointed at
another system they either find instances or they do not.

\textbf{One failure is unexplained and one is unresolved.} A single segmentation
fault was observed once during full-suite runs, in garbage collection inside the
validation layer with a vector-store event loop live; it has not been root-caused.
Separately, the generation-fence scenario failed one repetition in 1{,}200 through a
transient writer unavailability that is now diagnosable but not diagnosed.

\textbf{The cost of governance is measured by component, not end to end.}
\Cref{sec:eval:perf} withdraws the previous version's overhead figure because the two
arms it differenced are not comparable, and \Cref{sec:eval:envelope} replaces it by
timing the envelope's components where they run. That result is in-process and on a
temporary filesystem, so the journal components in particular will differ on a
persistent device, and the sum of component medians is not the median of the sum. It
establishes the \emph{distribution} of cost within the envelope---which is the part
that carries the finding---more firmly than it establishes any single absolute
number.

\textbf{Reliability evaluation is component-level.} The harness drives real
production modules over fresh per-trial SQLite databases but does not exercise the
HTTP/MCP/WebSocket server stack, inter-process communication, or production-scale
concurrency. The \code{sqlite-vec} ANN index is not loadable in the harness
environment; exp1's vector erasure is scoped to the \code{embedding\_metadata} SQL
table. Retained result artifacts record package version and platform but not a source
commit or wheel hash.

\textbf{External long-context benchmarks are future work.} No new V4 LoCoMo run is
reported; V3-architecture LoCoMo results are carried forward, protocol-scoped. An
end-to-end V4 benchmark against LoCoMo and LongMemEval is preregistered.

\textbf{Single-node local-first scope.} All reliability results are from single-machine
deployments; multi-node mesh behavior and distributed consistency are not
characterized.

\textbf{Concurrent and closely related prior art.} MemTX and MemTxn~\citep{memtx,memtxn}
on adjacent transaction layers; MemClaw / Governed Memory~\citep{memclaw,governedmemory}
on governed multi-tenant shared memory; PROJECTMEM~\citep{projectmem} on local-first
event-sourced agent memory. \slm{}'s position is a distinct local-first composition,
not priority over any of these.

\textbf{Corpus-bounded novelty.} ``Not found in the searched corpus'' is a dated
review through 2026-08-03, not proof of first invention.

\section{Novelty and Contributions}
\label{sec:contributions}

We state novelty with corpus-bounded language: ``not found in the searched corpus''
means not described in primary sources reviewed through 2026-08-03---a dated review,
not a priority claim. Several capabilities \slm{} provides---multi-tenant isolation,
compliance controls, temporal memory, local-first operation---exist individually in
shipping systems. We claim (a) their \emph{integration} into one governed, local-first,
verifiable runtime under a governing admission invariant (enforced on the primary write
path; target architecture for all surfaces), and (b) specific innovations the V2--V4
line of work contributes.

\paragraph{C1 --- A failure class, two invariants that expose it, and the repair one of them produced (lead).}
Ten mechanisms in this system were implemented, reachable on a live call path, and
ineffective or unobservable at their final connection, every one of them silently
(\Cref{sec:eval:taxonomy}). Two properties recur: \emph{implemented},
\emph{reachable} and \emph{effective} are three different questions, of which only
the third requires querying the store; and a signal derived from the mechanism it
evaluates cannot detect that mechanism's failure. We contribute two mechanical checks, released in 4.1.5, that
answer the third question---a prior-distance assertion over Bayesian learners, and a
join-liveness assertion over schema-guarded paths that reports where a guard's
missing data actually resides (\Cref{subsec:invariants}). The contribution does not
stop at detection. The prior-distance verdict prompted an investigation that found
two defects no counter in the subsystem could reach---two individually correct
components disagreeing on which identifiers count, and a neutral-by-default
settlement that shrank posterior variance while holding the mean---and their fixes
are public in 4.1.6--4.1.9 (\Cref{subsec:repair}). We then test the repair under
control rather than assert it. A three-arm ablation varying only the identifier
namespace (\Cref{tab:exp12}) moves no posterior with the defect present, moves every
instantiated arm with it absent, and---in a negative control that writes every
outcome ticket but supplies no engagement---correctly settles nothing at all. We
still make no attribution claim about the historical field state, because we ran no
replay against it, and a controlled ablation is not a deployment result.

Not found in the searched corpus \emph{for agent memory}, with three honest
qualifications. \citet{silentnarratives} already established silent failure in a
production agent runtime as a studied phenomenon with a mechanism-oriented
taxonomy, and we claim no priority over it; our class is the sub-case in which no
error event occurs, which its five classes do not cover and which its
human-observation finding could not surface. Detecting a conditional branch that never executes in production is
long established in software engineering as dead-conditional and flag-controlled
dead-code analysis, and we claim no novelty for the idea; our narrow claim is its
application to schema-presence guards on a learning path, where the consequence is a
learner that is arithmetically inert rather than merely disabled. And MemFail%
~\citep{memfail} diagnoses memory failure modes from outside the system across
summarization, storage and retrieval---a different axis, on which none of our ten
instances appears, because a black-box benchmark sees a plausible answer throughout.
\emph{Scope:} these are admission and observability claims. We present no evidence
that repairing any of the ten improves agent behaviour, because we have run no such
experiment.

\paragraph{C2 --- Verifiable memory transactions (co-lead).}
An atomic canonical commit with a transactional obligation ledger, per-projection
apply/verify/compensate/erase owners under generation fencing, and a hash-checkable
completion manifest (COMPLETE / DEGRADED / FAILED). Distinct from and composable with
concurrent belief-level~\citep{memtx} and logical-state~\citep{memtxn} transactions.
Manifest correctness and real \code{compensate()} behavior are \emph{measured} (exp2).

\paragraph{C3 --- Governance-native memory with deployment-context assessment (co-lead).}
A scoped governance control plane in a local-first runtime: immutable server-derived actor
identity, pure kind-keyed policy registry, role-based access, write-time
generation-fenced isolation with verified projection-store erasure, hash-chained
audit, a non-authoritative EU~AI~Act checklist that abstains without intended-use
context, and operable recovery (auto-reconciliation, a
write-path stall watchdog, and an RBAC-gated dashboard/CLI/MCP remediation surface for stuck
or degraded operations). Cross-tenant isolation (exp5), erasure completeness (exp1),
generation-fenced admission (exp7), and policy evaluation (exp8) are \emph{measured}. \emph{Scope:} authorization-layer isolation, not legal
certification.

\paragraph{C4 --- SLM-Mesh: serverless governed coordination (major).}
Serverless, per-tenant-isolated coordination through a local SQLite broker, with
optional remote peer discovery and message proxying, across all three operating modes.
\emph{Scope:} leaderless LWW, opt-in, default-profile, two-node-loopback validated;
not consensus/quorum/CRDT; WAN-scale and per-device revocation are future work.

\paragraph{C5 --- Time-aware memory (major).}
A shipped temporal stack: three-date model, bi-temporal storage, dedicated temporal
retrieval channel, non-destructive superseded-fact demotion, Ebbinghaus recency model,
and cross-surface \code{as\_of} recall. Three temporal behaviors measured directly.

\paragraph{C6 --- Information-geometric retrieval brain, productionized (major).}
Five candidate producers fused by RRF, over Fisher-Rao scoring, sheaf-cohomology
consistency, and a Riemannian--Langevin lifecycle, together with an entity knowledge
graph and unified caching and compression. Contribution is productionizing and
governing this brain inside a multi-tenant runtime.

\paragraph{C7 --- Direct reliability evaluation method (supporting).}
A bracketed, positive-controlled, fail-loud fault-injection harness measuring
agent-memory durability, isolation, admission, and temporal guarantees against the
candidate package.
\paragraph{C8 --- Governed skill evolution: a design, not a result.}
A blind-verified, budgeted pipeline under per-cycle budget controls with append-only,
hash-linked status transitions to a quarantine store. The previous version presented
this as a major contribution. \textbf{We demote it.} The pipeline is implemented and
gated and \emph{has never executed}: its cycle is entered and exits having found no
candidate, and the evolution log is empty. A design for bounding self-modification is
worth describing, and we describe it in those terms; it is not a delivered
capability, and the corpus comparison is therefore withdrawn until it runs. The
withdrawal is not merely procedural: \citet{msce} crystallises evidence-backed
policies from agent memory into callable skills carrying verification rules and
applicability boundaries, and reports results on two benchmarks. That is the same
territory, executing, where ours is gated and idle.
\emph{Scope:} default stops at quarantine; an explicit configuration flag---not human
or RBAC approval---can activate promotion; parent-version lineage and generation
tracking are not reported.

\section{Future Work}
\label{sec:futurework}

\paragraph{Memory scaling via quantization.}
\slm{}'s context-optimization layer combines exact-match caching with reversible
lossless compression to reduce working-set pressure at the session boundary. As
deployment stores grow to tens or hundreds of gigabytes---a realistic trajectory for
long-lived enterprise agents---a near-lossless vector-quantization tier for the
semantic-channel embedding store becomes necessary. \citet{Gao2024RaBitQ} showed that a
codebook-free, random-rotation 1-bit quantizer achieves a provably tight error bound and
outperforms product quantization on ANN recall at near-zero indexing cost;
\citet{Zandieh2025TurboQuant} extended this line, combining a polar-coordinate
preprocessing stage~\citep{Han2025PolarQuant} with a 1-bit quantized
Johnson--Lindenstrauss residual corrector~\citep{Zandieh2024QJL} to reach near-optimal
distortion across bit-widths. The natural \slm{} extension is a pluggable, data-oblivious
quantization tier in the optimize layer that compresses stored embeddings without changing
the retrieval interface, with the existing lossless path as a strict-reversibility fallback.

\paragraph{Graph engineering and knowledge-graph maturation.}
\slm{} maintains an entity knowledge graph with spreading-activation retrieval, but graph
construction currently relies on LLM-extracted entities without automatic temporal-edge
invalidation or community-level organization. Graphiti~\citep{graphitipaper} demonstrates
one concrete realization---an episode/entity/community architecture where a contradiction
sets an \code{expired\_at} timestamp on the superseded edge rather than deleting it, and a
label-propagation step groups entities into communities with maintained summaries.
Integrating temporal-edge invalidation into \slm{}'s entity-graph channel and adding
community summaries as an additional retrieval signal are well-scoped extensions that would
strengthen multi-hop and cross-entity recall.

\paragraph{Temporal memory: point-in-time reconstruction and calibration.}
\slm{}'s \code{as\_of} path performs superseded-fact demotion at query time but does not
reconstruct full system state at an arbitrary past instant---in particular it does not yet
apply \code{as\_of} to the backup store or the sqlite-vec ANN index. Extending
temporal-validity filtering to the ANN index and the restore path would close that scope
gap and make point-in-time audits complete; wiring the (currently unwired) Fisher--Langevin
posterior coupling would make the Ebbinghaus decay adaptive rather than fixed-parameter; and
an ablation against the temporal categories of LoCoMo~\citep{locomo} and
LongMemEval~\citep{longmemeval} would quantify the accuracy contribution of bi-temporal
demotion.

\paragraph{End-to-end agent retrieval benchmark.}
The evaluation here measures structural correctness at the component level; it is not a
retrieval-accuracy benchmark. A dedicated V4 suite should run \slm{} as a live recall server
across the LongMemEval~\citep{longmemeval} abilities and the LoCoMo~\citep{locomo}
categories, exercise multi-session write/recall cycles that trigger the generation fence and
the completion manifest, and report retrieval degradation under fault injection.
\citet{He2026MemoryArena} further show that LoCoMo-saturating agents still fail when
memorization and action are coupled across sessions, motivating an additional pass on
interdependent multi-session workflows.

\paragraph{SLM-Mesh at WAN scale and per-device revocation.}
Extending the mesh beyond a LAN raises three problems: consistency under arbitrary partition
(a causal-consistency or explicit CRDT model to remain correct across disconnect/reconnect);
per-device revocation (invalidating a decommissioned device's state across peers without a
central coordinator); and message-content scrubbing before transit (mesh \emph{state} values
pass a secrets filter, but \code{mesh\_send} payloads are not yet scrubbed pre-transmission).
Each is a scoped, engineering-grade extension with clear acceptance criteria.

\paragraph{Persistent generation fence.}
The generation fence is enforced in process memory and does not survive a restart;
persisting the fence epoch to the control store and rehydrating on startup is a
self-contained fix and remains open.

\paragraph{The watchdog item from the previous version, and what doing it taught.}
The previous version listed a kill-and-replace path for a stalled write-path worker,
with restart backoff and alerting, as future work. It was implemented. The result is
the first row of \Cref{tab:failure-class}: the revival monitor reported healthy while
semantic retrieval was off for over an hour, because it inferred liveness from a
probe search and discounted zero results---which is exactly what a dead embedder
produces. The stated fix was correct and insufficient, and the insufficiency
generalises: \textbf{a liveness check must not derive its signal from the output of
the component it is checking.} Whether that rule holds beyond this instance is
unproven and is the more interesting open question.

\paragraph{Showing that a repaired learner learns.}
\Cref{subsec:repair} removes the barrier on two of the three mechanisms and stops
short of the result that matters: at the last reading every arm was still at its
prior, because the outcomes that did arrive were themselves neutral. The open
question is not whether the signal now reaches the learner---the ledger shows it
can---but whether a reward derived from observed engagement moves a posterior in a
direction an operator would endorse. That needs a multi-operator deployment over
weeks and a held-out comparison against uniform selection, neither of which a
single-operator store supplies. The third mechanism is untouched: re-keying the
trust join onto the provenance table makes trust-modulated retention live with no
migration, and the join-liveness check already names the table and the row count.
Binding the loop ledger into the completion manifest would make execution and write
verification a single chain rather than two.

\paragraph{Isolating the cost of governance --- done differently than planned.}
The previous version of this paper named a build-time switch---one pipeline measured
with the envelope enabled and disabled---as the way to isolate this cost, and we no
longer think that is the right instrument. A switch is still a subtraction, and it
carries a worse problem: a governance control plane with a documented off position is
a governance control plane with a bypass, and shipping one to measure it would trade a
security property for a number. \Cref{sec:eval:envelope} takes the other route and
times the components in place, which needs no switch and leaves nothing to disable.

What remains open is the part measurement cannot settle. The envelope is dominated by
journal durability, so the interesting question is now an engineering one: how much of
that $7.7$\,ms survives group commit, batched journal writes, or a journal on a
separate device, and at what recovery cost. That is a design study we have not run.

\paragraph{Generalising the two invariants.}
The prior-distance and join-liveness checks of \Cref{subsec:invariants} are
implemented for this system's learners and guards, and the implementations are
specific to it in ways the pattern is not. Prior distance names two tables, assumes a
$\mathrm{Beta}(1,1)$ prior and tests a mean; it cannot inspect a contextual bandit, a
categorical posterior, a gradient learner or a learned ranker without being extended.
Join liveness registers one guard by table and column name and assumes a SQLite store.
What generalises is the question each asks, not the code that asks it. We have
general patterns and codebase-specific probes, and calling the second general would be
the kind of claim this paper exists to object to. A
useful test of the claim in \Cref{sec:eval:taxonomy} would be to run both against
other agent-memory implementations and report what they find, including nothing.

\paragraph{Machine-checked formal proofs.}
The Verified Design Invariants of \Cref{sec:formal} are established here by implementation
trace and fault-injection measurement. Upgrading them to machine-checked proofs (e.g.\ in
Lean~4 or TLA$^{+}$) would lift the guarantee from \emph{empirically upheld at $N{=}200$ per
trial} to \emph{proven for all executions conforming to the protocol}. The completion
manifest's \textsc{complete}/\textsc{degraded}/\textsc{failed} derivation is a natural first
target: a small, bounded state space tractable for model checking.

\section{Conclusion}
\label{sec:conclusion}

Agent memory is becoming shared infrastructure, judged by properties that
retrieval-accuracy benchmarks do not capture: isolation between tenants, verifiable
correctness of canonical operations, governed self-improvement, coordination without a
central server, first-class time, and operable governance. \slm{} addresses a specific,
under-served point in this space---a local-first runtime that co-locates multi-scope
isolation, verifiable memory transactions, and governance controls under one admission path,
without requiring a cloud service. The contribution is not any single mechanism---bi-temporal
storage, knowledge-graph retrieval, and GDPR-aware erasure each have strong prior art---but
their composition into one verifiable, governed, open-source runtime, together with a
physical-store transaction spine and a measured reliability evaluation.

The fault-injection result---2{,}199 of 2{,}200 repetitions---is deliberately narrow: it shows the
spine services upheld eleven deterministic structural properties at the component level;
it does not claim
retrieval-accuracy superiority over commercial services, nor does it substitute for an
end-to-end multi-agent benchmark. Within that honest scope, \slm{} delivers what its
architecture promises---tenants isolated by profile-scoped predicates on every retrieval
channel, with a write-time generation fence rejecting stale-epoch writes on top, every write through that path either verifiably complete across
its registered projection owners or recorded as a hash-checkable \textsc{degraded}
manifest, and erasure proved complete for the three canonical projection
stores---as one instance of \textbf{AI Reliability Engineering} for the memory layer. The
open threads of \Cref{sec:futurework}---near-lossless quantization as stores scale, graph
engineering, an end-to-end V4 retrieval benchmark, WAN-scale mesh, a persistent fence, and
machine-checked proofs---are each bounded extensions. \slm{} lays the control plane; the
reliability and quality properties reported here are a lower bound on what a production
deployment requires.

This version also retracts. A learning layer the previous version called governed had
never learned, and we can now show that exactly rather than argue it approximately.
A per-write governance cost we quoted differenced two arms that are not comparable. We
withdraw the figure, and we withdraw both of the explanations we later gave for it. A skill-evolution
pipeline we called a major contribution has never executed and is demoted to a
design. The reliability harness that returned 2{,}200 of 2{,}200 in the previous
version returns 2{,}199 of 2{,}200 here, and we report the one failure rather than
re-running until it disappears.

We think that is the more useful contribution. Ten mechanisms in this system were
built correctly, reached on a live call path, and left ineffective at their final
connection, every one of them silently---and the reason they were invisible is
structural rather than careless: \emph{implemented}, \emph{reachable} and
\emph{effective} are three different questions, and a signal derived from the
mechanism it evaluates cannot detect that mechanism's failure. The two invariants
this paper contributes answer the third question mechanically. The first thing they
did was find three of the ten in our own store; the second was to name a cause that
no counter in the subsystem could reach, and that named cause is repaired in a
public release a reader can install and inspect (\Cref{subsec:repair}). That is the
loop we want to argue for---detect, name, repair, and state plainly what remains
unproven---because holding a memory system to an evidence standard is only
meaningful if the standard is allowed to take claims away.

\section*{Availability and Full Report}
\label{sec:availability}

SuperLocalMemory is open source under AGPL-3.0 at
\url{https://github.com/qualixar/superlocalmemory} and is packaged for PyPI and npm as
\code{superlocalmemory}. It installs and runs on a single machine in three modes---fully
local (no provider in the memory path), local with an on-device model, and
provider-assisted---with no Docker, external graph database, or API key required for the
default local mode. The reliability evaluation harness of \Cref{sec:evaluation} and its
per-experiment result records are released alongside this paper. Reproducibility is
\emph{per-result}: the harness re-runs the fault-injection experiments (this version's
results were produced on Python~3.14.5, macOS-26.4.1-arm64, package 4.1.3),
whereas the carried-forward performance and V3 LoCoMo figures depend on artifacts (a
$\sim$1\,GB store and external model calls) that are not fully hash-pinned---retained
records note package version and platform but not a source commit or wheel hash. The exact
audited build is pinned at release.

This preprint is self-contained. A longer internal technical report---with extended
assumptions, full proofs, per-layer implementation detail, and an exhaustive capability
inventory---underlies this work; it is not part of this preprint's artifact.

\section*{Author Biography}
\label{sec:bio}

\textbf{Varun Pratap Bhardwaj} is a Solution Architect and the founder of Qualixar.
His work centres on
\textbf{AI Reliability Engineering}---the systematic application of reliability, correctness,
and governance to AI-augmented software. SuperLocalMemory is his line of work on local-first,
governed memory for AI agents, consolidated across the V2--V4 releases into the system
described here. ORCID: \texttt{0009-0002-8726-4289}.

\section*{Availability and Licensing}
\addcontentsline{toc}{section}{Availability and Licensing}

\slm{} is released as open source under the \textbf{GNU Affero General Public
License v3.0 (AGPL-3.0-or-later)}, with a separate commercial licence available for
closed-source, proprietary, or hosted deployment. Source, packaging metadata
(\code{pyproject.toml}, \code{package.json}) and the \code{LICENSE} file all declare
AGPL-3.0-or-later; earlier preprints in this series that stated a permissive licence are
superseded by this statement.

The reliability evidence in \Cref{sec:evaluation} is not merely reported: the full experiment
harness ships with the source under \code{benchmark/}, and the bundle in this paper is
regenerated by a single runner, \code{benchmark/run\_all.py}. A repository test asserts
that every experiment on disk is reachable from that runner, so a result cited here
cannot become unrunnable without the test suite failing. Readers are invited to
regenerate the numbers rather than take them on trust.

\section*{Author Contributions}
\addcontentsline{toc}{section}{Author Contributions}

\textbf{Varun Pratap Bhardwaj} conceived the system, designed and implemented SuperLocalMemory~4.0,
specified the reliability guarantees and their formal statements, designed and executed
the evaluation, and wrote the manuscript.

\textbf{Garima Singh} contributed to the product conception and requirements definition, shaped
the evaluation criteria and acceptance conditions from a product-ownership perspective,
and carried out sustained independent validation over several months on separate
hardware. The resulting defect reports, usability findings, and prioritisation feedback
materially changed the system's scope, its interface design, and which guarantees were
treated as load-bearing.

\textbf{Arun Pratap Bhardwaj} established and ran the iterative validation feedback loop across the
three operating modes and successive releases, contributed an operational-technology and
industrial-systems perspective on reliability and failure handling, and reported the
behavioural findings that drove several correctness fixes.

All authors reviewed the manuscript, take responsibility for the work as presented, and
approved the submitted version.

\bibliographystyle{plainnat}
\bibliography{references}

\end{document}